\documentclass[twoside]{article}

\usepackage[accepted]{aistats2026}
\usepackage{amssymb}
\usepackage{amsthm}
\usepackage{amsmath}
\usepackage{bm}
\usepackage{hyperref}
\usepackage{enumitem}
\usepackage{graphicx}
\usepackage{subcaption}
\usepackage{booktabs}
\usepackage{siunitx}
\usepackage[round]{natbib}

\newtheorem{theorem}{Theorem}
\newtheorem{assumption}{Assumption}
\newtheorem{lemma}{Lemma}
\newtheorem{proposition}{Proposition}
\newtheorem{corollary}{Corollary}
\newenvironment{sketchproof}{\begin{proof}[Proof sketch]}{\end{proof}}

\theoremstyle{remark}
\newtheorem{remark}{Remark}

\begin{document}

\twocolumn[

\aistatstitle{Provable Affine Identifiability of Nonlinear CCA Under Latent Distributional Priors}

\aistatsauthor{ Zhiwei Han \And Stefan Matthes \And Hao Shen }

\aistatsaddress{fortiss GmbH, Munich, Germany\\Technical university of Munich\\ \{han,matthes,shen\}@fortiss.org} ]

\begin{abstract}
In this work, we establish the sufficient conditions under which
nonlinear Canonical Correlation Analysis (CCA) recovers ground-truth
latent factors up to an affine transformation. By transporting the
analysis from the observation space to the source space, we extend
classical statistical results on
orthogonal polynomial expansions of bivariate distributions
to representation learning, proving affine identifiability under
specific distributional priors. We formally demonstrate that whitening
is strictly necessary to ensure the boundedness and well-conditioning
of the learned mappings. Furthermore, we bridge the gap between theory
and practice by proving that ridge-regularized empirical CCA converges
to its population counterpart in the finite-sample regime. Finally,
our findings provide a rigorous theoretical foundation explaining the
empirical success of recent correlation-based non-contrastive learning
methods. Experiments on synthetic and rendered image datasets, alongside
systematic ablations, validate the predicted recovery behavior and
illustrate the failure modes that arise when the assumptions are
violated.
\end{abstract}
\section{INTRODUCTION}

Identifying explanatory factors from raw sensory data is a 
fundamental challenge in machine learning.
Originated from the problem of blind source separation, 
independent component analysis (ICA) and its nonlinear variations 
\citep{comon1994independent, hyvarinen2000independent, hyvarinen2019nonlinear} 
aim to identify independent latent factors that improve 
downstream interpretability, controllability, fairness, and sample efficiency 
\citep{bengio2013representation, higgins2017beta, locatello2019challenging}. 
Unfortunately, directly learning these factors from nonlinear mixtures is 
unidentifiable without additional structural constraints or weak supervision 
\citep{hyvarinen1999nonlinear, locatello2019challenging, locatello2020weakly}.

% Without additional structure, identifying latent factors is provably
% impossible under nonlinear mixtures
% \citep{locatello2019challenging, hyvarinen1999nonlinear, locatello2020weakly}.
% However, leveraging structural priors, such as temporal dependency
% \citep{hyvarinen2016unsupervised}, nonstationarity \citep{hyvarinen2017nonlinear},
% auxiliary variables \citep{hyvarinen2019nonlinear, khemakhem2020variational},
% interventions \citep{brehmer2022weakly, lippe2022citris}, or mechanism sparsity
% \citep{lachapelle2022disentanglement, zheng2022identifiability}, enables the provable
% recovery of these factors. Within this framework, contrastive learning approaches
% have successfully established identifiability guarantees up to permutation
% transformations in multi-view and conditional settings
% \citep{zimmermann2021contrastive, daunhawer2023identifiability, matthes2023towards},
% elegantly connecting classical ICA theory with modern SSL.

Canonical Correlation Analysis (CCA) \citep{hotelling1936relation} 
is another classic method, which aims to learn representations 
by maximizing the cross-correlation between paired views. 
While nonlinear CCA and its extensions 
\citep{bach2002kernel, fukumizu2007statistical, andrew2013deep, hardoon2011sparse, ermolov2021whitening, zbontar2021barlow, bardes2021vicreg} 
demonstrate empirical performance comparable to strong contrastive
learning \citep{gutmann2010noise, oord2018representation} baselines, 
the theoretical understanding of nonlinear CCA identifiability
lags behind. Contrastive learning now offers robust 
block- and factor-level identifiability guarantees by leveraging specific 
data or model assumptions 
\citep{von2021self, yao2024multi, zimmermann2021contrastive, daunhawer2023identifiability, matthes2023towards, hyvarinen2016unsupervised, brehmer2022weakly, lachapelle2022disentanglement}. 
In contrast, existing analyses for nonlinear CCA either assume restricted 
post-nonlinear mixing \citep{lyu2020nonlinear} or only guarantee recovery up 
to broad, arbitrary invertible transformations 
\citep{lyu2021understanding, karakasis2023revisiting}. 
This discrepancy raises a fundamental open question: 
\textit{Under what conditions can nonlinear 
CCA provide strictly stronger identifiability guarantees, up to simpler 
transformations that are comparable to those of contrastive learning?}

% As a principled approach to representation learning, 
% canonical correlation analysis (CCA) \citep{hotelling1936relation} 
% maximize the cross-correlation of the representations between paired views. 
% Empirically, CCA and its nonlinear extensions 
% \citep{bach2002kernel, fukumizu2007statistical, andrew2013deep, hardoon2011sparse,ermolov2021whitening,zbontar2021barlow, bardes2021vicreg} 
% demonstrate competitive performance with contrastive methods, 
% another line of methods \citep{gutmann2010noise,oord2018representation} 
% showcase etremendoeus success in both theoretical and empirical settings. 
% Unlike contrastive learning, demonstrating block to disentanglement level
% identifiability guarantees 
% \citep{von2021self,yao2024multi,zimmermann2021contrastive, daunhawer2023identifiability, matthes2023towards}
%  by leverating model or data assumptions\citep{hyvarinen2016unsupervised, brehmer2022weakly,lachapelle2022disentanglement},
% the theoretical understanding in terms of identifiability of CCA lags behind 
% empirical success. The existing
% factor-level identifiability analyses for nonlinear CCA either assume restricted 
% post-nonlinear mixing \citep{lyu2020nonlinear} or only guarantee recovery up to 
% arbitrary invertible transformations \citep{lyu2021understanding, karakasis2023revisiting}.
% This raises an open question: \textit{Under what conditions can nonlinear 
% CCA provide strictly stronger identifiability guarantees, up to simpler 
% transformations that are comparable to those of contrastive learning?}

This work answers the above question by revisiting the classical theory of
bivariate distributions \citep{lancaster1958structure, eagleson1964polynomial}.
We establish the sufficient conditions under which nonlinear CCA provably recovers ground-truth
latent variables up to an affine transformation. Crucially, we formalize the role of representation
whitening in ensuring the boundedness and well-conditioning strictly necessary for this identifiability.
We empirically validate these theoretical guarantees and systematically ablate our core assumptions to
characterize failure modes when these conditions are violated. We summarize our contributions as follows:
\begin{enumerate}
\item We characterize the invariance of the population CCA objective and its maximizers under reparameterization via composition with the generators, enabling direct analysis in the latent source space (\autoref{prop:rep-inv}).
\item We extend the classical theory of bivariate distributions to representation learning and derive affine identifiability results for nonlinear CCA in the population setting across a broad class of latent distributional priors (\autoref{thm:affine-id}).
\item We analyze ridge-regularized empirical CCA and show its convergence to the population counterpart, linking the identifiability results to the finite-sample regime (\autoref{thm:emp-consistency}).
\item We assess these theoretical predictions across candidate latent distributions using both a fully controlled synthetic dataset and a rendered 3D image dataset.
\item We study the failure modes of nonlinear CCA under partial violations of the core assumptions through systematic ablations.
\end{enumerate}
\section{RELATED WORK}
\paragraph{Disentangled Representation Learning.}
Disentangled representation learning aims to isolate
independent explanatory factors of variation, originating from blind source
separation \citep{cardoso1998blind,comon1994independent}. Because disentanglement is
provably impossible for i.i.d.\ nonlinear mixtures without structural assumptions \citep{locatello2019challenging,hyvarinen1999nonlinear}, research has focused on exploiting suitable inductive biases to invert nonlinear generative processes.
One prevalent approach leverages shared or co-observed variables,
such as temporal dynamics \citep{hyvarinen2016unsupervised}, nonstationarity \citep{hyvarinen2017nonlinear}, auxiliary labels \citep{hyvarinen2019nonlinear,khemakhem2020variational}, weak supervision \citep{locatello2020weakly,shu2019weakly},
and multi-view observations
\citep{daunhawer2023identifiability, lyu2020nonlinear, von2021self}. A complementary direction restricts the model class or source distribution, enforcing structural constraints like mechanism sparsity \citep{lachapelle2022disentanglement}, conformal mappings \citep{buchholz2022function}, or specific latent priors (e.g., exponential families \citep{hyvarinen2019nonlinear}, causal structures \citep{shen2022weakly},
or energy-based models \citep{khemakhem2020ice}).
Our work aligns with the latter by imposing latent distributional priors
within a two-view setting.

\paragraph{Nonlinear CCA.}
Nonlinear CCA extends classical CCA to arbitrary nonlinear mappings via
kernels \citep{fukumizu2007statistical} or deep neural networks
\citep{andrew2013deep, benton2017deep}. While these methods excel at maximizing
cross-view correlations, achieving strictly identifiable representations
was not their original objective. Recent efforts have explored the identifiability
of nonlinear CCA-based models
\citep{lyu2020nonlinear, lyu2021understanding, karakasis2023revisiting, sidiropoulos2022canonical}.
However, existing theoretical guarantees of factor-level
identifiability rely heavily on restrictive
post-nonlinear mixing assumptions \citep{lyu2020nonlinear} or only guarantee
recovery up to broad equivalence classes, such as arbitrary invertible
transformations \citep{lyu2021understanding, karakasis2023revisiting}.
In contrast, modern contrastive learning frameworks successfully invert
the data-generating process to recover latent variables up to strict affine
or permutation ambiguities
\citep{matthes2023towards,zimmermann2021contrastive,daunhawer2023identifiability}.
A detailed theoretical comparison between the identifiability frameworks of nonlinear
CCA and contrastive learning is provided in \autoref{app:cca_vs_cl}.
An equivalently rigorous, affine identifiability theory for nonlinear
CCA remains absent.

We bridge this gap by revisiting classical bivariate distribution theory,
which establishes that linear mappings uniquely maximize canonical
correlations under Gaussian priors \citep{lancaster1958structure}, a result
later generalized to broader distribution families \citep{eagleson1964polynomial}.
Building on these foundations, we extend this classical theory to modern
representation learning. By proving the affine identifiability of nonlinear
CCA under a family of latent distributional priors, we cement CCA as a
theoretically principled alternative to contrastive learning for exact
latent recovery, moving beyond mere view alignment.
\section{CANONICAL CORRELATION ANALYSIS FOR DISENTANGLED REPRESENTATION}
\paragraph{Notations.}Throughout this work, scalars are denoted by plain letters, (random) 
vectors or vector-valued functions by bold lowercase, matrices by bold 
uppercase symbols, and sets/spaces by calligraphic letters.
\subsection{Data Generating Process}
\label{subsec:dgp}

We consider a latent pair $(\mathbf{s},\mathbf{s}') \in \mathcal{S}\times\mathcal{S}$,
where $\mathcal{S}\subseteq\mathbb{R}^{d_{\mathcal S}}$ and $d_{\mathcal S}\ge 2$.
For the main population identifiability theorem, we
assume that $(\mathbf{s},\mathbf{s}')$ is a non-degenerate jointly Gaussian
pair and parameterize its dependence through its canonical correlations, which
are the natural quantities for CCA.
Our primary analysis focuses on the Gaussian
case with extensions to other latent distribution families admitting orthogonal
polynomial expansions. Formal proofs for these Lancaster-type
distributions \citep{lancaster1958structure}, namely the negative
binomial, Gamma, Poisson, and hypergeometric cases, are deferred to
\autoref{sec:sketch}.

\begin{assumption}[Non-degenerate Joint Gaussian Latent Pair]
\label{assump:gaussian-latent-pair}
The latent pair $(\mathbf{s},\mathbf{s}')\in\mathcal{S}\times\mathcal{S}$ is jointly
Gaussian with mean $\bm{\mu}=(\bm{\mu}_{\mathbf{s}},\bm{\mu}_{\mathbf{s}'})$ and block
covariance
\[
    \mathbf{\Sigma} =
    \begin{bmatrix}
    \mathbf{\Sigma}_{\mathbf{s}\mathbf{s}} & \mathbf{\Sigma}_{\mathbf{s}\mathbf{s}'}\\
    \mathbf{\Sigma}_{\mathbf{s}'\mathbf{s}} & \mathbf{\Sigma}_{\mathbf{s}'\mathbf{s}'}
    \end{bmatrix},
    \quad \text{where} \quad
    \mathbf{\Sigma}_{\mathbf{s}\mathbf{s}} \succ 0, \: \mathbf{\Sigma}_{\mathbf{s}'\mathbf{s}'} \succ 0.
\]
Let $\mathbf{K} := \mathbf{\Sigma}_{\mathbf{s}\mathbf{s}}^{-1/2} \mathbf{\Sigma}_{\mathbf{s}\mathbf{s}'} \mathbf{\Sigma}_{\mathbf{s}'\mathbf{s}'}^{-1/2}$
be the normalized cross-covariance matrix. We assume the singular values of $\mathbf{K}$, i.e.,
the canonical correlations between $\mathbf{s}$ and $\mathbf{s}'$, satisfy:
\[
    1 > \rho_1 \ge \rho_2 \ge \cdots \ge \rho_{d_{\mathcal S}} > 0.
\]
\end{assumption}

The spectral bounds on $\rho_i$ guarantee a full-rank
correlated subspace while excluding deterministic equivalence between the views.
We then give the practical implication of this assumption
and present the corresponding latent model for better structural clarity.
\begin{remark}[Canonical Additive Latent Model]
\label{rem:canonical-additive}
Let $\mathbf{K} = \mathbf{U} \mathbf{\Lambda} \mathbf{V}^\top$ be the singular value decomposition, where $\mathbf{\Lambda} = \operatorname{diag}(\rho_1,\dots,\rho_{d_{\mathcal S}})$. Defining the canonical coordinates
\[
    \bar{\mathbf s} := \mathbf U^\top \mathbf{\Sigma}_{\mathbf{s}\mathbf{s}}^{-1/2} (\mathbf s-\bm{\mu}_{\mathbf s}), \quad
    \bar{\mathbf s}' := \mathbf V^\top \mathbf{\Sigma}_{\mathbf{s}'\mathbf{s}'}^{-1/2} (\mathbf s'-\bm{\mu}_{\mathbf s'}),
\]
yields $\mathrm{Cov}(\bar{\mathbf s}) = \mathrm{Cov}(\bar{\mathbf s}') = \mathbf{I}_{d_{\mathcal S}}$ and $\mathrm{Cov}(\bar{\mathbf s},\bar{\mathbf s}') = \mathbf{\Lambda}$. Consequently, there exist mutually independent, centered Gaussian vectors $\mathbf{a}, \mathbf{b}, \mathbf{c}$ such that
\(
    (\bar{\mathbf s},\bar{\mathbf s}') \mathrel{:=} (\mathbf{a}+\mathbf{c}, \mathbf{b}+\mathbf{c})
\)
with $\mathrm{Cov}(\mathbf{c}) = \mathbf{\Lambda}$ and $\mathrm{Cov}(\mathbf{a}) = \mathrm{Cov}(\mathbf{b}) = \mathbf{I}_{d_{\mathcal S}} - \mathbf{\Lambda}$.
Thus, an additive shared-private interpretation is inherently guaranteed \citep{cramer1994eigenschaft}
in the whitened
canonical space, even if the cross-covariance matrix $\mathbf{\Sigma}_{\mathbf{s}\mathbf{s}'}$
is asymmetric.
\end{remark}

To construct the joint distributions in our experiments, we employ the explicit
additive generative model as follows:
\begin{equation}
\label{equ:additive}
    \mathbf{s} = \mathbf{a}+\mathbf{c}, \qquad \mathbf{s}' = \mathbf{b}+\mathbf{c},
\end{equation}
where $\mathbf{a}, \mathbf{b}, \mathbf{c}$ are mutually independent latent vectors.
We emphasize that \autoref{equ:additive} serves as an experimental device rather than a prerequisite
for our population identifiability theorem.

Let $\mathbf{g}\colon \mathcal{S} \to \mathcal{X} \subset \mathbb{R}^{d_\mathcal{X}}$
and $\mathbf{g}'\colon \mathcal{S} \to \mathcal{X}' \subset \mathbb{R}^{d_{\mathcal{X}'}}$
be injective and Borel-measurable mappings that generate the high-dimensional
observed pairs $(\mathbf{x}, \mathbf{x}') = (\mathbf{g}(\mathbf{s}),
\mathbf{g}'(\mathbf{s}'))$, where
$\mathcal{X}, \mathcal{X}'$
denote the observation spaces of two modalities with
$d_{\mathcal{S}} \ll d_{\mathcal{X}},
d_{\mathcal{X}'}$. Under \autoref{assump:gaussian-latent-pair},
our objective is to learn
encoders $\mathbf{f}\colon \mathcal{X} \to \mathcal{Z}$ and $\mathbf{f}'\colon
\mathcal{X}' \to \mathcal{Z}$ that invert the generative
 process to recover the true
latent sources $(\mathbf{s}, \mathbf{s}')$.
To establish baseline affine identifiability, we assume a
dimension-matched latent space ($d_{\mathcal{Z}} = d_{\mathcal{S}}$), deferring
the analysis of mismatched regimes ($d_{\mathcal{Z}} \neq d_{\mathcal{S}}$) to
the ablation study.
\subsection{Canonical Correlation Analysis}
Nonlinear CCA aims to learn a pair of nonlinear
encoders $\mathbf{f}\colon \mathcal{X}\to \mathcal{Z}$ and $\mathbf{f}'\colon
\mathcal{X}'\to \mathcal{Z}$ that maximize the sum of the singular values of
their normalized cross-covariance. To ensure well-conditioned representations, we restrict
the search space to zero-mean, identity-covariance encoders:

\begin{assumption}[Whitened Encoder Classes]\label{assump:function-class}
Let $\mathcal{H}_{\mathcal{X}} \subset L^2(P_{\mathbf{x}}; \mathbb{R}^{d_{\mathcal{Z}}})$
be a Hilbert space of square-integrable vector-valued functions. For any base
encoder $\mathbf{f} \in \mathcal{H}_{\mathcal{X}}$ with a positive definite covariance
matrix $\mathbf{\Sigma}_{\mathbf{f}} \succ 0$, let $\mathbf{W}_{\mathbf{f}} \in
\mathrm{GL}(d_{\mathcal{Z}})$ be a whitening matrix satisfying
$\mathbf{W}_{\mathbf{f}} \mathbf{\Sigma}_{\mathbf{f}} \mathbf{W}_{\mathbf{f}}^\top
= \mathbf{I}_{d_{\mathcal{Z}}}$. We define the whitened encoder class on $\mathcal{X}$ as
\begin{equation*}
\tilde{\mathcal{F}}_{\mathcal{X}} := \Big\{\, \mathbf{W}_{\mathbf{f}}\big(\mathbf{f}(\mathbf{x})
- \mathbb{E}[\mathbf{f}(\mathbf{x})]\big) : \mathbf{f} \in \mathcal{H}_{\mathcal{X}}, \,
\mathbf{\Sigma}_{\mathbf{f}} \succ 0 \,\Big\},
\end{equation*}
and assume it is closed under orthogonal transformations, i.e.,
$\mathbf{Q}\tilde{\mathbf{f}} \in \tilde{\mathcal{F}}_{\mathcal{X}}$ for all
$\tilde{\mathbf{f}} \in \tilde{\mathcal{F}}_{\mathcal{X}}$ and $\mathbf{Q} \in
O(d_{\mathcal{Z}})$. The corresponding class $\tilde{\mathcal{F}}'_{\mathcal{X}'}$
is defined analogously for $\mathcal{H}_{\mathcal{X}'} \subset L^2(P_{\mathbf{x}'};
\mathbb{R}^{d_{\mathcal{Z}}})$.
\end{assumption}

\paragraph{CCA Population Objective.} Given these feasible encoder classes, nonlinear
CCA maximizes the following population objective:
\begin{equation}\label{equ:CCA_objective}
\max_{\tilde{\mathbf{f}} \in \tilde{\mathcal{F}}_{\mathcal{X}}, \, \tilde{\mathbf{f}}'
\in \tilde{\mathcal{F}}'_{\mathcal{X}'}} J(\tilde{\mathbf{f}}, \tilde{\mathbf{f}}')
= \sum_{i=1}^{d_{\mathcal{Z}}} \sigma_i\big(\mathrm{Cov}(\tilde{\mathbf{f}}(\mathbf{x}),
\tilde{\mathbf{f}}'(\mathbf{x}'))\big),
\end{equation}
where $\sigma_i(\cdot)$ denotes the $i$-th largest singular value
and $\mathrm{Cov}(\cdot)$ the population covariance.
Due to the orthogonal invariance of the objective and the post-orthogonal
closure of the encoder classes, every population maximizer generates an
equivalent \(O(d_{\mathcal Z})\times O(d_{\mathcal Z})\)-orbit of maximizers.
In general, this does not imply that all maximizers lie in a single orbit.
Stronger single-orbit statement is obtained later under the additional
assumptions of \autoref{thm:affine-id} as shown in
Appendix \ref{app:cca_orbit} and \autoref{cor:cca_single_orbit}.

\subsection{Population Affine Identifiability}
\label{subsec:identifiability}
In this section, we derive precise conditions under which solving
nonlinear CCA leads to provable affine identification of the ground-truth
latent factors at the population level.
To leverage the distributional priors of the source,
we begin by transporting the CCA problem
from the observation domain to the underlying source domain.
By composing the observation space encoders with the generators,
$\mathbf h=\mathbf f\!\circ\!\mathbf g$ and
$\mathbf h'=\mathbf f'\!\circ\!\mathbf g'$,
we re-express the learned latent via representation maps
$\mathbf h, \mathbf h'$ and ground-truth latent pair.
This change of variables preserves the first- and second-order
structure of the CCA problem,
so the whitening constraints and the CCA objective are unaffected.
We formally justify the invariance of transporting CCA
problems from observation spaces to source space.
\begin{proposition}[Reparameterization Invariance and Representational Universality]
\label{prop:rep-inv}
Let's assume $\mathcal{S}, \mathcal{X}, \mathcal{X}'$ be standard Borel spaces,
and let $\mathbf{g}, \mathbf{g}'$ be injective, Borel-measurable mappings.
Further assume the base encoders underlying the whitened classes $\tilde{\mathcal{F}}_{\mathcal{X}},
\tilde{\mathcal{F}}'_{\mathcal{X}'}$ in \autoref{assump:function-class} are
dense in $L^2(P_{\mathbf{x}}; \mathbb{R}^{d_{\mathcal{Z}}})$ and
$L^2(P_{\mathbf{x}'}; \mathbb{R}^{d_{\mathcal{Z}}})$, respectively.
Define whitened representable latent classes:
\[
\tilde{\mathcal F}_{\mathcal{S}}:=\{\tilde{\mathbf f}\!\circ \mathbf{g}:\ \tilde{\mathbf f}\in\tilde{\mathcal F}_{\mathcal{X}}\},
\qquad
\tilde{\mathcal F}'_{\mathcal{S}}:=\{\tilde{\mathbf f}'\!\circ \mathbf{g}':\ \tilde{\mathbf f}'\in\tilde{\mathcal F}'_{\mathcal{X}'}\},
\]
and the whitened feasible latent classes:
\begin{align*}
    \hat{\mathcal F}_{\mathcal S} &:= \{\, \hat{\mathbf h} \in L^2(P_{\mathbf s};\mathbb R^{d_\mathcal Z}):
\mathbb E[\hat{\mathbf h}]=0, \mathrm{Cov}(\hat{\mathbf h})=\mathbf I_{d_\mathcal Z} \,\},\\
\hat{\mathcal F'_\mathcal S} &:= \{\, \hat{\mathbf h'} \in L^2(P_{\mathbf s'};\mathbb R^{d_\mathcal Z}):
\mathbb E[\hat{\mathbf h'}]=0, \mathrm{Cov}(\hat{\mathbf h'})=\mathbf I_{d_\mathcal Z} \,\}.
\end{align*}
Consider the source-space CCA objective defined over the feasible
whitened latent classes on $\mathcal{S} \times \mathcal{S}$:
for all $\hat{\mathbf{h}} \in \hat{\mathcal{F}}_{\mathcal{S}}$ and
$\hat{\mathbf{h}}' \in \hat{\mathcal{F}}'_{\mathcal{S}}$,
\begin{equation}
\label{equ:CCA_objective_source}
J_\mathcal{S}(\hat{\mathbf h}, \hat{\mathbf h'})=\sum_{i=1}^{d_\mathcal{Z}}
\sigma_i\big(\mathrm{Cov}(\hat{\mathbf h}(\mathbf{s}),\,\hat{\mathbf h'}(\mathbf{s'}))\big),
\end{equation}
where $\sigma_i(\cdot)$ denotes the $i$-th largest singular value. Then the following properties hold.

\begin{enumerate}[label=\arabic*.]
\item \textbf{Objective Preservation.}
$\forall (\tilde{\mathbf f},\tilde{\mathbf f}')\in \tilde{\mathcal{F}}_{\mathcal{X}}\times\tilde{\mathcal{F}}'_{\mathcal{X}'}$,
\(
J(\tilde{\mathbf f},\tilde{\mathbf f}')\;=\;
J_\mathcal S\big(\tilde{\mathbf f}\!\circ \mathbf{g},\ \tilde{\mathbf f}'\!\circ \mathbf{g}'\big).
\)

\item \textbf{Maximizer Correspondence.}
A pair
$(\tilde{\mathbf f}^*,\tilde{\mathbf f}^{\prime *})$
maximizes $J$ over
$\tilde{\mathcal F}_{\mathcal X}\times \tilde{\mathcal F}'_{\mathcal X'}$
if and only if
$(\tilde{\mathbf f}^*\circ \mathbf g,\tilde{\mathbf f}^{\prime *}\circ \mathbf g')$
maximizes $J_{\mathcal S}$ over
$\tilde{\mathcal F}_{\mathcal S}\times \tilde{\mathcal F}'_{\mathcal S}$.

\item \textbf{Representation Universality.}
For any $\hat{\mathbf h}\in\hat{\mathcal F}_{\mathcal S}$
and $\hat{\mathbf h}'\in\hat{\mathcal F}'_{\mathcal S}$,
and any $\epsilon,\epsilon'>0$,
there exist representable maps
$\tilde{\mathbf h}\in\tilde{\mathcal F}_{\mathcal S}$ and
$\tilde{\mathbf h}'\in\tilde{\mathcal F}'_{\mathcal S}$ such that
$\mathbb{E}[\|\hat{\mathbf{h}}(\mathbf{s}) - \tilde{\mathbf{h}}(\mathbf{s})\|^2] < \epsilon^2$
and $\mathbb{E}[\|\hat{\mathbf{h}}'(\mathbf{s}') - \tilde{\mathbf{h}}'(\mathbf{s}')\|^2] < \epsilon'^2$.
Consequently, since $J_{\mathcal S}$ is continuous under the product $L^2$ topology,
\[
\sup_{\tilde{\mathcal F}_{\mathcal X}\times\tilde{\mathcal F}'_{\mathcal X'}} J
=
\sup_{\tilde{\mathcal F}_{\mathcal S}\times\tilde{\mathcal F}'_{\mathcal S}} J_{\mathcal S}
=
\sup_{\hat{\mathcal F}_{\mathcal S}\times\hat{\mathcal F}'_{\mathcal S}} J_{\mathcal S}.
\]
\end{enumerate}
\end{proposition}

\begin{sketchproof}
    The pushforward identity preserves all moments, so the CCA 
    objective and whitening constraints are invariant under composition 
    with $\mathbf{g},\mathbf{g}'$. Injectivity on standard Borel spaces 
    yields measurable inverses, allowing any target in 
    $\hat{\mathcal F}_{\mathcal S}$ or $\hat{\mathcal F'}_{\mathcal S}$ 
    to be pulled back and approximated by the dense base encoders. 
    See Appendix \ref{app:proof-rep-inv} for details.
\end{sketchproof}

By \autoref{prop:rep-inv}, the observation-space CCA problem
is isometrically equivalent to the representable source-space problem, and this representable
problem attains the same optimal value as the full feasible source-space problem.
To establish affine identifiability, it suffices to characterize the maximizers
of $J_{\mathcal S}$ over $\hat{\mathcal F}_{\mathcal S}\times\hat{\mathcal F}'_{\mathcal S}$,
any observation-space maximizer induces one such source-space maximizer.

\begin{assumption}[First-Order Canonical Dominance]
\label{assump:canonical-correlation-separation}
Under \autoref{assump:gaussian-latent-pair}, we further assume that the
smallest canonical correlation $\rho_{d_{\mathcal S}}$ is
strictly larger than the product of any two canonical
correlations, i.e., $\rho_{d_{\mathcal S}} > \rho_i \rho_j, \forall 1 \le i \le j \le d_{\mathcal S}$. Equivalently,
$\rho_{d_{\mathcal S}} > \rho_1^2$, where $\rho_1$ is the largest 
canonical correlation.
\end{assumption}

We then state our central population identifiability result as follows:
\begin{theorem}[Population Affine Identifiability]
\label{thm:affine-id}
Assume the conditions of \autoref{prop:rep-inv},
\autoref{assump:gaussian-latent-pair}, and
\autoref{assump:canonical-correlation-separation}, and let
$d_{\mathcal Z}=d_{\mathcal S}$.
For the whitened encoder classes in \autoref{assump:function-class},
any population maximizer pair
$(\tilde{\mathbf f}^*,\tilde{\mathbf f}^{\prime *})$
of \autoref{equ:CCA_objective} identifies the marginally whitened latent
factors up to orthogonal transformations. Specifically, there exist
orthogonal matrices $\mathbf Q,\mathbf Q' \in O(d_{\mathcal Z})$ such that
\[
\tilde{\mathbf h}^*(\mathbf s)
:= (\tilde{\mathbf f}^*\circ \mathbf g)(\mathbf s)
= \mathbf Q\,\mathbf\Sigma_{\mathbf s\mathbf s}^{-1/2}
(\mathbf s-\bm\mu_{\mathbf s}),
\]
\[
\tilde{\mathbf h}^{\prime *}(\mathbf s')
:= (\tilde{\mathbf f}^{\prime *}\circ \mathbf g')(\mathbf s')
= \mathbf Q'\,\mathbf\Sigma_{\mathbf s'\mathbf s'}^{-1/2}
(\mathbf s'-\bm\mu_{\mathbf s'}).
\]
Equivalently, the corresponding unwhitened representation maps recover
$\mathbf s$ and $\mathbf s'$ up to invertible affine transformations.
\end{theorem}
\begin{sketchproof}
    By \autoref{prop:rep-inv}, the CCA optimization can be transported
    to the source space, allowing us to characterize the optimal
    representation mappings directly in terms of the ground-truth latents.
    For joint
    Gaussian priors,
    Mehler-Hermite expansion \citep{mehler1866ueber}
     effectively diagonalizes the
    CCA objective, revealing that the canonical correlations of the
    learned features are bounded by combinations of the true latent
    canonical correlations.
    under \autoref{assump:canonical-correlation-separation},
    the CCA objective
    is uniquely maximized by selecting only the first-order Hermite
    polynomials and all higher-order nonlinearities vanish.
    A complete proof is deferred to Appendix \ref{app:proof-affine-id}.
\end{sketchproof}
At the population optimum under Gaussian priors with
$d_{\mathcal{Z}} = d_{\mathcal{S}}$,
the learned encoders successfully invert the data-generating process up to an
affine ambiguity, i.e., centering, scaling, and an orthogonal
rotation. In the proof of \autoref{thm:affine-id}, we formally prove that explicit
whitening or unit-variance regularization, as employed by
non-contrastive methods
like Barlow Twins, is strictly necessary to ensure the boundedness and
well-conditioning of the learned mappings.
This provides a partial but rigorous explanation for the
effectiveness of whitening-based non-contrastive methods.
While our theoretical guarantees
require matched dimensions ($d_{\mathcal{S}} = d_{\mathcal{Z}}$),
we extend our analysis to mismatched regimes, i.e.,
$d_{\mathcal{S}} \neq d_{\mathcal{Z}}$ and
finite-sample convergence in the subsequent empirical sections.
A detailed discussion of the role of canonical-correlation separation and the consequences of relaxing this condition is provided in \autoref{app:remarks}.
\subsection{Empirical Estimation and Statistical Consistency}
While population identifiability characterizes the infinite-sample
optimum of \autoref{equ:CCA_objective}, practical representation
learning requires statistical consistency with finite data. To bridge
this gap, we introduce a regularized empirical CCA estimator
and analyze its statistical consistency in
the finite-sample regime. We
prove that as the regularization parameter decays asymptotically,
the empirical maximizers strictly converge to the population solution,
formally linking theoretical identifiability to finite-sample learnability.

To construct this empirical objective, we parameterize the nonlinear
encoders $\mathbf{f}_{\bm\theta}\colon \mathbb{R}^{d_\mathcal{X}} \to
\mathbb{R}^{d_\mathcal{Z}}$ and $\mathbf{f}'_{\bm\theta'}\colon
\mathbb{R}^{d_{\mathcal{X}'}} \to \mathbb{R}^{d_\mathcal{Z}}$ using
neural networks with sufficient capacity to act as universal function
approximators. Let $\bm\theta$ and $\bm\theta'$ denote their respective
weights. The regularized finite-sample objective is formulated as follows:

\paragraph{Empirical Objective.}
Let $\mathbf{X} \in \mathbb{R}^{n \times d_{\mathcal{X}}}$ and $\mathbf{X}' \in
\mathbb{R}^{n \times d_{\mathcal{X}'}}$ denote observation matrices comprising
$n$ mutually independent samples drawn from $P_{\mathbf{x}\mathbf{x}'}$. We define the
learned representations as $\mathbf{Z} = \mathbf{f}_{\bm{\theta}}(\mathbf{X})$
and $\mathbf{Z}' = \mathbf{f}'_{\bm{\theta}'}(\mathbf{X}') \in \mathbb{R}^{n
\times d_{\mathcal{Z}}}$. Let $\mathbf{Z}_c$ and $\mathbf{Z}'_c$ denote their
respective mean-centered counterparts. The empirical auto- and cross-covariance matrices
are given by $\hat{\mathbf{\Sigma}}_{\mathbf{z}\mathbf{z}} := \frac{1}{n-1}
\mathbf{Z}_c^\top \mathbf{Z}_c$, $\hat{\mathbf{\Sigma}}_{\mathbf{z}'\mathbf{z}'}
:= \frac{1}{n-1}{\mathbf{Z}'_c}^\top \mathbf{Z}'_c$, and
$\hat{\mathbf{\Sigma}}_{\mathbf{z}\mathbf{z}'} := \frac{1}{n-1}\mathbf{Z}_c^\top
\mathbf{Z}'_c$. To ensure well-conditioning in the finite-sample regime, we
apply a ridge penalty $\epsilon > 0$ to construct the regularized whitening
matrices:
\[
\hat{\mathbf{W}}_{\mathbf{z}} := (\hat{\mathbf{\Sigma}}_{\mathbf{z}\mathbf{z}}
+ \epsilon\mathbf{I})^{-1/2}, \quad \hat{\mathbf{W}}_{\mathbf{z}'} :=
(\hat{\mathbf{\Sigma}}_{\mathbf{z}'\mathbf{z}'} + \epsilon\mathbf{I})^{-1/2}.
\]
The empirical CCA objective maximizes the nuclear norm, i.e., the sum of singular
values,  of the normalized cross-covariance $\hat{\mathbf{K}} := \hat{\mathbf{W}}_{\mathbf{z}}
\hat{\mathbf{\Sigma}}_{\mathbf{z}\mathbf{z}'} \hat{\mathbf{W}}_{\mathbf{z}'}$,
formulated via the variational trace characterization:
\begin{align} \label{equ:empirical_CCA}
\max_{\bm{\theta}, \bm{\theta}'} \hat{J}(\bm{\theta}, \bm{\theta}')
&= \max_{\bm{\theta}, \bm{\theta}'} \sum_{k=1}^{d_{\mathcal{Z}}} \sigma_k(\hat{\mathbf{K}}) \nonumber \\
&= \max_{\substack{\bm{\theta}, \bm{\theta}' \\
\mathbf{U}^\top \mathbf{U} = \mathbf{I} \\ \mathbf{V}^\top \mathbf{V} = \mathbf{I}}}
\operatorname{tr}\big(\mathbf{U}^\top \hat{\mathbf{K}}\mathbf{V}\big).
\end{align}
The following theorem establishes our central statistical
consistency result.
By demonstrating the uniform convergence of the
empirical covariance operators
and the stability of both the regularized whitening and
spectral mappings,
we prove the asymptotic consistency of the empirical maximizers.

\begin{theorem}[Consistency of Empirical Maximizers]
\label{thm:emp-consistency}
Assume the conditions of \autoref{thm:affine-id} hold with $d_{\mathcal{Z}} =
d_{\mathcal{S}}$. Let $(\tilde{\mathbf{f}}_{\bm{\theta}}, \tilde{\mathbf{f}}'_{\bm{\theta}'})$
denote a pair of empirical whitened encoders and consider the regularized empirical
objective $\hat{J}$ in \autoref{equ:empirical_CCA} with ridge penalty $\epsilon
\to 0^+$. Suppose the following standard consistency assumptions hold as $n$ goes
to infinity:
\begin{enumerate}[label=A\arabic*.]
    \item \textbf{Realizability and Orbit Uniqueness.} The population maximizer
    of $J$ in \autoref{equ:CCA_objective} exists in the whitened encoder classes
    in \autoref{assump:function-class} and is unique up to postorthogonal
    transformations.
    \item \textbf{Capacity and Nondegeneracy.} The second moments of
    $(\tilde{\mathbf{f}}_{\bm{\theta}}, \tilde{\mathbf{f}}'_{\bm{\theta}'})$
    are uniformly bounded,
    and the empirical covariances $\hat{\mathbf{\Sigma}}_{\mathbf{zz}}$ and
    $\hat{\mathbf{\Sigma}}_{\mathbf{z}'\mathbf{z}'}$ are uniformly symmetric
    positive definite.
    \item \textbf{Uniform Convergence and Ridge Schedule.} The empirical auto-
    and cross-covariances converge uniformly at a rate faster than the ridge
    penalty, i.e., \(
    \|\hat{\mathbf \Sigma}_{\mathbf z\mathbf z}-\mathbf \Sigma_{\mathbf z\mathbf z}\|=o_p(\epsilon), \|\hat{\mathbf \Sigma}_{\mathbf z'\mathbf z'}-\mathbf \Sigma_{\mathbf z'\mathbf z'}\|=o_p(\epsilon),
     \|\hat{\mathbf \Sigma}_{\mathbf z\mathbf z'}-\mathbf \Sigma_{\mathbf z\mathbf z'}\|=o_p(\epsilon).
\)
    \item \textbf{Approximate Maximization.} The empirical whitened encoders
    $(\tilde{\mathbf{f}}_{\bm{\theta}}, \tilde{\mathbf{f}}'_{\bm{\theta}'})$
    are $\delta_n$-maximizers of $\hat{J}$ for some sequence $\delta_n \to 0^+$, i.e.,
    \[
    \hat{J}(\tilde{\mathbf{f}}_{\bm{\theta}}, \tilde{\mathbf{f}}'_{\bm{\theta}'})
    \;\geq\; \sup_{\tilde{\mathbf f}'\in\tilde{\mathcal F}_{\mathcal X},\tilde{\mathbf f}'\in\tilde{\mathcal F}'_{\mathcal X}} \hat{J}(\tilde{\mathbf f},\tilde{\mathbf f}') - \delta_n,
    \quad \delta_n \to 0.
    \]
\end{enumerate}
Then, as $n \to \infty$, the following convergence properties hold in probability:
\begin{enumerate}[label=\arabic*.]
    \item \textbf{Objective Consistency.}
    \[\sup_{\tilde{\mathbf{f}} \in
    \tilde{\mathcal{F}}_{\mathcal{X}}, \tilde{\mathbf{f}}' \in
    \tilde{\mathcal{F}}'_{\mathcal{X}'}} \big|\hat{J}(\tilde{\mathbf{f}},
    \tilde{\mathbf{f}}') - J(\tilde{\mathbf{f}}, \tilde{\mathbf{f}}')\big|
    \xrightarrow{\mathbb{P}} 0.
    \]
    \item \textbf{Estimator Consistency.} For any population maximizer
    $(\tilde{\mathbf{f}}^\star, \tilde{\mathbf{f}}^{\prime\star})$, there exist
    orthogonal matrices $\mathbf{Q}, \mathbf{Q}' \in O(d_{\mathcal{Z}})$ such
    that the optimal empirical encoders $(\tilde{\mathbf{f}}^*_{\bm{\theta}^*},
    \tilde{\mathbf{f}}'^*_{\bm{\theta}'^*})$ satisfy:
    \[
        \big\|\mathbf{Q}\tilde{\mathbf{f}}^*_{\bm{\theta}^*} - \tilde{\mathbf{f}}^\star\big\|_{L^2(P_{\mathbf{x}})}
        + \big\|\mathbf{Q}'\tilde{\mathbf{f}}'^*_{\bm{\theta}'^*} -
        \tilde{\mathbf{f}}^{\prime\star}\big\|_{L^2(P_{\mathbf{x}'})}
        \xrightarrow{\mathbb{P}} 0.
    \]
    \item \textbf{Latent Recovery.} By \autoref{prop:rep-inv} and
    \autoref{thm:affine-id}, the composed encoders recover the ground-truth
    latents up to orthogonal transformations:
    \begin{align*}
        \inf_{\mathbf{Q},\mathbf{Q}'\in O(d_\mathcal Z)}
        \Big(
        &\big\|\mathbf{Q}\,(\tilde{\mathbf f}^*_{\bm\theta^*}\circ \mathbf g) - \tilde{\mathbf h}^*\big\|_{L^2(P_{\mathbf s})}
        + \\
        &\big\|\mathbf{Q}'\,(\tilde{\mathbf f}'^*_{\bm\theta'^*}\circ \mathbf g') - \tilde{\mathbf h'}^*\big\|_{L^2(P_{\mathbf s'})}
        \Big)\ \xrightarrow{\mathbb{P}}\ 0,
    \end{align*}
\end{enumerate}
\end{theorem}
\begin{sketchproof}

    The complete proof, deferred to Appendix \ref{app:proof-emp-consistency}, proceeds in four steps. First, uniform convergence
of the empirical covariances establishes the asymptotic stability of the ridge-regularized
whitening operators. Second, this stability ensures the empirical normalized cross-covariance
$\hat{\mathbf{K}}$ converges uniformly to its population counterpart $\mathbf{K}$ in operator norm;
the continuity of the CCA objective then guarantees uniform convergence of the empirical
objective (\autoref{thm:emp-consistency}.1). Third, given the uniqueness of the population
maximizer up to orthogonal transformations, standard $M$-estimation arguments on the quotient
space yield estimator consistency (\autoref{thm:emp-consistency}.2). Finally, coupling this
consistency with reparameterization invariance (\autoref{prop:rep-inv}) and affine identifiability
(\autoref{thm:affine-id}) proves that the learned encoders recover the true latents up to an
orthogonal ambiguity (\autoref{thm:emp-consistency}.3).
\end{sketchproof}
An interpretation of Assumptions A3–A4 and their role in the consistency argument is provided in \autoref{app:remarks}.

\section{EXPERIMENTS}\label{sec:experiments}
\begin{table*}
    \centering
    \resizebox{\linewidth}{!}{
    \setlength{\tabcolsep}{3pt}
    \large
    \begin{tabular}{l*{10}{c}}
        Methods   & \multicolumn{2}{c}{Gaussian}  & \multicolumn{2}{c}{Negative Binomial}  & \multicolumn{2}{c}{Gamma}    & \multicolumn{2}{c}{Poisson}    & \multicolumn{2}{c}{Hypergeometric} \\
        \cmidrule(r){2-3}
        \cmidrule(r){4-5}
        \cmidrule(r){6-7}
        \cmidrule(r){8-9}
        \cmidrule(r){10-11}
            & $\mathbf f$ & $\mathbf f'$   & $\mathbf f$ & $\mathbf f'$   &$\mathbf f$ & $\mathbf f'$   &$\mathbf f$ & $\mathbf f'$   &$\mathbf f$ & $\mathbf f'$  \\
        \midrule
        SwAV & 32.96 $\pm$4.05 & 37.88 $\pm$0.45 & 33.17 $\pm$3.67 & 28.85 $\pm$2.62 & 29.37 $\pm$4.85 & 28.38 $\pm$2.74 & 36.43 $\pm$0.88 & 35.23 $\pm$1.69 & 12.88 $\pm$8.25 & 12.84 $\pm$8.66\\
        BarlowTwins    & 79.44$\pm$0.26 &79.51 $\pm$0.3             & 67.22 $\pm$5.67 & 68.33 $\pm$4.11 & 61.99 $\pm$2.4 & 61.98 $\pm$2.43 & 55.77 $\pm$5.82 & 55.29 $\pm$4.59 &25.47 $\pm$0.49 &23.9 $\pm$1.53\\
        VICReg  &  14.54 $\pm$1.14 & 13.05 $\pm$1.1 & 13.83 $\pm$1.82 & 12.82 $\pm$1.08 & 12.27 $\pm$0.56 & 13.61 $\pm$0.82 & 11.88 $\pm$0.48 & 11.37 $\pm$0.82 & 18.94 $\pm$2.37 & 16.7 $\pm$0.64 \\
        W-MSE & 99.17 $\pm$0.21 & 99.4 $\pm$0.04 & \textbf{99.07 $\pm$0.04} & 99.23 $\pm$0.04 & 99.11 $\pm$0.15 & 99.34 $\pm$0.05 & \textbf{99.18 $\pm$0.08} & \textbf{99.4 $\pm$0.04} & \textbf{98.97 $\pm$0.04} & \textbf{99.02 $\pm$0.32} \\
        DGCCA & 21.9 $\pm$0.79 & 18.17 $\pm$1.4 & 21.45 $\pm$1.52 & 23.9 $\pm$0.86 & 22.0 $\pm$0.73 & 21.17 $\pm$1.65 & 22.16 $\pm$1.43 & 18.94 $\pm$1.2 & 23.81 $\pm$1.09 & 23.95 $\pm$1.25\\
        rDGCCA & 90.57 $\pm$2.89 & 82.48 $\pm$3.7 & 90.62 $\pm$3.33 & 78.51 $\pm$2.73 & 90.35 $\pm$3.23 & 83.35 $\pm$4.16 & 92.97 $\pm$2.26 & 83.33 $\pm$1.99 & 91.74 $\pm$3.04 & 86.93 $\pm$2.89\\
        DeepCCA   & \textbf{99.4 $\pm$0.05} & \textbf{99.41 $\pm$0.03} &  99.05 $\pm$0.04 & \textbf{99.23 $\pm$0.03}  & \textbf{99.12 $\pm$0.14} & \textbf{99.35 $\pm$0.03} & 99.18 $\pm$0.09& 99.36 $\pm$0.03 &98.91 $\pm$0.15&98.37 $\pm$0.1\\
        \bottomrule
    \end{tabular}}
        \caption{Comparison of the  coefficient of determination $R^2$$\uparrow$(\%) of both encoders $\mathbf f, \mathbf f'$ on synthetic data ($d_\mathcal{S}=d_\mathcal{Z}=10$).}
        \label{tab:synthetic_identifiability}
\end{table*}
In this section, we empirically validate our theoretical guarantees for
nonlinear CCA. We evaluate affine identifiability across diverse latent
distributions, verify reparameterization invariance and finite-sample
consistency, and systematically ablate our core assumptions.
We adopt the experimental setup from previous works
\citep{zimmermann2021contrastive, matthes2023towards} on disentangled
representations and extend it to a multi-modal setting. The code is available at
\url{https://github.com/ZhiweiHan9277/AISTATS2026_Provable_Affine_Identifiability}.
\subsection{Experimental Setup}
\paragraph{Datasets.} We evaluate our theoretical claims on a synthetic
dataset and the 3DIdent image dataset \citep{zimmermann2021contrastive}.
For the synthetic setup, we adapt the online sampling mechanism
from prior work \citep{zimmermann2021contrastive,matthes2023towards},
generalizing their single-generator framework to two independent
generators to simulate multi-modal observations. 3DIdent comprises
images of a 3D object rendered from 11 latent
factors, e.g., position, rotation and illumination,
yielding (image, latent) tuples.

To map our continuous generative process to the finite 3DIdent dataset,
we employ a nearest-neighbor matching strategy. We first independently sample
$\mathbf{a}, \mathbf{b}, \mathbf{c}$ to construct continuous latent
pairs $(\mathbf{s}, \mathbf{s}')$ as in \autoref{equ:additive}.
We then retrieve the corresponding rendered images $(\mathbf{x}, \mathbf{x}')$
by finding the Euclidean nearest neighbors of $\mathbf{s}$ and $\mathbf{s}'$
in the 3DIdent ground-truth latent space. To emulate distinct multi-view
modalities, we retain the original color image for $\mathbf{x}$ and use a
grayscale conversion for $\mathbf{x}'$, preserving shared generative factors
while marginalizing out color.

\paragraph{Baselines.}
We benchmark the CCA-like objectives and strong non-contrastive SSL baselines to test affine identifiability under a unified encoder architecture and a family of distributions aligned with our theory.
\begin{itemize}
\item \textbf{SwAV} \citep{caron2020unsupervised} learns view-invariant features via online clustering with prototype assignments.
\item \textbf{BarlowTwins} \citep{zbontar2021barlow} enforces invariance and minimizes redundancy through cross-correlation diagonalization.
\item \textbf{VICReg} \citep{bardes2021vicreg} encourages invariance while regularizing variance and covariance to decorrelate neural representations.
\item \textbf{W-MSE} \citep{ermolov2021whitening} aligns whitened multi-view neural representations by minimizing mean-squared error.
\item \textbf{DGCCA} \citep{benton2017deep} is a nonlinear extension of generalized CCA, optimizing a shared latent that reconstructs all view projections.
\item \textbf{rDGCCA} \citep{karakasis2023revisiting} improves DGCCA by extracting shared representations under conditionally independent private latents, offering theoretical recovery guarantees.
\item \textbf{DeepCCA} \citep{andrew2013deep} maximizes the CCA objective with whitened multi-view representations.
\end{itemize}
We exclude Kernel CCA due to poor high-dimensional scalability.
Among the evaluated baselines, only DeepCCA faithfully enforces exact whitening and maximizes canonical correlations. We therefore use its converged solution as a proxy for population CCA with the data  generated from online sampling of the underlying distributions.
All baselines are implemented using their official repositories when
available, otherwise we reimplement by following their paper specifications.

\paragraph{Metrics.}
Following prior work \citep{hyvarinen2016unsupervised,zimmermann2021contrastive,matthes2023towards}, we assess affine recoverability by regressing ground-truth
latents onto learned latents and reporting the coefficient of determination
($R^2$) averaged across dimensions. For settings with partial distributional
violations or dimensional mismatch, we evaluate subspace error using the mean
($PA_{\text{mean}}$) and maximum ($PA_{\text{max}}$) principal angles between
the learned span and the ground-truth canonical subspace, in line with
subspace identification studies \citep{ma2020subspace,gao2017sparse,cai2018rate}.
To quantify the alignment between two encoders up to
orthogonal transformations,
we measure the orbit distance defined as
\(
\min_{\mathbf{Q},\mathbf{Q}'\in O(d_\mathcal Z)} \; \tfrac{1}{n}\Big( \|\hat{\mathbf{Z}}-\mathbf{Z} \mathbf{Q}\|_F^2 \;+\; \|\hat{\mathbf{Z}}'-\mathbf{Z}' \mathbf{Q}'\|_F^2 \Big),
\)
where $n$ is the batch size, $\mathbf{Z}, \mathbf{Z}'$ and
$\hat{\mathbf{Z}}, \hat{\mathbf{Z}}'$ are the learned latents
of two sets of encoders.

% To enforce the function class constraints of \autoref{assump:function-class}, we explicitly whiten the learned latent representations.

\subsection{Validation of Theoretical Findings}
\paragraph{Affine Identifiability.}
For the synthetic experiments ($d_{\mathcal S}=10$), latent pairs
$(\mathbf{s}, \mathbf{s}')$ are generated via \autoref{equ:additive} by
sampling $\mathbf{a}, \mathbf{b}$, and $\mathbf{c}$ independently from
five candidate distributions, and observations $(\mathbf{x}, \mathbf{x}')$
are produced by independent decoders $\mathbf{g}$ and $\mathbf{g}'$.
The $R^2$ scores in \autoref{tab:synthetic_identifiability} support the
predicted affine identifiability under this additive model: DeepCCA and
W-MSE consistently achieve near-perfect recovery across all distributions.
rDGCCA forms a weaker second tier with noticeable asymmetry across the
two views, while Barlow Twins deteriorates substantially on the
hypergeometric setting. SwAV, VICReg, and DGCCA remain far from exact
recovery. Detailed distribution settings and subspace recovery errors are
deferred to \autoref{sec:results}.

On 3DIdent (\autoref{tab:3dident_identifiability}), DeepCCA again
performs best, with W-MSE close behind in both $R^2$ and principal-angle
errors. rDGCCA and Barlow Twins provide partial recovery but remain
clearly below these two methods, whereas DGCCA fails outright and SwAV
and VICReg also perform poorly. Because 3DIdent is constructed by
nearest-neighbor matching in latent space rather than from exact
generative pairs, exact recovery is inherently harder. We therefore also
report subspace recovery errors as a more robust measure of
representation alignment under finite-sample and matching noise.

\begin{table}
    \centering
    \resizebox{\linewidth}{!}{
    \setlength{\tabcolsep}{2pt}
    \begin{tabular}{l*{6}{c}}
        \toprule
        Methods   & \multicolumn{2}{c}{$R^2\uparrow$(\%)}  & \multicolumn{2}{c}{$PA_{\text{mean}}\downarrow$(°)}  & \multicolumn{2}{c}{$PA_{\max}\downarrow$(°)} \\
        \cmidrule(r){2-3}
        \cmidrule(r){4-5}
        \cmidrule(r){6-7}
            & $\mathbf f$ & $\mathbf f'$   & $\mathbf f$ & $\mathbf f'$   &$\mathbf f$ & $\mathbf f'$ \\
        \midrule
        SwAV & 22.45$\pm$5.59&26.21$\pm$5.99&67.22$\pm$3.64&65.94$\pm$4.26&89.28$\pm$0.58&89.28$\pm$0.41\\
        BarlowTwins & 85.20$\pm$0.38 & 86.07$\pm$0.48&17.39$\pm$0.61&16.76$\pm$0.67&73.69$\pm$3.86&70.17$\pm$4.29\\
        VICReg  & 22.34$\pm$1.37&24.45$\pm$1.39&66.13$\pm$1.46&65.40$\pm$1.31&89.07$\pm$0.75&89.74$\pm$0.95\\
        W-MSE & 97.40$\pm$0.80 & 97.00$\pm$0.87 & 8.82$\pm$0.31&9.65$\pm$1.06&13.98$\pm$1.97&13.00$\pm$2.01\\
        DGCCA & 0.37$\pm$0.77&0.47$\pm$1.36&87.70$\pm$1.21&86.72$\pm$1.73&89.97$\pm$0.21&89.42$\pm$0.28\\
        rDGCCA & 89.13$\pm$0.89 & 79.98$\pm$0.70 & 11.39$\pm$0.13 & 19.51$\pm$0.33 & 57.78$\pm$5.61 & 82.55$\pm$7.12 \\
        DeepCCA  & \textbf{97.80$\pm$0.23} & \textbf{97.79$\pm$0.35}&\textbf{8.33$\pm$0.46}&\textbf{8.33$\pm$0.54}&\textbf{11.11$\pm$0.91}&\textbf{10.76$\pm$1.04}\\
        \bottomrule
    \end{tabular}}
        \caption{Comparison of $R^2$, the mean principal angle $PA_{\text{mean}}$, and the maximal principal angle $PA_{\max}$ of both encoders on 3DIdent ($d_{\mathcal S}=d_{\mathcal Z}=7$).}
        \label{tab:3dident_identifiability}
\end{table}
\begin{figure}[t]
  \centering
  \begin{subfigure}[b]{0.23\linewidth}
    \centering
    \includegraphics[width=\linewidth]{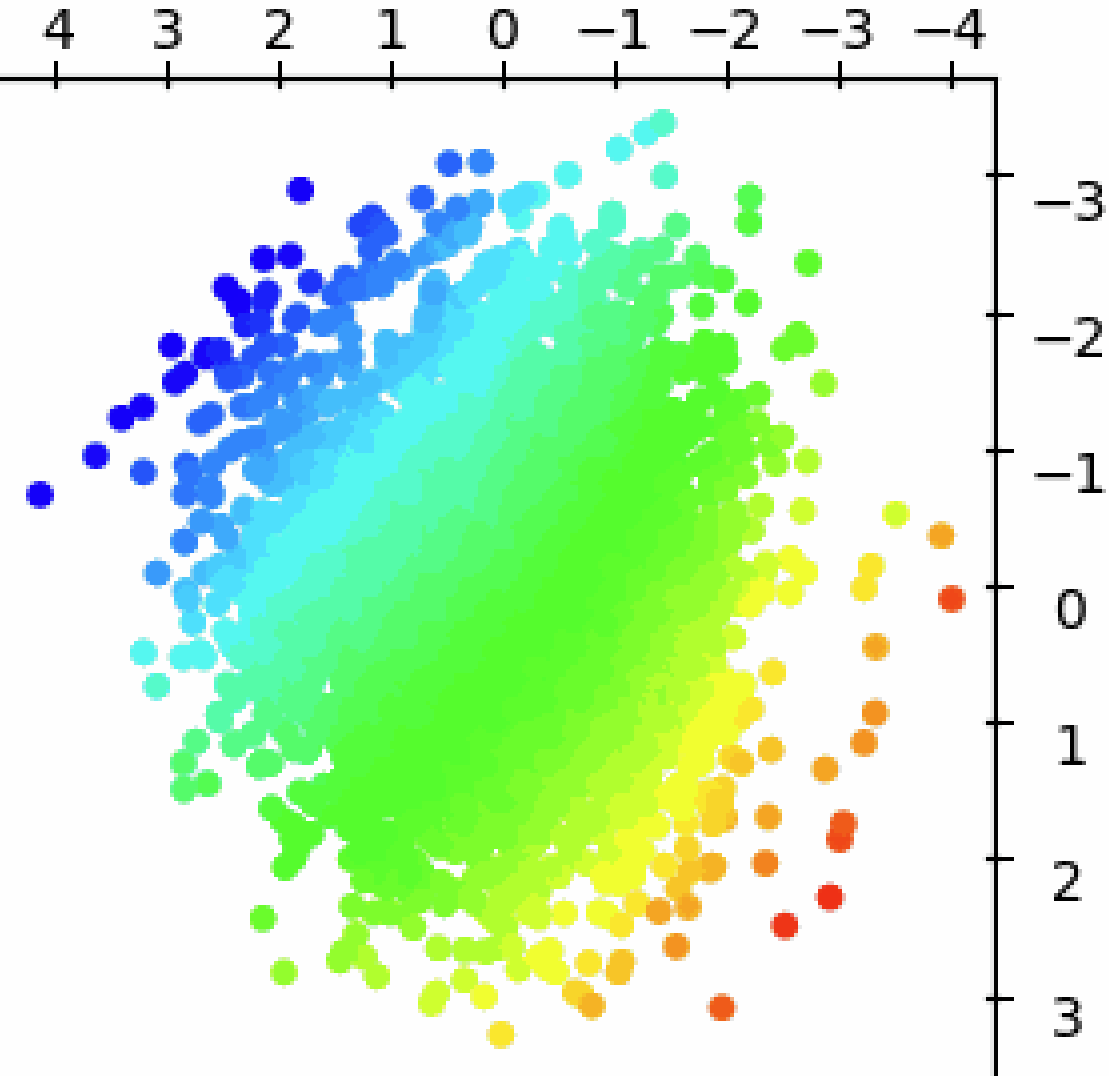}
  \end{subfigure}
  \hfill
  \begin{subfigure}[b]{0.23\linewidth}
    \centering
    \includegraphics[width=\linewidth]{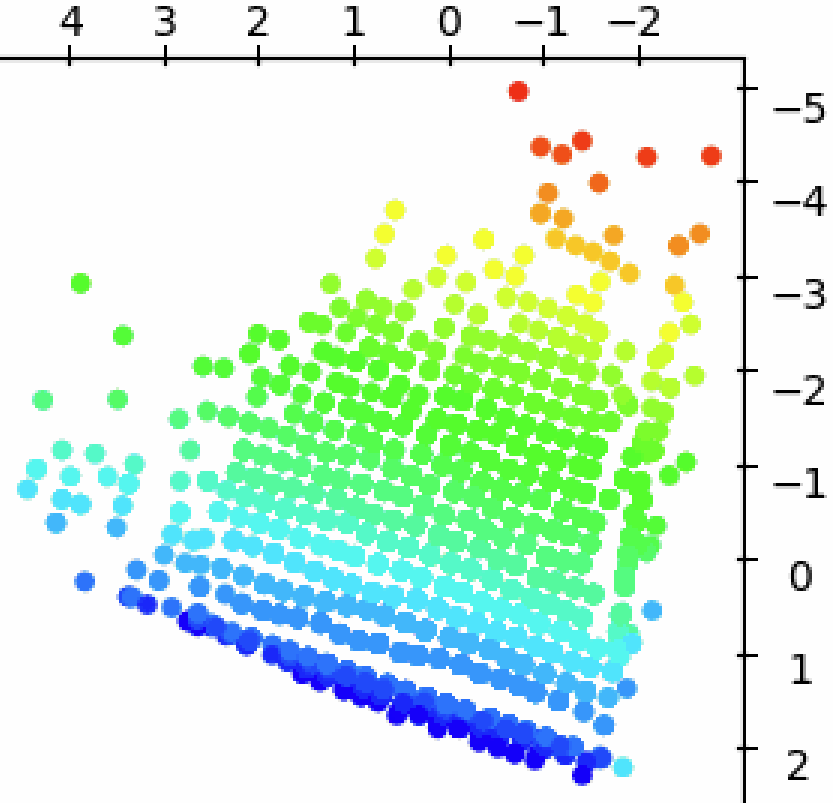}
  \end{subfigure}
  \hfill
  \begin{subfigure}[b]{0.23\linewidth}
    \centering
    \includegraphics[width=\linewidth]{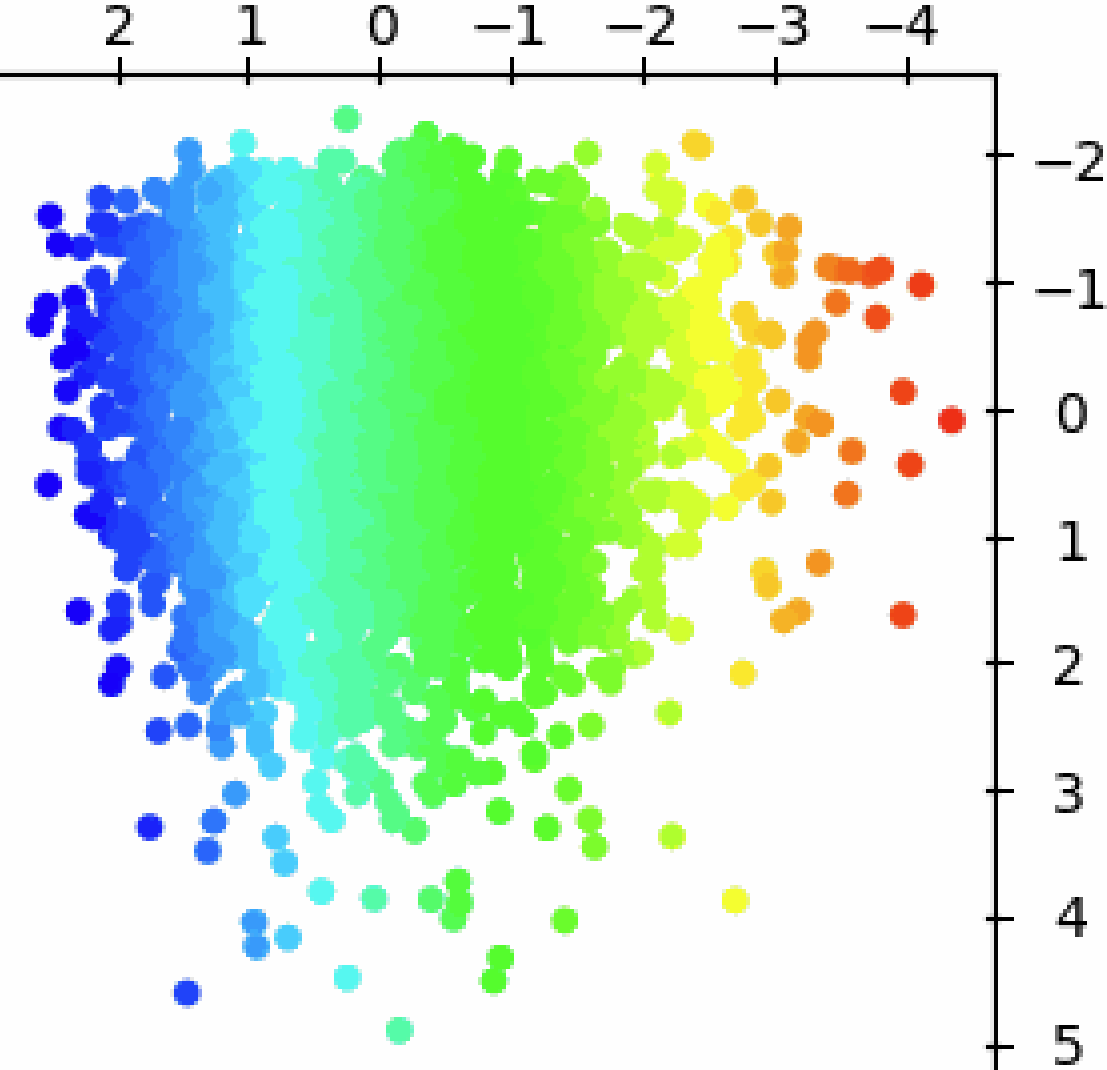}
  \end{subfigure}
  \hfill
  \begin{subfigure}[b]{0.23\linewidth}
    \centering
    \includegraphics[width=\linewidth]{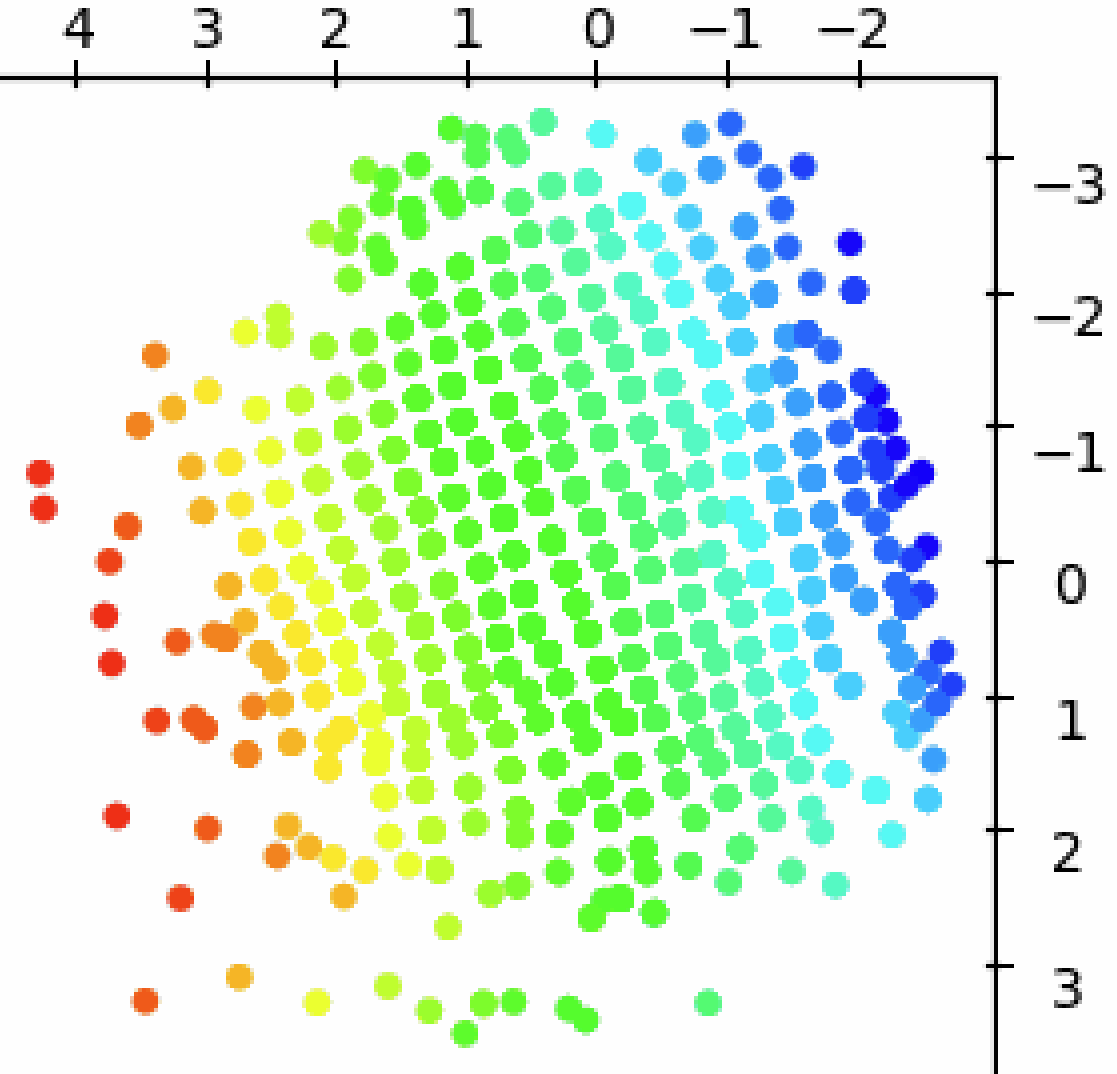}
  \end{subfigure}

  \vskip\baselineskip
  \begin{subfigure}[b]{0.23\linewidth}
    \centering
    \includegraphics[width=\linewidth]{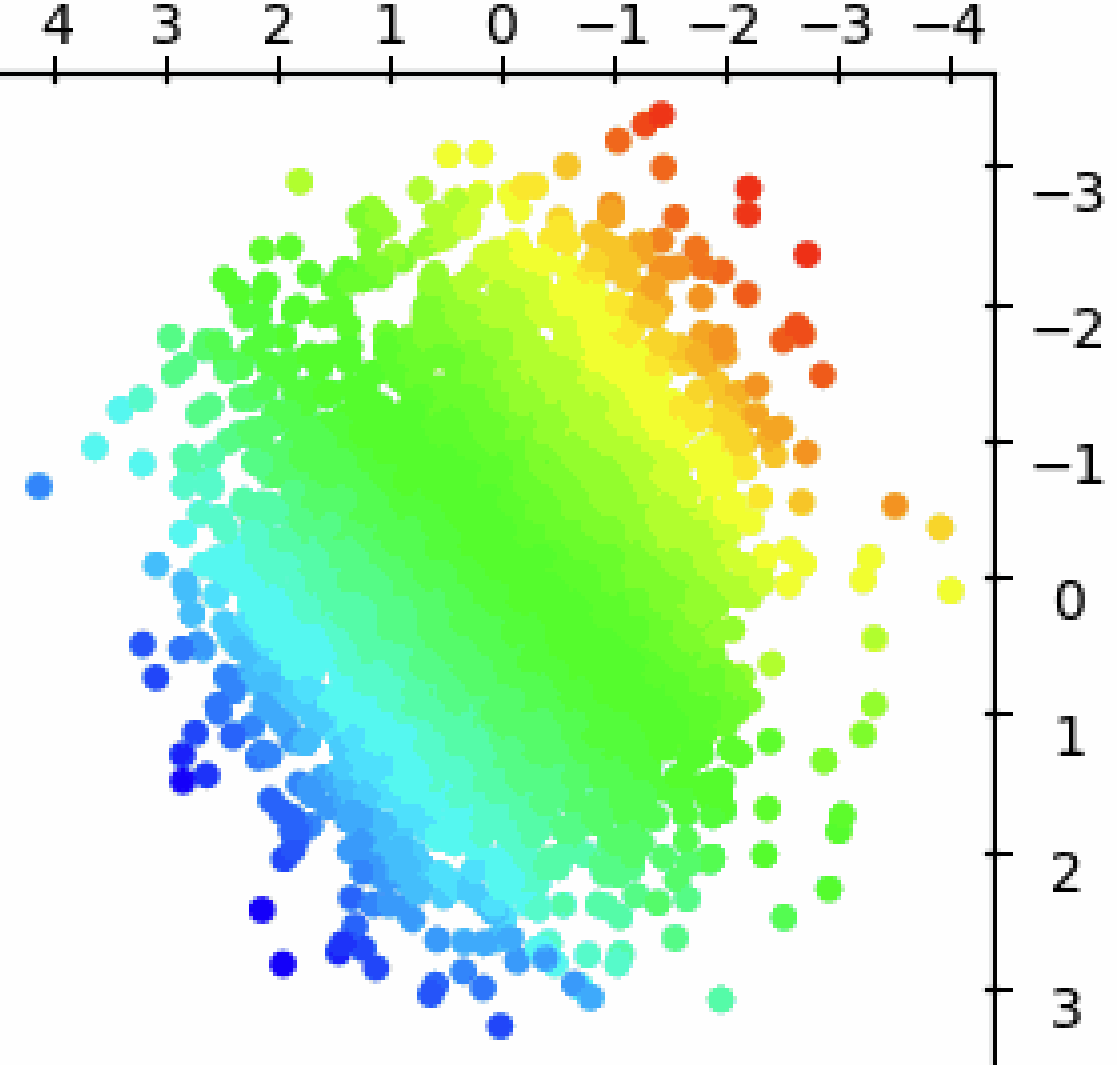}
    \caption{Gaussian}
  \end{subfigure}
  \hfill
  \begin{subfigure}[b]{0.23\linewidth}
    \centering
    \includegraphics[width=\linewidth]{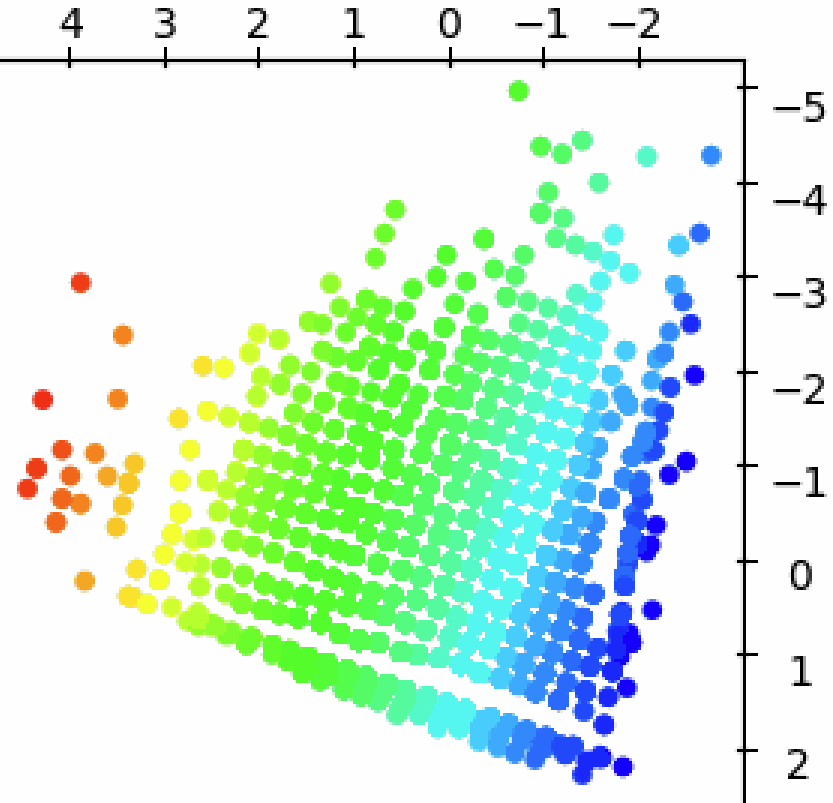}
    \caption{Neg. Bi.}
  \end{subfigure}
  \hfill
  \begin{subfigure}[b]{0.23\linewidth}
    \centering
    \includegraphics[width=\linewidth]{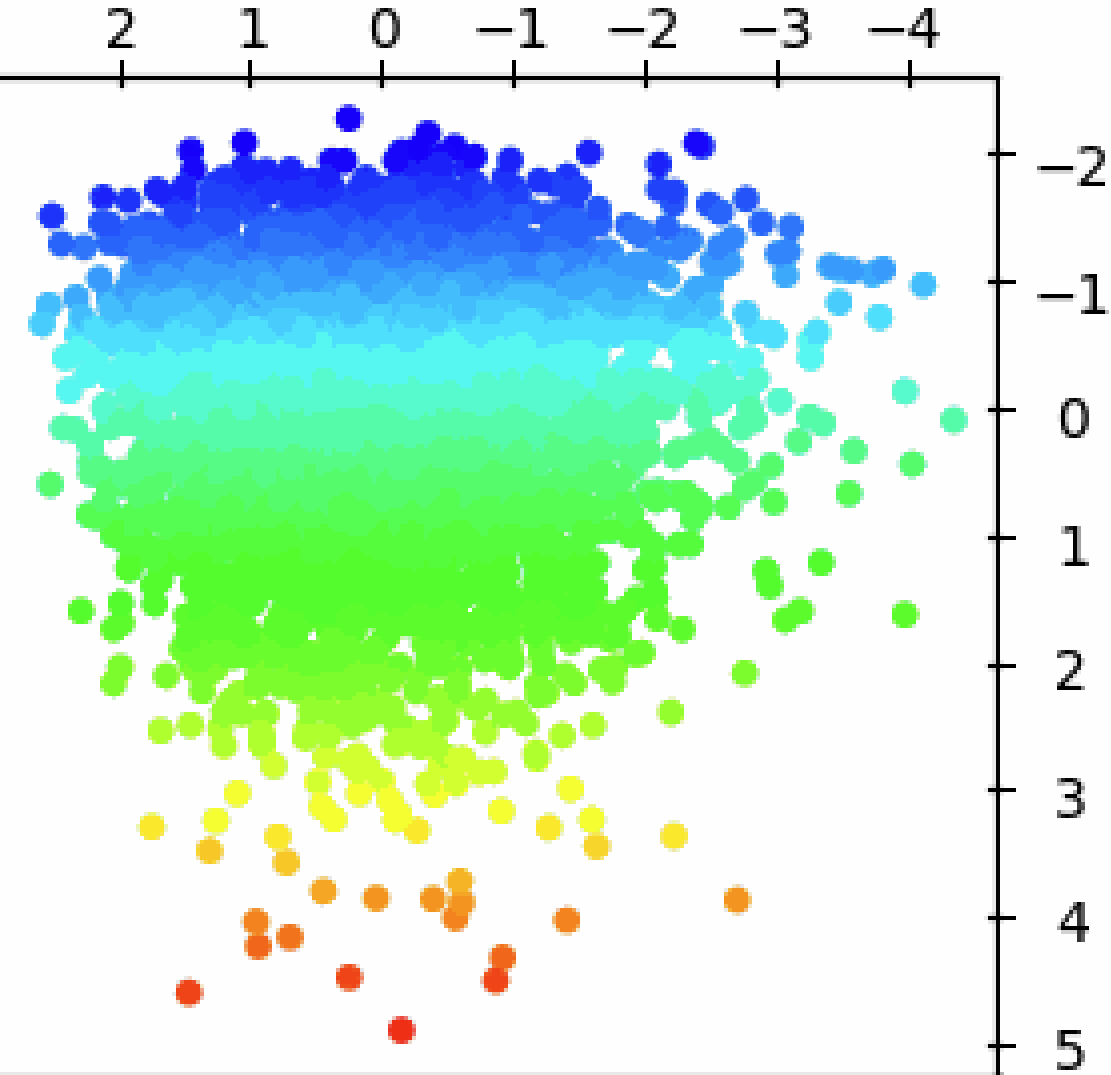}
    \caption{Gamma}
  \end{subfigure}
  \hfill
  \begin{subfigure}[b]{0.23\linewidth}
    \centering
    \includegraphics[width=\linewidth]{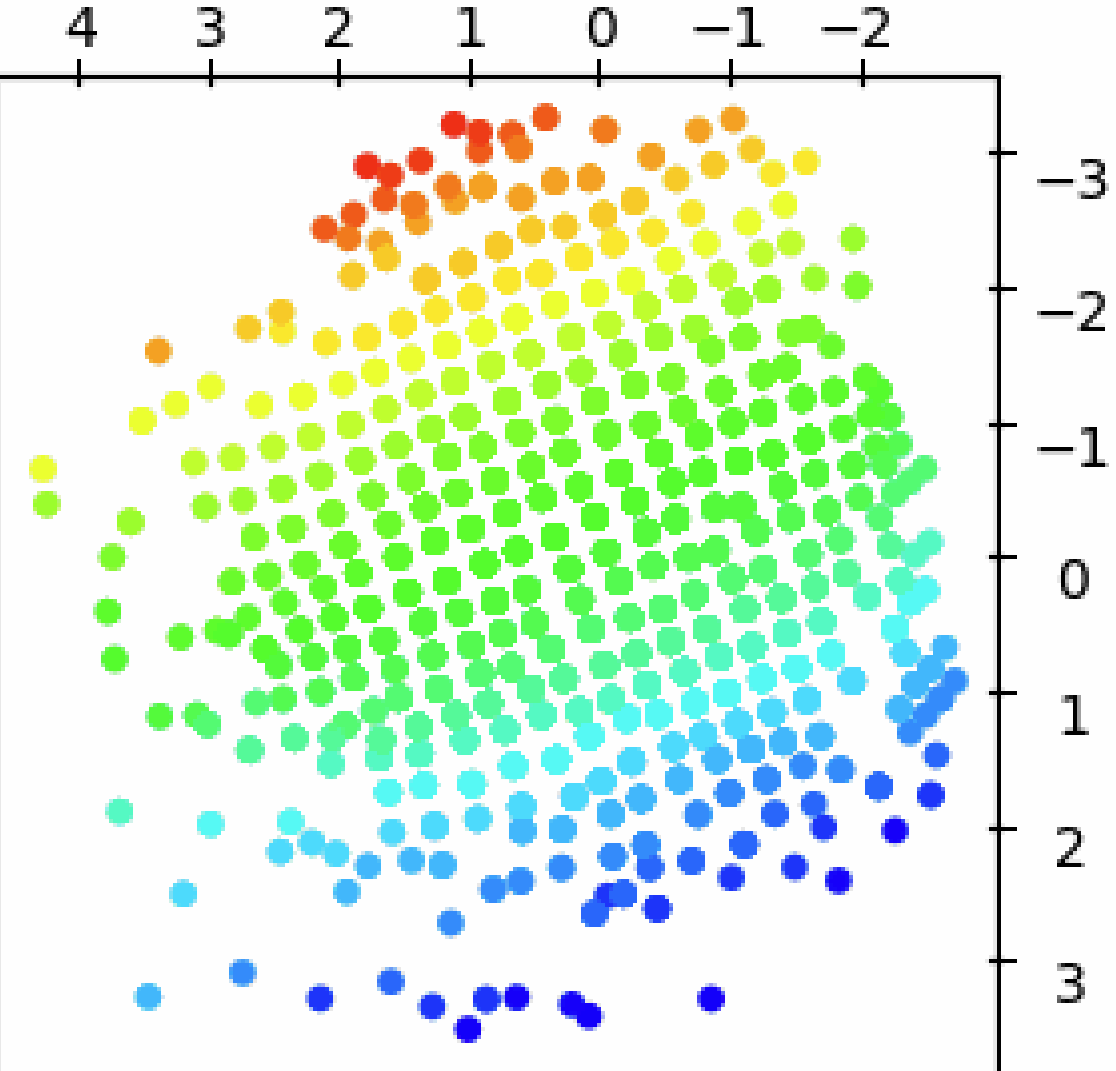}
    \caption{Poisson}
  \end{subfigure}

  \caption{Whitened latent learned representations by DeepCCA on synthetic data ($d_\mathcal{S}=d_\mathcal{Z}=2$). Color gradients in different rows illustrate the variations along a single coordinate in $\mathcal{S}$.}
  \label{fig:color_gradient}
\end{figure}

\autoref{fig:color_gradient} visualizes the 2D whitened
representations learned by DeepCCA across four distributions,
omitting Hypergeometric due to representation collapse.
The source coordinates manifest as smoothly varying,
mutually orthogonal color gradients. As theoretically expected, these
gradients are rotated relative to the canonical axes, empirically
validating that whitening identifies the true subspace only up to an
orthogonal transformation.
\begin{figure}[t]
    \centering
    \includegraphics[width=\linewidth]{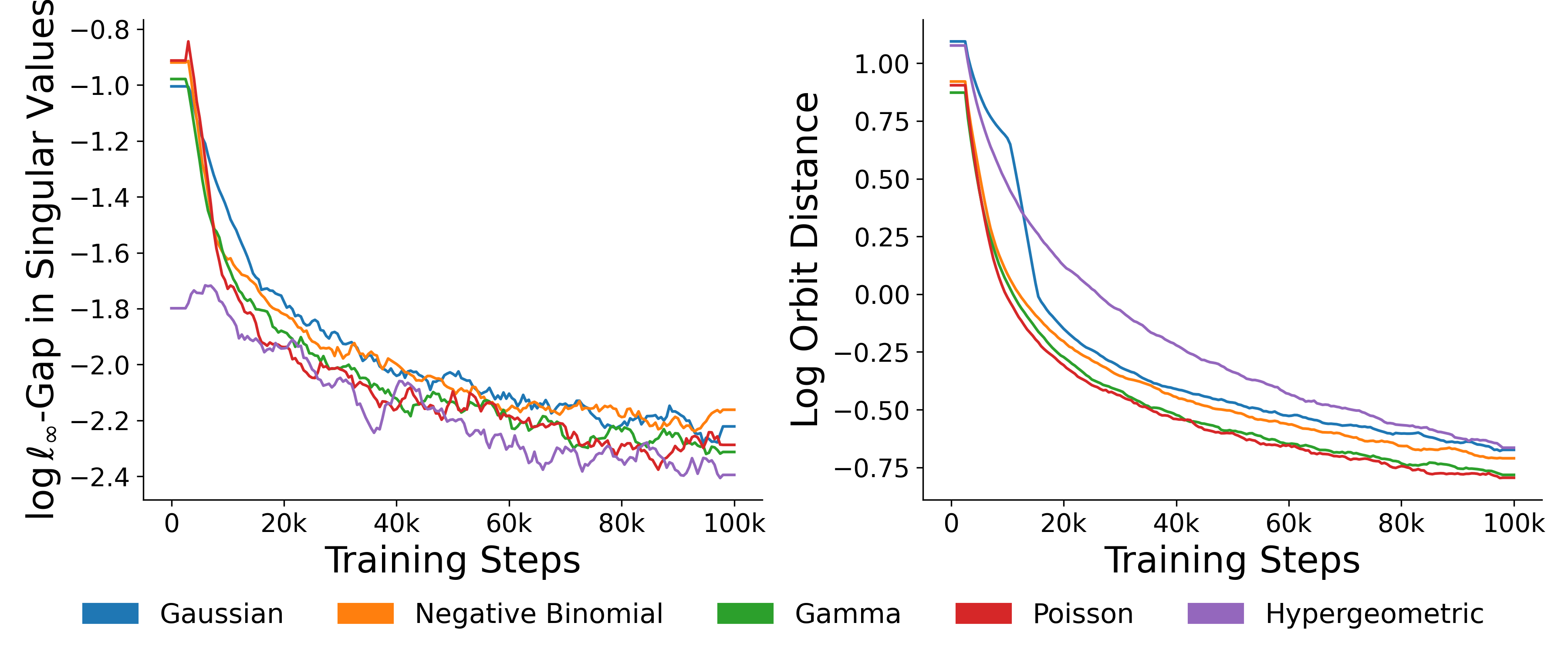}

\caption{The $\ell_\infty$-gap in singular values and log orbit distance over
training steps in Gaussian case.}
\label{fig:reparam_inv}
\end{figure}
\paragraph{Reparameterization Invariance.}
This experiment is not a downstream-performance benchmark and therefore
does not admit a meaningful chance-level baseline. The relevant
reference point is exact agreement, i.e., zero singular-value gap and zero
orbit distance before the logarithmic transform.
Our goal is only to test Proposition~1, namely whether CCA trained in observation
space and in source space converges to the same $O(d_{\mathcal Z})$-orbit and
canonical spectrum. To this end, we independently optimize two CCA encoders under
identical synthetic configurations: one operating on the observation space and
another directly on the latent source space.
\autoref{fig:reparam_inv} tracks the $l_\infty$-norm
singular-value gap and the orbit distance between the two learned encoders
for a given view.
Smaller values signify closer alignment, exact invariance implies a zero
singular value gap and zero orbit distance prior to logarithmic transformation.
Across five random seeds, both metrics rapidly converge and remain consistently
small throughout training. The median orbit distance remains below 0.2,
while the $l_\infty$ singular-value gap stays below $0.01$.
Crucially, this demonstrates that the
observation-space maximizer of the CCA objective tightly aligns with its
source-space counterpart, providing strong empirical corroboration for
\autoref{prop:rep-inv}.
\begin{figure}[t]
    \centering
    \includegraphics[width=\linewidth]{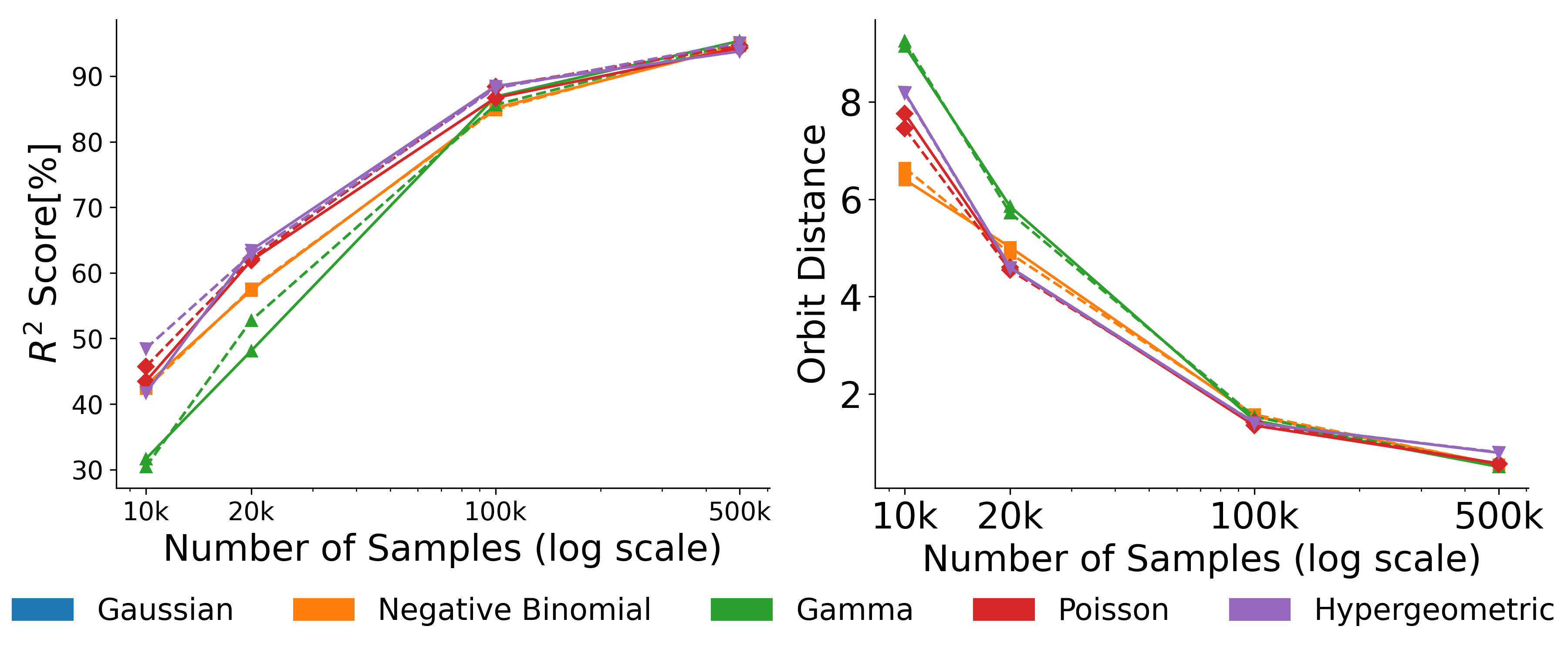}

\caption{$R^2\uparrow$ and Orbit distance over changing sample size in fixed dataset setup.}
\label{fig:consistency}
\end{figure}
\paragraph{Finite-sample Consistency.}
We evaluated the finite-sample consistency of CCA by varying the
sample size $n$
while keeping the remaining synthetic setup fixed,
utilizing $\epsilon=0.01\cdot n^{-1/4}$ balancing bias and stability.
We fix the source dimension to $d_{\mathcal S}=10$ and report
both the $R^2$ score and the orbit distance between the empirical
and population CCA maximizers for a single view in
\autoref{fig:consistency}.
As the sample size increases, the $R^2$ improves steadily across all distributions, exceeding 90\% at roughly $4\cdot 10^5$ samples,
while orbit distance rapidly decays to near zero, indicating
convergence of the learned representation to the population solution
up to an orthogonal transformation. These results demonstrate that
empirical CCA recovers the affine structure of the latent sources
as the sample size grows as stated in \autoref{thm:emp-consistency}.
\subsection{Ablation Study}
\begin{figure}[t]
  \centering
  \begin{subfigure}[b]{0.49\linewidth}
    \centering
    \includegraphics[width=\linewidth]{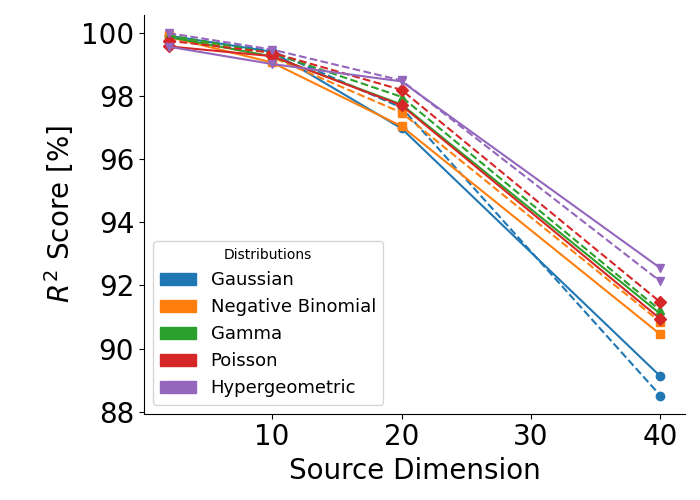}
  \end{subfigure}
  \hfill
  \begin{subfigure}[b]{0.49\linewidth}
    \centering
    \includegraphics[width=\linewidth]{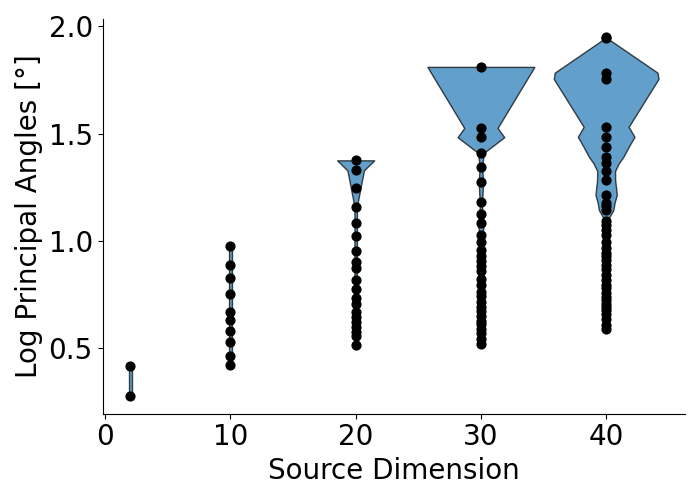}
  \end{subfigure}
\caption{Ablation over the source dimension $d_\mathcal S$ ($d_\mathcal S =d_\mathcal Z$). Left: $R^2\uparrow$. Solid lines denote encoder $\mathbf{f}$ and dashed lines encoder $\mathbf{f}'$. Right: log principal angles of $\mathbf{f}$ in the Gaussian case. Black dots denote principal angles and shaded region indicates the log-standard deviation.}
        \label{fig:source_dimension}
\end{figure}

We provide an ablation study to isolate the effects of source dimensionality, first-order canonical dominance and latent–dimension mismatch on affine
identifiability. Following the synthetic experimental setup, we vary
only one parameter at a time while keeping all others fixed.
\paragraph{Source Dimension.}
CCA consistently achieves affine identifiability in low-dimensional
settings ($d_\mathcal{S}=2,10,20$), as evidenced by small principal angles
between the recovered and ground-truth subspaces and high $R^2$ scores,
as shown in the left panel of \autoref{fig:source_dimension}. As the source
dimension increases ($d_{\mathcal S}=30,40$), CCA continues to recover a
substantial portion of the correlated subspace, but the recovery quality
gradually deteriorates. Both the mean and variance of the principal angles
increase, indicating reduced estimation stability in higher dimensions, as
illustrated in the right panel of \autoref{fig:source_dimension}.
\begin{figure}[t]
  \centering
    \includegraphics[width=\linewidth]{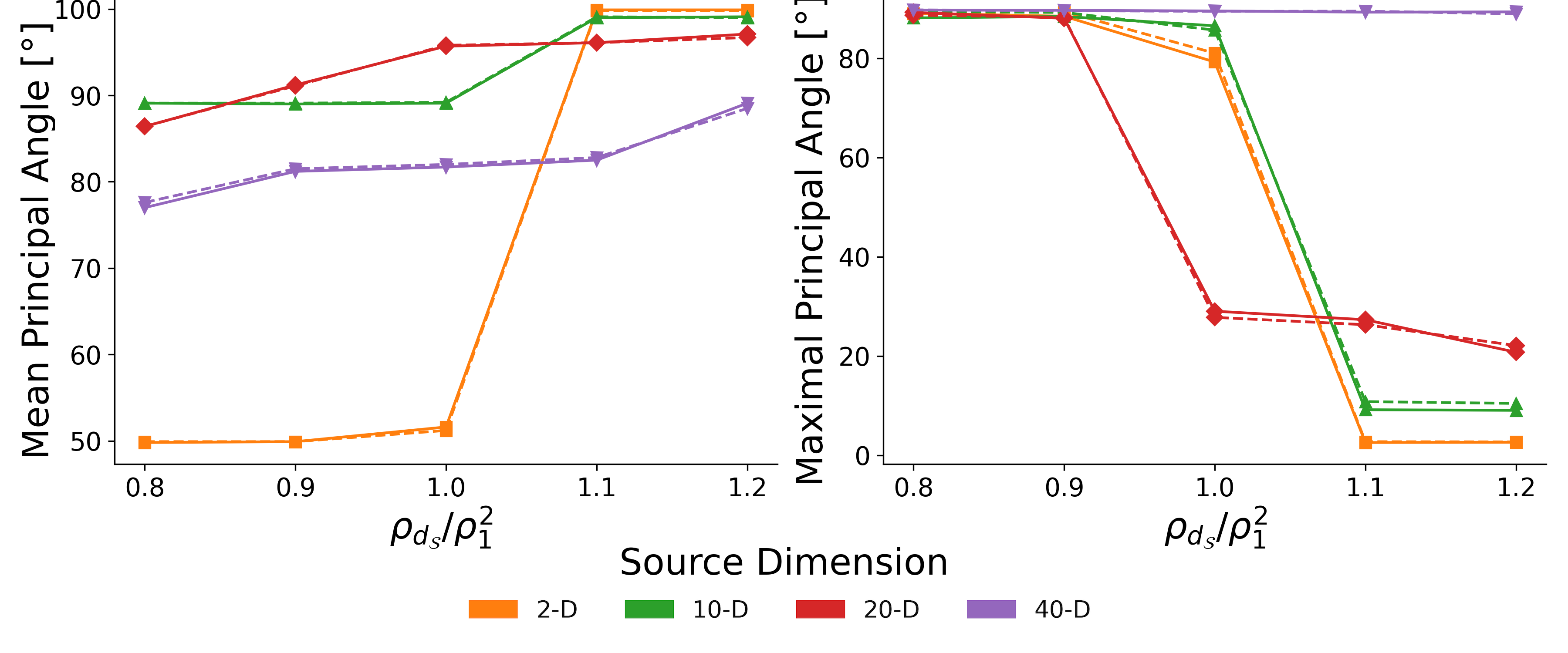}
\caption{Ablation over the first order canonical ratio $\rho_{d_{\mathcal S}}/\rho_1^2$ ($d_\mathcal S =d_\mathcal Z$). Left: $R^2\uparrow$. Right: $PA_{max}\downarrow$. Colors denote different source dimensions.}
\label{fig:first_order_canonical_dominance}
\end{figure}
\paragraph{First-order Canonical Dominance.}
We introduce the \textit{first-order canonical dominance ratio},
$\rho_{d_{\mathcal S}} / \rho_1^2$, to characterize the weakest
canonical correlation's strength relative to the squared strongest
correlation. Assuming source canonical correlations are uniformly
sampled from $[\rho_{d_{\mathcal S}}, \rho_1]$,
\autoref{fig:first_order_canonical_dominance} illustrates the impact
of this ratio on affine identifiability. Increasing the dominance
ratio consistently improves $R^2$ scores across latent dimensions,
indicating enhanced source recoverability. Notably, identifiability
exhibits a threshold effect at $1$. For $\rho_{d_{\mathcal S}} / \rho_1^2 < 1$,
elevated maximum principal angles ($PA_{\max}$) reveal that at least
one canonical direction remains poorly identifiable. Once the ratio
exceeds $1$, $PA_{\max}$ rapidly converges to zero, confirming full
recovery of all canonical directions and strict alignment with the
ground-truth source subspace.
\begin{table}[h]
    \centering
    \resizebox{\linewidth}{!}{
    \begin{tabular}{c*{5}{c}}
        \toprule
        Encoder &
        Gaussian & Negative Binomial & Gamma & Poisson & Hypergeometric \\
        \midrule
        $\mathbf f$   & 98.93$\pm$0.04 & 98.9$\pm$0.03 & 99.09$\pm$0.03 & 99.12$\pm$0.03 & 98.56$\pm$0.05 \\
        $\mathbf f'$  & 98.94$\pm$0.05 & 99.17$\pm$0.03 & 99.33$\pm$0.02 & 99.31$\pm$0.017 & 98.56$\pm$0.03 \\
        \bottomrule
    \end{tabular}}
    \caption{$R^2\uparrow$ [\%] for both encoders $\mathbf f$ and $\mathbf f'$ in the overcomplete setup ($d_{\mathcal S}=10,\ d_{\mathcal Z}=13$).}
    \label{tab:r_square_dimension_mismatch}
\end{table}
\begin{figure}[t]
  \centering

    \includegraphics[width=\linewidth]{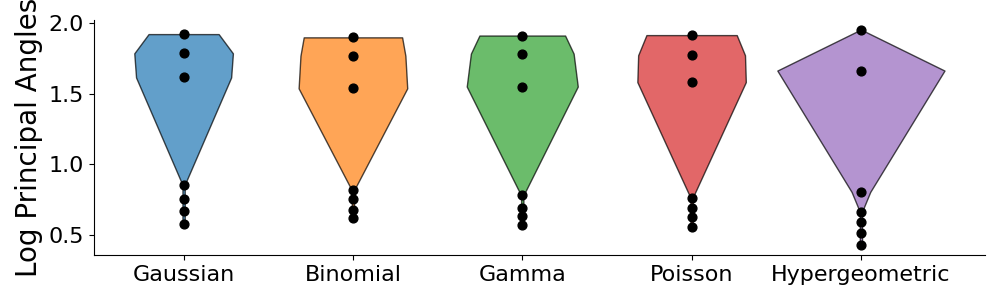}
\caption{Log principal angles of encoder $\mathbf{f}$ in the undercomplete setup ($d_{\mathcal S}=10,\ d_{\mathcal Z}=7$). Black dots denote principal angles and the shaded region indicates the log-standard deviation. }
  \label{fig:log_principle_angle_dimension_mismatch}
\end{figure}
\paragraph{Dimension Mismatch.}
We investigate both under- and overcomplete learning regimes.
In the overcomplete case ($d_{\mathcal S} < d_{\mathcal Z}$),
consistently high $R^2$ scores (\autoref{tab:r_square_dimension_mismatch})
demonstrate that the true source subspace is recoverable despite
redundant latent coordinates, though the stability of these redundancies
remains unclear.
Conversely, in the undercomplete regime ($d_{\mathcal S} > d_{\mathcal Z}$),
insufficient capacity precludes full $d_{\mathcal S}$-dimensional recovery.
Although four canonical directions are recovered with minimal principal
angle deviation (\autoref{fig:log_principle_angle_dimension_mismatch}),
affine identifiability cannot be broadly guaranteed.

\paragraph{Discussion on Whitening.}
Whitening, enforcing unit variance and decorrelation across latent
dimensions for self-supervised learning w
\citep{hua2021feature, weng2022investigation, weng2023modulate},
is critical for affine identifiability as it ensures
well-conditioned representations and stabilizes canonical direction
estimation. Empirically, methods explicitly enforcing whitening,
e.g., DeepCCA and W-MSE, achieve superior affine identifiability.
Furthermore, even relaxed decorrelation constraints, as in Barlow Twins,
yield substantial improvements over unregularized baselines. These
empirical results corroborate the theoretical analysis detailed
in \autoref{app:whitening}.
\section{CONCLUSION}
We characterized the sufficient conditions under which nonlinear CCA achieves
affine identifiability, advancing the theoretical understanding of
correlation-based non-contrastive learning.
Crucially, our framework provides a principled
explanation for the recent remarkable empirical performance of non-contrastive
methods: it proves that explicit whitening and unit-variance regularizers,
e.g., in Barlow Twins, natively drive the recovery of ground-truth
latent factors under (near-)Gaussian priors.
Although the affine guarantee is naturally weaker than
the permutation-level identifiability of contrastive
methods, CCA provides practical advantages by
circumventing negative sampling and exhibiting robustness
to small batch sizes. Finally, our proof of finite-sample
consistency guarantees that scaling dataset size directly
improves empirical source recovery and downstream disentanglement.

\bibliographystyle{apalike}
\bibliography{reference}

\clearpage

\onecolumn
\appendix
\clearpage
\thispagestyle{empty}
\aistatstitle{Supplementary Material}
\section{OVERVIEW OF THE SUPPLEMENTARY MATERIAL}
This supplementary material is organized as follows. \autoref{app:cca_vs_cl}
compares the identifiability frameworks of nonlinear CCA and contrastive learning.
\autoref{sec:proofs} contains complete proofs of the main theoretical results
and \autoref{app:whitening} isolates the role of whitening in the identifiability
argument. \autoref{sec:sketch} provides proof sketches for additional latent distribution
families. \autoref{sec:implmentation} describes implementation details
and \autoref{sec:results} reports additional experimental results. Accompanying
source code is provided to support reproducibility and independent verification.
\section{EXTENDED RELATED WORK: COMPARISON OF IDENTIFIABILITY FRAMEWORKS: NONLINEAR CCA VERSUS CONTRASTIVE LEARNING}
\label{app:cca_vs_cl}

While both nonlinear Canonical Correlation Analysis (CCA) and Contrastive Learning (CL),
e.g., InfoNCE, serve as foundational paradigms for multi-view representation
learning, their underlying theoretical mechanisms for achieving identifiability
differ fundamentally. Recent works have established strong identifiability guarantees
for contrastive methods
\citep{hyvarinen2019nonlinear, zimmermann2021contrastive, matthes2023towards}.
However, unifying CL and CCA under a single theoretical framework remains challenging
due to divergent structural assumptions, regularization mechanisms, and proof techniques.
We systematically compare these two theoretical frameworks below.

\paragraph{1. Distributional Assumptions}
\begin{itemize}
    \item \textbf{Contrastive Learning:} CL identifiability typically assumes the ground-truth latents are sampled from an exponential family distribution. Crucially, it relies on non-Gaussianity (often requiring that at most one latent dimension is Gaussian, following classical ICA theory) or imposes no constraints on the marginal distributions \citep{hyvarinen2019nonlinear}. CL models remain identifiable even when the latent spaces are strictly bounded or distributed on a hypersphere (e.g., von Mises-Fisher distributions).
    \item \textbf{Nonlinear CCA:} In contrast, our nonlinear CCA framework requires strong joint distributional assumptions across all latent dimensions, specifically relying on families that admit an orthogonal polynomial expansion (such as the joint Gaussian priors driving the Mehler expansion). Consequently, nonlinear CCA becomes theoretically non-identifiable under bounded or hyperspherical latent spaces---precisely the regimes where CL theories thrive.
\end{itemize}

\paragraph{2. Regularization Mechanisms}
\begin{itemize}
    \item \textbf{Contrastive Learning:} CL objectives inherently regularize the geometry of the representation space. This is typically achieved via unit-norm constraints on the embeddings (e.g., temperature-scaled InfoNCE) or by assuming bounded latent spaces, which prevents the representations from diverging.
    \item \textbf{Nonlinear CCA:} CCA relies strictly on representation whitening to achieve well-conditioning. As demonstrated in our theoretical analysis, unit variance ensures the compactness of the representation functions, while zero cross-correlation guarantees orthogonality between the components in the function space. Without whitening, the orthogonal polynomial expansion collapses, and the CCA objective cannot reliably separate first-order components from higher-order artifacts.
\end{itemize}

\paragraph{3. Proof Techniques}
\begin{itemize}
    \item \textbf{Contrastive Learning:} The standard proof mechanism for CL demonstrates that, under exponential family generative models, the contrastive loss forces the learned representation to act as an isometry. By preserving the underlying metric structure of the data-generating process, the learned features asymptotically align with the true latents \citep{zimmermann2021contrastive}.
    \item \textbf{Nonlinear CCA:} Our analysis of CCA operates via function expansion in an infinitesimal function space. The proof mechanism relies on decomposing the representation mapping into an orthogonal polynomial basis (e.g., normalized Hermite polynomials). Identifiability is established by proving that the CCA objective uniquely isolates the first-order basis components that maximize canonical correlations, provided that structural conditions---such as the First-Order Canonical Dominance (\autoref{assump:canonical-correlation-separation})---hold.
\end{itemize}

\paragraph{4. Identifiability Guarantees}
\begin{itemize}
    \item \textbf{Contrastive Learning:} Because CL forces metric preservation, it yields exceptionally strong global identifiability conclusions. When the order of the sufficient statistics is known, CL can achieve permutation or block-permutation disentanglement (up to sign flips) \citep{matthes2023towards}.
    \item \textbf{Nonlinear CCA:} CCA achieves strictly \textit{affine} identifiability; the ground-truth representation is recovered up to a linear shift and an orthogonal rotation (after whitening). While this represents a broader equivalence class than permutation disentanglement, our theory directly explains the empirical success of purely non-contrastive, correlation-based methods (e.g., Barlow Twins \citep{zbontar2021barlow}, W-MSE \citep{ermolov2021whitening}). It proves that exact latent recovery is possible under minimal architectural complexity, entirely circumventing the need for negative sampling.
\end{itemize}

\begin{remark}
Modern non-contrastive pipelines often incorporate architectural mechanisms such as momentum encoders and stop-gradient pathways (e.g., BYOL, SimSiam). These dynamics are not yet rigorously captured by current identifiability frameworks for either ICA/CL or CCA-type methods, representing an important frontier for future theoretical unification.
\end{remark}
\section{Further Remarks on Theoretical Results}\label{app:remarks}
\begin{remark}[Role of Canonical-Correlation Separation]
The condition in \autoref{assump:canonical-correlation-separation} is precisely the threshold between strictly affine recovery and contamination by higher-order nonlinearities. It requires that the weakest first-order canonical correlation strictly exceed the largest correlation induced by any second-order term in the normalized Hermite expansion. Since nonlinear CCA ranks candidate directions solely by their inter-view correlations, irrespective of polynomial degree, this strict separation prevents any higher-order Hermite component from entering the top-$d_{\mathcal{Z}}$ representation. Consequently, the learned subspace is supported only on first-order terms, which is exactly what yields strictly affine identifiability.

If this separation condition is relaxed, the ordering argument no 
longer applies. Higher-order terms associated with strongly correlated 
latent factors can then outrank first-order terms associated with weaker 
factors, so the learned representation generally becomes a degree-mixed 
polynomial subspace determined jointly by the full canonical spectrum 
$\{\rho_i\}_{i=1}^{d_{\mathcal{S}}}$ and the representation dimension 
$d_{\mathcal{Z}}$. Nevertheless, because the canonical correlations of 
Hermite components decay exponentially with the polynomial degree, 
enlarging the representation budget eventually brings all first-order 
terms into the retained subspace in the limit $d_{\mathcal{Z}} \to \infty$, 
albeit entangled with higher-order artifacts. See 
Appendix \ref{app:higher_order_nonlinearities}.
\end{remark}
\begin{remark}[Interpretation of A3–A4]
    Assumptions A3 and A4 do not impose further structural restrictions on the
    latent data generation process. Instead, they constitute standard conditions
    for establishing the argmax consistency of M-estimators applied to the empirical CCA objective.
    Specifically, Assumption A3 gives a uniform law of large numbers alongside a ridge schedule
    that ensures stable empirical whitening operators. Assumption A4 requires the optimization
    procedure to yield approximate maximizers of the empirical objective. We formulate these
    conditions uniformly to facilitate a transparent application of standard argmax theorems,
    weaker pointwise variants, complemented by compactness and continuity requirements, would
    also suffice at the expense of increased technical complexity.
\end{remark}
\section{Proofs of Main Theoretical Results}\label{sec:proofs}
\subsection{Proof of \autoref{prop:rep-inv}}
\label{app:proof-rep-inv}
\begin{lemma}[Pushforward identities and whitening preservation]
\label{lem:pushforward}
Let $(\mathbf s,\mathbf s')$ be square-integrable random vectors on
$\mathcal S\times\mathcal S$ with joint law $P_{\mathbf s\mathbf s'}$.
Let
$\mathbf g\colon \mathcal S\to\mathcal X$
and
$\mathbf g'\colon \mathcal S\to\mathcal X'$
be Borel-measurable, and define
\[
\mathbf x=\mathbf g(\mathbf s),
\qquad
\mathbf x'=\mathbf g'(\mathbf s').
\]
Denote
\[
P_{\mathbf x}=\mathbf g_{\#}P_{\mathbf s},
\qquad
P_{\mathbf x'}=\mathbf g'_{\#}P_{\mathbf s'},
\qquad
P_{\mathbf x\mathbf x'}=(\mathbf g,\mathbf g')_{\#}P_{\mathbf s\mathbf s'}.
\]
Then the following properties hold.
\begin{enumerate}[label=\arabic*.]
\item \textbf{Square integrability and isometry.}
For any
$\bm\phi\in L^2(P_{\mathbf x};\mathbb R^{d_{\mathcal Z}})$ and
$\bm\phi'\in L^2(P_{\mathbf x'};\mathbb R^{d_{\mathcal Z}})$,
\[
\bm\phi\circ \mathbf g\in L^2(P_{\mathbf s};\mathbb R^{d_{\mathcal Z}}),
\qquad
\bm\phi'\circ \mathbf g'\in L^2(P_{\mathbf s'};\mathbb R^{d_{\mathcal Z}}),
\]
and
\[
\|\bm\phi\circ \mathbf g\|_{L^2(P_{\mathbf s})}
=
\|\bm\phi\|_{L^2(P_{\mathbf x})},
\qquad
\|\bm\phi'\circ \mathbf g'\|_{L^2(P_{\mathbf s'})}
=
\|\bm\phi'\|_{L^2(P_{\mathbf x'})}.
\]

\item \textbf{Expectation and covariance preservation.}
For all
$\bm\phi\in L^2(P_{\mathbf x};\mathbb R^{d_{\mathcal Z}})$ and
$\bm\phi'\in L^2(P_{\mathbf x'};\mathbb R^{d_{\mathcal Z}})$,
\[
\mathbb E[\bm\phi(\mathbf x)]
=
\mathbb E[\bm\phi(\mathbf g(\mathbf s))],
\qquad
\mathbb E[\bm\phi'(\mathbf x')]
=
\mathbb E[\bm\phi'(\mathbf g'(\mathbf s'))],
\]
\[
\mathrm{Cov}(\bm\phi(\mathbf x))
=
\mathrm{Cov}(\bm\phi(\mathbf g(\mathbf s))),
\qquad
\mathrm{Cov}(\bm\phi'(\mathbf x'))
=
\mathrm{Cov}(\bm\phi'(\mathbf g'(\mathbf s'))),
\]
and
\[
\mathrm{Cov}(\bm\phi(\mathbf x),\bm\phi'(\mathbf x'))
=
\mathrm{Cov}(\bm\phi(\mathbf g(\mathbf s)),\bm\phi'(\mathbf g'(\mathbf s'))).
\]

\item \textbf{Whitening preservation.}
If
$\tilde{\mathbf f}\in\tilde{\mathcal F}_{\mathcal X}$ and
$\tilde{\mathbf f}'\in\tilde{\mathcal F}'_{\mathcal X'}$,
then
\[
\tilde{\mathbf f}\circ \mathbf g\in \hat{\mathcal F}_{\mathcal S},
\qquad
\tilde{\mathbf f}'\circ \mathbf g'\in \hat{\mathcal F}'_{\mathcal S}.
\]
Equivalently,
\[
\tilde{\mathcal F}_{\mathcal S}\subseteq \hat{\mathcal F}_{\mathcal S},
\qquad
\tilde{\mathcal F}'_{\mathcal S}\subseteq \hat{\mathcal F}'_{\mathcal S}.
\]
\end{enumerate}
\end{lemma}

\begin{proof}
Since
$P_{\mathbf x}=\mathbf g_{\#}P_{\mathbf s}$
and
$P_{\mathbf x'}=\mathbf g'_{\#}P_{\mathbf s'}$,
the defining property of pushforward measures yields, for every nonnegative
or integrable Borel function $\varphi$,
\[
\int_{\mathcal X}\varphi(x)\,dP_{\mathbf x}(x)
=
\int_{\mathcal S}\varphi(\mathbf g(s))\,dP_{\mathbf s}(s),
\qquad
\int_{\mathcal X'}\varphi(x')\,dP_{\mathbf x'}(x')
=
\int_{\mathcal S}\varphi(\mathbf g'(s'))\,dP_{\mathbf s'}(s').
\]

For Part~1, apply this identity with
$\varphi(x)=\|\bm\phi(x)\|^2$
and
$\varphi(x')=\|\bm\phi'(x')\|^2$.
Then
\[
\mathbb E\!\left[\|\bm\phi(\mathbf g(\mathbf s))\|^2\right]
=
\mathbb E\!\left[\|\bm\phi(\mathbf x)\|^2\right]
<\infty,
\]
and similarly for $\bm\phi'$. The stated $L^2$-norm equalities follow immediately.

For Part~2, the same pushforward identity with vector-valued integrands gives
the expectation formulas. The covariance identities then follow from
\[
\mathrm{Cov}(\mathbf u,\mathbf v)
=
\mathbb E[\mathbf u\mathbf v^\top]
-
\mathbb E[\mathbf u]\mathbb E[\mathbf v]^\top.
\]
Indeed, since
$P_{\mathbf x\mathbf x'}=(\mathbf g,\mathbf g')_{\#}P_{\mathbf s\mathbf s'}$,
we have
\[
\mathbb E[\bm\phi(\mathbf x)\bm\phi'(\mathbf x')^\top]
=
\mathbb E[\bm\phi(\mathbf g(\mathbf s))\bm\phi'(\mathbf g'(\mathbf s'))^\top].
\]

Part~3 is immediate from Part~2: if $\tilde{\mathbf f}$ and $\tilde{\mathbf f}'$
are centered and whitened under $P_{\mathbf x}$ and $P_{\mathbf x'}$, then
\[
\mathbb E[(\tilde{\mathbf f}\circ \mathbf g)(\mathbf s)] = \mathbf 0,
\qquad
\mathrm{Cov}((\tilde{\mathbf f}\circ \mathbf g)(\mathbf s)) = \mathbf I_{d_{\mathcal Z}},
\]
and likewise for $\tilde{\mathbf f}'\circ \mathbf g'$.
Hence the compositions belong to the feasible source-space classes.
\end{proof}

\begin{proof}[Proof of \autoref{prop:rep-inv}]
Throughout this proof, functions in $L^2$ are identified up to almost-sure equality.

\paragraph{1. Objective preservation.}
Let
$(\tilde{\mathbf f},\tilde{\mathbf f}')
\in
\tilde{\mathcal F}_{\mathcal X}\times\tilde{\mathcal F}'_{\mathcal X'}$.
By \autoref{lem:pushforward}.2,
\[
\mathrm{Cov}(\tilde{\mathbf f}(\mathbf x),\tilde{\mathbf f}'(\mathbf x'))
=
\mathrm{Cov}((\tilde{\mathbf f}\circ \mathbf g)(\mathbf s),(\tilde{\mathbf f}'\circ \mathbf g')(\mathbf s')).
\]
Therefore the two cross-covariance matrices have the same singular values, and hence
\[
J(\tilde{\mathbf f},\tilde{\mathbf f}')
=
J_{\mathcal S}(\tilde{\mathbf f}\circ \mathbf g,\tilde{\mathbf f}'\circ \mathbf g').
\]

\paragraph{2. Maximizer correspondence on representable classes.}
Define
\[
\Psi:
\tilde{\mathcal F}_{\mathcal X}\times\tilde{\mathcal F}'_{\mathcal X'}
\to
\tilde{\mathcal F}_{\mathcal S}\times\tilde{\mathcal F}'_{\mathcal S},
\qquad
\Psi(\tilde{\mathbf f},\tilde{\mathbf f}')
=
(\tilde{\mathbf f}\circ \mathbf g,\tilde{\mathbf f}'\circ \mathbf g').
\]
By definition of
$\tilde{\mathcal F}_{\mathcal S}$ and $\tilde{\mathcal F}'_{\mathcal S}$,
the map $\Psi$ is surjective. Moreover, for any
$\tilde{\mathbf f}_1,\tilde{\mathbf f}_2\in\tilde{\mathcal F}_{\mathcal X}$,
\[
\|\tilde{\mathbf f}_1-\tilde{\mathbf f}_2\|_{L^2(P_{\mathbf x})}^2
=
\|(\tilde{\mathbf f}_1-\tilde{\mathbf f}_2)\circ \mathbf g\|_{L^2(P_{\mathbf s})}^2
=
\|\tilde{\mathbf f}_1\circ \mathbf g-\tilde{\mathbf f}_2\circ \mathbf g\|_{L^2(P_{\mathbf s})}^2
\]
by \autoref{lem:pushforward}.1, and the same holds on the primed side.
Hence $\Psi$ is an isometric bijection modulo null sets.
Combining this with Part~1 gives
\[
(\tilde{\mathbf f}^*,\tilde{\mathbf f}^{\prime *})
\in
\arg\max_{\tilde{\mathcal F}_{\mathcal X}\times\tilde{\mathcal F}'_{\mathcal X'}} J
\quad\Longleftrightarrow\quad
(\tilde{\mathbf f}^*\circ \mathbf g,\tilde{\mathbf f}^{\prime *}\circ \mathbf g')
\in
\arg\max_{\tilde{\mathcal F}_{\mathcal S}\times\tilde{\mathcal F}'_{\mathcal S}} J_{\mathcal S}.
\]

\paragraph{3. Representation universality.}
We prove the claim for $\hat{\mathcal F}_{\mathcal S}$; the primed case is identical.
Since $\mathcal S$ and $\mathcal X$ are standard Borel spaces and
$\mathbf g$ is injective and Borel measurable, the Lusin--Souslin theorem implies that
$\mathbf g(\mathcal S)\subseteq \mathcal X$ is Borel and that
\[
\mathbf g^{-1}:\mathbf g(\mathcal S)\to \mathcal S
\]
is Borel measurable.

Fix $\hat{\mathbf h}\in \hat{\mathcal F}_{\mathcal S}$.
Define the pullback $\bar{\mathbf h}:\mathcal X\to\mathbb R^{d_{\mathcal Z}}$ by
\[
\bar{\mathbf h}(x)
=
\begin{cases}
\hat{\mathbf h}(\mathbf g^{-1}(x)), & x\in \mathbf g(\mathcal S),\\[2mm]
\mathbf 0, & x\notin \mathbf g(\mathcal S).
\end{cases}
\]
Then $\bar{\mathbf h}$ is Borel measurable.
Since $P_{\mathbf x}$ is supported on $\mathbf g(\mathcal S)$,
\[
\bar{\mathbf h}(\mathbf g(\mathbf s))
=
\hat{\mathbf h}(\mathbf s)
\qquad\text{for }P_{\mathbf s}\text{-a.e. }\mathbf s.
\]
By \autoref{lem:pushforward}.1,
\[
\|\bar{\mathbf h}\|_{L^2(P_{\mathbf x})}
=
\|\hat{\mathbf h}\|_{L^2(P_{\mathbf s})}
<
\infty,
\]
so $\bar{\mathbf h}\in L^2(P_{\mathbf x};\mathbb R^{d_{\mathcal Z}})$.
Moreover, by \autoref{lem:pushforward}.2 and the fact that
$\hat{\mathbf h}\in\hat{\mathcal F}_{\mathcal S}$,
\[
\mathbb E[\bar{\mathbf h}(\mathbf x)] = \mathbf 0,
\qquad
\mathrm{Cov}(\bar{\mathbf h}(\mathbf x)) = \mathbf I_{d_{\mathcal Z}}.
\]

By density of the base encoder class $\mathcal H_{\mathcal X}$ in
$L^2(P_{\mathbf x};\mathbb R^{d_{\mathcal Z}})$,
there exists a sequence
$\mathbf f_n\in\mathcal H_{\mathcal X}$
such that
\[
\|\mathbf f_n-\bar{\mathbf h}\|_{L^2(P_{\mathbf x})}\to 0.
\]
Set
\[
\bm\mu_n:=\mathbb E[\mathbf f_n(\mathbf x)],
\qquad
\mathbf\Sigma_n:=\mathrm{Cov}(\mathbf f_n(\mathbf x)).
\]
Because convergence in $L^2$ implies convergence of first and second moments,
\[
\bm\mu_n\to \mathbf 0,
\qquad
\mathbf\Sigma_n\to \mathbf I_{d_{\mathcal Z}}.
\]
In particular, $\mathbf\Sigma_n\succ 0$ for all sufficiently large $n$.
For such $n$, choose the symmetric whitener
\[
\mathbf W_n:=\mathbf\Sigma_n^{-1/2},
\qquad
\tilde{\mathbf f}_n:=\mathbf W_n(\mathbf f_n-\bm\mu_n)\in \tilde{\mathcal F}_{\mathcal X}.
\]
Since the map $A\mapsto A^{-1/2}$ is continuous on the cone of symmetric positive definite matrices and
$\mathbf\Sigma_n\to \mathbf I_{d_{\mathcal Z}}$, we have
$\mathbf W_n\to \mathbf I_{d_{\mathcal Z}}$.
Hence
\begin{align*}
\|\tilde{\mathbf f}_n-\bar{\mathbf h}\|_{L^2(P_{\mathbf x})}
&\le
\|(\mathbf W_n-\mathbf I_{d_{\mathcal Z}})(\mathbf f_n-\bm\mu_n)\|_{L^2(P_{\mathbf x})}
+
\|\mathbf f_n-\bar{\mathbf h}\|_{L^2(P_{\mathbf x})}
+
\|\bm\mu_n\| \\
&\le
\|\mathbf W_n-\mathbf I_{d_{\mathcal Z}}\|\,
\|\mathbf f_n-\bm\mu_n\|_{L^2(P_{\mathbf x})}
+
\|\mathbf f_n-\bar{\mathbf h}\|_{L^2(P_{\mathbf x})}
+
\|\bm\mu_n\|
\to 0.
\end{align*}
Now define
\[
\tilde{\mathbf h}_n:=\tilde{\mathbf f}_n\circ \mathbf g\in \tilde{\mathcal F}_{\mathcal S}.
\]
By \autoref{lem:pushforward}.1,
\[
\|\tilde{\mathbf h}_n-\hat{\mathbf h}\|_{L^2(P_{\mathbf s})}
=
\|\tilde{\mathbf f}_n-\bar{\mathbf h}\|_{L^2(P_{\mathbf x})}
\to 0.
\]
Therefore, for every $\epsilon>0$, there exists $n$ such that
\[
\mathbb E\!\left[
\|\tilde{\mathbf h}_n(\mathbf s)-\hat{\mathbf h}(\mathbf s)\|^2
\right]
<\epsilon^2.
\]
Repeating the same construction for
$\hat{\mathbf h}'\in \hat{\mathcal F}'_{\mathcal S}$
proves the density claim on the primed side.

\paragraph{4. Continuity of $J_{\mathcal S}$ and equality of suprema.}
Let
$(\mathbf h,\mathbf h'),(\mathbf u,\mathbf u')
\in
\hat{\mathcal F}_{\mathcal S}\times\hat{\mathcal F}'_{\mathcal S}$,
and define
\[
\Delta
:=
\mathrm{Cov}(\mathbf h(\mathbf s),\mathbf h'(\mathbf s'))
-
\mathrm{Cov}(\mathbf u(\mathbf s),\mathbf u'(\mathbf s')).
\]
Since all four maps are centered,
\[
\Delta
=
\mathbb E[(\mathbf h-\mathbf u)(\mathbf s)\,\mathbf h'(\mathbf s')^\top]
+
\mathbb E[\mathbf u(\mathbf s)\,(\mathbf h'-\mathbf u')(\mathbf s')^\top].
\]
Hence, by Cauchy--Schwarz,
\[
\|\Delta\|_F
\le
\|\mathbf h-\mathbf u\|_{L^2(P_{\mathbf s})}\,
\|\mathbf h'\|_{L^2(P_{\mathbf s'})}
+
\|\mathbf u\|_{L^2(P_{\mathbf s})}\,
\|\mathbf h'-\mathbf u'\|_{L^2(P_{\mathbf s'})}.
\]
Because every element of
$\hat{\mathcal F}_{\mathcal S}$ and $\hat{\mathcal F}'_{\mathcal S}$
is whitened,
\[
\|\mathbf h\|_{L^2(P_{\mathbf s})}
=
\|\mathbf u\|_{L^2(P_{\mathbf s})}
=
\|\mathbf h'\|_{L^2(P_{\mathbf s'})}
=
\sqrt{\mathrm{tr}(\mathbf I_{d_{\mathcal Z}})}
=
\sqrt{d_{\mathcal Z}}.
\]
Therefore,
\[
\|\Delta\|_F
\le
\sqrt{d_{\mathcal Z}}
\Big(
\|\mathbf h-\mathbf u\|_{L^2(P_{\mathbf s})}
+
\|\mathbf h'-\mathbf u'\|_{L^2(P_{\mathbf s'})}
\Big).
\]
Since $J_{\mathcal S}$ is the nuclear norm of the cross-covariance matrix,
\[
|J_{\mathcal S}(\mathbf h,\mathbf h')-J_{\mathcal S}(\mathbf u,\mathbf u')|
\le
\|\Delta\|_*
\le
\sqrt{d_{\mathcal Z}}\|\Delta\|_F
\le
d_{\mathcal Z}
\Big(
\|\mathbf h-\mathbf u\|_{L^2(P_{\mathbf s})}
+
\|\mathbf h'-\mathbf u'\|_{L^2(P_{\mathbf s'})}
\Big).
\]
Thus $J_{\mathcal S}$ is continuous under the product $L^2$ topology on
$\hat{\mathcal F}_{\mathcal S}\times \hat{\mathcal F}'_{\mathcal S}$.

Because
$\tilde{\mathcal F}_{\mathcal S}$ is dense in $\hat{\mathcal F}_{\mathcal S}$
and
$\tilde{\mathcal F}'_{\mathcal S}$ is dense in $\hat{\mathcal F}'_{\mathcal S}$,
this continuity implies
\[
\sup_{\tilde{\mathcal F}_{\mathcal S}\times\tilde{\mathcal F}'_{\mathcal S}} J_{\mathcal S}
=
\sup_{\hat{\mathcal F}_{\mathcal S}\times\hat{\mathcal F}'_{\mathcal S}} J_{\mathcal S}.
\]
Together with Part~1,
\[
\sup_{\tilde{\mathcal F}_{\mathcal X}\times\tilde{\mathcal F}'_{\mathcal X'}} J
=
\sup_{\tilde{\mathcal F}_{\mathcal S}\times\tilde{\mathcal F}'_{\mathcal S}} J_{\mathcal S}
=
\sup_{\hat{\mathcal F}_{\mathcal S}\times\hat{\mathcal F}'_{\mathcal S}} J_{\mathcal S}.
\]

Finally, if
$(\tilde{\mathbf f}^*,\tilde{\mathbf f}^{\prime *})$
maximizes $J$ over
$\tilde{\mathcal F}_{\mathcal X}\times\tilde{\mathcal F}'_{\mathcal X'}$,
then by Part~2 its composition maximizes $J_{\mathcal S}$ over
$\tilde{\mathcal F}_{\mathcal S}\times\tilde{\mathcal F}'_{\mathcal S}$;
since the supremum over representable classes equals the supremum over the
full feasible classes, the same composed pair is also a maximizer of
$J_{\mathcal S}$ over
$\hat{\mathcal F}_{\mathcal S}\times\hat{\mathcal F}'_{\mathcal S}$.
This proves the proposition.
\end{proof}
\subsection{Proof of the rotation-invariance of CCA}
\label{app:cca_orbit}
In this appendix, we formalize the rotational symmetry underlying the population
CCA objective in \autoref{equ:CCA_objective}. Throughout, for any feasible pair
\(
(\tilde{\mathbf f},\tilde{\mathbf f}')
\in
\tilde{\mathcal F}_{\mathcal X}\times \tilde{\mathcal F}'_{\mathcal X'}
\),
we write
\[
\mathbf C_{\tilde{\mathbf f},\tilde{\mathbf f}'}
:=
\mathrm{Cov}\!\big(\tilde{\mathbf f}(\mathbf x),\tilde{\mathbf f}'(\mathbf x')\big)
\in \mathbb R^{d_{\mathcal Z}\times d_{\mathcal Z}}.
\]

\begin{proposition}[Orthogonal invariance of the whitened CCA objective]
\label{prop:cca_orbit_invariance}
Under \autoref{assump:function-class}, let
\(
(\tilde{\mathbf f},\tilde{\mathbf f}')
\in
\tilde{\mathcal F}_{\mathcal X}\times \tilde{\mathcal F}'_{\mathcal X'}
\)
be any feasible pair, and let
\(
\mathbf Q,\mathbf Q' \in O(d_{\mathcal Z})
\).
Then
\[
\mathbf Q\tilde{\mathbf f}\in \tilde{\mathcal F}_{\mathcal X},
\qquad
\mathbf Q'\tilde{\mathbf f}'\in \tilde{\mathcal F}'_{\mathcal X'},
\]
and the population CCA objective is invariant under these post-transformations:
\[
J(\mathbf Q\tilde{\mathbf f},\,\mathbf Q'\tilde{\mathbf f}')
=
J(\tilde{\mathbf f},\,\tilde{\mathbf f}').
\]
Equivalently, \(J\) is constant on every
\(
O(d_{\mathcal Z})\times O(d_{\mathcal Z})
\)-orbit.
\end{proposition}

\begin{proof}
By construction of the whitened encoder classes, every feasible encoder is
centered and whitened. Hence
\[
\mathbb E[\tilde{\mathbf f}(\mathbf x)] = \mathbf 0,
\qquad
\mathrm{Cov}(\tilde{\mathbf f}(\mathbf x)) = \mathbf I_{d_{\mathcal Z}},
\]
and analogously,
\[
\mathbb E[\tilde{\mathbf f}'(\mathbf x')] = \mathbf 0,
\qquad
\mathrm{Cov}(\tilde{\mathbf f}'(\mathbf x')) = \mathbf I_{d_{\mathcal Z}}.
\]

We first verify feasibility after orthogonal post-transformation. Since
\(
\mathbf Q,\mathbf Q'
\in O(d_{\mathcal Z})
\),
we have
\[
\mathbb E[\mathbf Q\tilde{\mathbf f}(\mathbf x)]
=
\mathbf Q\,\mathbb E[\tilde{\mathbf f}(\mathbf x)]
=
\mathbf 0,
\]
and
\[
\mathrm{Cov}(\mathbf Q\tilde{\mathbf f}(\mathbf x))
=
\mathbf Q\,\mathrm{Cov}(\tilde{\mathbf f}(\mathbf x))\,\mathbf Q^\top
=
\mathbf Q\mathbf I_{d_{\mathcal Z}}\mathbf Q^\top
=
\mathbf I_{d_{\mathcal Z}}.
\]
Thus \(\mathbf Q\tilde{\mathbf f}\) is again zero-mean and whitened. By the
post-orthogonal closure in \autoref{assump:function-class},
\(
\mathbf Q\tilde{\mathbf f}\in \tilde{\mathcal F}_{\mathcal X}
\).
The same argument yields
\(
\mathbf Q'\tilde{\mathbf f}'\in \tilde{\mathcal F}'_{\mathcal X'}
\).

Next, using the zero-mean property of the whitened encoders, the transformed
cross-covariance satisfies
\[
\mathbf C_{\mathbf Q\tilde{\mathbf f},\,\mathbf Q'\tilde{\mathbf f}'}
=
\mathrm{Cov}\!\big(\mathbf Q\tilde{\mathbf f}(\mathbf x),\,\mathbf Q'\tilde{\mathbf f}'(\mathbf x')\big)
=
\mathbf Q\,\mathbf C_{\tilde{\mathbf f},\tilde{\mathbf f}'}\,\mathbf Q'^\top.
\]
Let
\[
\mathbf C_{\tilde{\mathbf f},\tilde{\mathbf f}'}
=
\mathbf U \mathbf D \mathbf V^\top
\]
be a singular value decomposition, where \(\mathbf U,\mathbf V\in O(d_{\mathcal Z})\)
and
\(
\mathbf D=\mathrm{diag}(\sigma_1,\dots,\sigma_{d_{\mathcal Z}})
\)
with \(\sigma_1\ge \cdots \ge \sigma_{d_{\mathcal Z}}\ge 0\). Then
\[
\mathbf C_{\mathbf Q\tilde{\mathbf f},\,\mathbf Q'\tilde{\mathbf f}'}
=
(\mathbf Q\mathbf U)\,\mathbf D\,(\mathbf Q'\mathbf V)^\top.
\]
Since \(\mathbf Q\mathbf U\) and \(\mathbf Q'\mathbf V\) are again orthogonal,
this is a singular value decomposition of
\(
\mathbf C_{\mathbf Q\tilde{\mathbf f},\,\mathbf Q'\tilde{\mathbf f}'}
\).
Therefore,
\[
\sigma_i\!\left(
\mathbf C_{\mathbf Q\tilde{\mathbf f},\,\mathbf Q'\tilde{\mathbf f}'}
\right)
=
\sigma_i\!\left(
\mathbf C_{\tilde{\mathbf f},\tilde{\mathbf f}'}
\right),
\qquad
i=1,\dots,d_{\mathcal Z}.
\]
Summing over \(i\) yields
\[
J(\mathbf Q\tilde{\mathbf f},\,\mathbf Q'\tilde{\mathbf f}')
=
\sum_{i=1}^{d_{\mathcal Z}}
\sigma_i\!\left(
\mathbf C_{\mathbf Q\tilde{\mathbf f},\,\mathbf Q'\tilde{\mathbf f}'}
\right)
=
\sum_{i=1}^{d_{\mathcal Z}}
\sigma_i\!\left(
\mathbf C_{\tilde{\mathbf f},\tilde{\mathbf f}'}
\right)
=
J(\tilde{\mathbf f},\,\tilde{\mathbf f}'),
\]
which proves the claim.
\end{proof}

\begin{corollary}[Orthogonal orbit of equivalent maximizers]
\label{cor:cca_orbit_maximizers}
If
\(
(\tilde{\mathbf f}^*,\tilde{\mathbf f}'^*)
\)
is a maximizer of \autoref{equ:CCA_objective}, then for every
\(
\mathbf Q,\mathbf Q' \in O(d_{\mathcal Z})
\),
the pair
\[
(\mathbf Q\tilde{\mathbf f}^*,\,\mathbf Q'\tilde{\mathbf f}'^*)
\]
is also a maximizer. In particular, each maximizer generates an equivalent
\(
O(d_{\mathcal Z})\times O(d_{\mathcal Z})
\)-orbit of solutions.
\end{corollary}

\begin{proof}
This follows immediately from \autoref{prop:cca_orbit_invariance}, since the
objective value is unchanged by orthogonal post-transformations and the feasible
set is closed under them.
\end{proof}

\begin{remark}[Why whitening leaves an orthogonal ambiguity]
\label{rem:whitening_orthogonal_ambiguity}
The preceding result shows that, after whitening, the residual linear ambiguity
is orthogonal. Indeed, let \(\tilde{\mathbf f}\) be any whitened encoder and let
\(\mathbf A\in \mathrm{GL}(d_{\mathcal Z})\). Then
\[
\mathrm{Cov}(\mathbf A\tilde{\mathbf f}(\mathbf x))
=
\mathbf A\,\mathrm{Cov}(\tilde{\mathbf f}(\mathbf x))\,\mathbf A^\top
=
\mathbf A\mathbf A^\top.
\]
Hence \(\mathbf A\tilde{\mathbf f}\) remains whitened if and only if
\[
\mathbf A\mathbf A^\top = \mathbf I_{d_{\mathcal Z}},
\]
that is, if and only if \(\mathbf A\in O(d_{\mathcal Z})\). Therefore whitening
eliminates arbitrary anisotropic rescalings and shears, leaving precisely the
orthogonal symmetry established above.
\end{remark}
\begin{proposition}[Orthogonal invariance does not imply uniqueness modulo orbit]
\label{prop:cca_orbit_nonunique}
Under \autoref{assump:function-class} alone, the set of maximizers of
\autoref{equ:CCA_objective} need not be a single
\(O(d_{\mathcal Z})\times O(d_{\mathcal Z})\)-orbit.
\end{proposition}

\begin{proof}
Take \(d_{\mathcal Z}=1\), \(\mathcal X=\mathcal X'=\mathbb R\), and let
\[
X=X'\sim \mathcal N(0,1).
\]
Let the base classes be
\[
\mathcal H_{\mathcal X}=\mathcal H_{\mathcal X'}=L^2(P_X;\mathbb R),
\]
so the corresponding whitened classes are exactly the centered, unit-variance
scalar functions.

For any feasible pair \((\tilde f,\tilde f')\),
\[
J(\tilde f,\tilde f')
=
\big|\mathrm{Cov}(\tilde f(X),\tilde f'(X))\big|
\le
\sqrt{\mathrm{Var}(\tilde f(X))\,\mathrm{Var}(\tilde f'(X))}
=1
\]
by Cauchy--Schwarz. Hence every pair of the form \((h,h)\), where
\[
\mathbb E[h(X)]=0,
\qquad
\mathrm{Var}(h(X))=1,
\]
is a population maximizer.

In particular, the two feasible functions
\[
h_1(X)=X,
\qquad
h_2(X)=\frac{X^2-1}{\sqrt 2}
\]
give two maximizers \((h_1,h_1)\) and \((h_2,h_2)\).
Since \(O(1)=\{\pm1\}\) and \(h_2\neq \pm h_1\) in \(L^2(P_X)\), these two
maximizers are not in the same orbit.
\end{proof}

\begin{remark}[Scope of \autoref{prop:cca_orbit_invariance}]
\label{rem:cca_orbit_scope}
\autoref{prop:cca_orbit_invariance} and \autoref{cor:cca_orbit_maximizers}
show only that each maximizer generates an equivalent
\(O(d_{\mathcal Z})\times O(d_{\mathcal Z})\)-orbit of maximizers.
They do not imply that all maximizers belong to a single orbit.

The stronger single-orbit statement used later in the paper requires the
additional distributional and spectral assumptions of
\autoref{thm:affine-id} (see \autoref{cor:cca_single_orbit}).
\end{remark}
\subsection{Proof of \autoref{thm:affine-id}}
\label{app:proof-affine-id}

Throughout this subsection, write
\[
d:=d_{\mathcal S}=d_{\mathcal Z}.
\]

\begin{lemma}[Hermite--Mehler expansion of a bivariate Gaussian pair]
\label{lemma:hermite_expansion_bivariate_gaussian}
Let $(S,S')$ be jointly Gaussian with means $\mu_S,\mu_{S'}$, variances
$\sigma_S^2,\sigma_{S'}^2$, and correlation coefficient $t\in(-1,1)$.
Define the standardized coordinates
\[
X:=\frac{S-\mu_S}{\sigma_S},\qquad
Y:=\frac{S'-\mu_{S'}}{\sigma_{S'}},\qquad
U:=\frac{X}{\sqrt 2},\qquad
V:=\frac{Y}{\sqrt 2}.
\]
Let $H_n$ denote the physicists' Hermite polynomials and define
\[
\psi_n(z):=\frac{1}{\sqrt{2^n n!}}\,H_n(z),\qquad n\in\mathbb N_0.
\]
Then $\{\psi_n\}_{n\ge 0}$ is an orthonormal basis of
$L^2(\nu)$, where
\[
\nu(dz)=\pi^{-1/2}e^{-z^2}\,dz,
\]
and the joint density of $(S,S')$ admits the Hermite--Mehler expansion
\begin{equation}
\label{eq:Hermite-expansion-density}
p_{S,S'}(s,s')
=
\frac{1}{2\pi\,\sigma_S\sigma_{S'}}
e^{-(u^2+v^2)}
\sum_{n=0}^{\infty} t^n\,\psi_n(u)\psi_n(v),
\end{equation}
where
\[
u=\frac{s-\mu_S}{\sqrt{2}\,\sigma_S},
\qquad
v=\frac{s'-\mu_{S'}}{\sqrt{2}\,\sigma_{S'}}.
\]
Equivalently, the joint density of $(U,V)$ is
\[
p_{U,V}(u,v)
=
\frac{1}{\pi}
e^{-(u^2+v^2)}
\sum_{n=0}^{\infty} t^n\,\psi_n(u)\psi_n(v).
\]
In particular,
\[
\mathbb E[\psi_m(U)\psi_n(V)] = t^n\,\delta_{mn},
\qquad m,n\in\mathbb N_0.
\]
\end{lemma}

\begin{proof}
The classical Mehler formula for the physicists' Hermite polynomials states
that, for $|t|<1$,
\[
\sum_{n=0}^{\infty}\frac{t^n}{2^n n!}H_n(u)H_n(v)
=
\frac{1}{\sqrt{1-t^2}}
\exp\!\left(
\frac{2tuv-t^2(u^2+v^2)}{1-t^2}
\right).
\]
Multiplying both sides by $\pi^{-1}e^{-(u^2+v^2)}$ gives
\[
\frac{1}{\pi}
e^{-(u^2+v^2)}
\sum_{n=0}^{\infty}\frac{t^n}{2^n n!}H_n(u)H_n(v)
=
\frac{1}{\pi\sqrt{1-t^2}}
\exp\!\left(
-\frac{u^2-2tuv+v^2}{1-t^2}
\right),
\]
which is precisely the density of $(U,V)$. Since
\[
\psi_n(u)\psi_n(v)=\frac{1}{2^n n!}H_n(u)H_n(v),
\]
this proves the expansion for $p_{U,V}$, and the formula for $p_{S,S'}$
follows by the change of variables $(s,s')\mapsto(u,v)$, whose Jacobian
determinant is $(2\sigma_S\sigma_{S'})^{-1}$.

Orthogonality is immediate from
\[
\int_{\mathbb R}\psi_m(z)\psi_n(z)\,\nu(dz)=\delta_{mn}.
\]
Finally,
\begin{align*}
\mathbb E[\psi_m(U)\psi_n(V)]
&=
\int_{\mathbb R^2}\psi_m(u)\psi_n(v)\,p_{U,V}(u,v)\,du\,dv \\
&=
\int_{\mathbb R^2}
\psi_m(u)\psi_n(v)
\Bigg(
\sum_{k=0}^{\infty} t^k\psi_k(u)\psi_k(v)
\Bigg)\nu(du)\nu(dv) \\
&=
\sum_{k=0}^{\infty} t^k
\left(\int_{\mathbb R}\psi_m(u)\psi_k(u)\,\nu(du)\right)
\left(\int_{\mathbb R}\psi_n(v)\psi_k(v)\,\nu(dv)\right) \\
&= t^n\,\delta_{mn}.
\end{align*}
\end{proof}

\begin{proposition}[Multivariate Hermite--Mehler expansion in canonical coordinates]
\label{prop:multi_variante_hermite}
Let $(\mathbf u,\mathbf v)\in\mathbb R^d\times\mathbb R^d$ be jointly Gaussian such that
$\sqrt 2\,\mathbf u$ and $\sqrt 2\,\mathbf v$ are standard Gaussian and
\[
\mathrm{Cov}(\sqrt 2\,\mathbf u,\sqrt 2\,\mathbf v)
=
\operatorname{diag}(\rho_1,\dots,\rho_d),
\qquad
1>\rho_1\ge \cdots \ge \rho_d>0.
\]
Define, for every multi-index
$\mathbf n=(n_1,\dots,n_d)\in\mathbb N_0^d$,
\[
\Psi_{\mathbf n}(\mathbf z):=\prod_{i=1}^d \psi_{n_i}(z_i),
\qquad
\rho^{\mathbf n}:=\prod_{i=1}^d \rho_i^{n_i}.
\]
Then the coordinate pairs $(u_i,v_i)$ are mutually independent across
$i=1,\dots,d$, and the joint density of $(\mathbf u,\mathbf v)$ is
\begin{equation}
\label{eq:multivariate-Hermite-Mehler}
p_{\mathbf u,\mathbf v}(\mathbf u,\mathbf v)
=
\frac{1}{\pi^d}
e^{-(\|\mathbf u\|^2+\|\mathbf v\|^2)}
\sum_{\mathbf n\in\mathbb N_0^d}
\rho^{\mathbf n}\,
\Psi_{\mathbf n}(\mathbf u)\Psi_{\mathbf n}(\mathbf v).
\end{equation}
Moreover, for all multi-indices $\mathbf m,\mathbf n\in\mathbb N_0^d$,
\[
\mathbb E\!\left[\Psi_{\mathbf m}(\mathbf u)\Psi_{\mathbf n}(\mathbf v)\right]
=
\rho^{\mathbf n}\,\delta_{\mathbf m\mathbf n},
\qquad
\delta_{\mathbf m\mathbf n}:=\prod_{i=1}^d \delta_{m_i n_i}.
\]
\end{proposition}

\begin{proof}
Because $(\sqrt2\,\mathbf u,\sqrt2\,\mathbf v)$ is jointly Gaussian with
block covariance
\[
\begin{pmatrix}
\mathbf I_d & \operatorname{diag}(\rho_1,\dots,\rho_d) \\
\operatorname{diag}(\rho_1,\dots,\rho_d) & \mathbf I_d
\end{pmatrix},
\]
the pairs $(u_i,v_i)$ are mutually independent across $i$. Hence
\[
p_{\mathbf u,\mathbf v}(\mathbf u,\mathbf v)
=
\prod_{i=1}^d p_{u_i,v_i}(u_i,v_i).
\]
Applying \autoref{lemma:hermite_expansion_bivariate_gaussian} coordinatewise,
\[
p_{u_i,v_i}(u_i,v_i)
=
\frac{1}{\pi}e^{-(u_i^2+v_i^2)}
\sum_{n_i=0}^{\infty}\rho_i^{n_i}\psi_{n_i}(u_i)\psi_{n_i}(v_i).
\]
Taking the product over $i=1,\dots,d$ yields
\[
p_{\mathbf u,\mathbf v}(\mathbf u,\mathbf v)
=
\frac{1}{\pi^d}
e^{-(\|\mathbf u\|^2+\|\mathbf v\|^2)}
\prod_{i=1}^d
\left(
\sum_{n_i=0}^{\infty}\rho_i^{n_i}\psi_{n_i}(u_i)\psi_{n_i}(v_i)
\right),
\]
which expands into \autoref{eq:multivariate-Hermite-Mehler}. The covariance
identity follows by Fubini and the one-dimensional orthogonality:
\begin{align*}
\mathbb E\!\left[\Psi_{\mathbf m}(\mathbf u)\Psi_{\mathbf n}(\mathbf v)\right]
&=
\prod_{i=1}^d
\mathbb E\!\left[\psi_{m_i}(u_i)\psi_{n_i}(v_i)\right] \\
&=
\prod_{i=1}^d \rho_i^{n_i}\delta_{m_i n_i}
=
\rho^{\mathbf n}\,\delta_{\mathbf m\mathbf n}.
\end{align*}
\end{proof}

\begin{corollary}[Orthogonality of multivariate normalized Hermite polynomials]
\label{cor:multi_hermite_orthogonality}
The family
\[
\big\{\Psi_{\mathbf n}:\mathbf n\in\mathbb N_0^d\big\}
\]
is an orthonormal basis of $L^2(\nu_d)$, where
\[
\nu_d(d\mathbf z)=\pi^{-d/2}e^{-\|\mathbf z\|^2}\,d\mathbf z.
\]
In particular,
\[
\Psi_{\mathbf 0}\equiv 1,
\qquad
\Psi_{\mathbf e_i}(\mathbf z)=\psi_1(z_i)=\sqrt2\,z_i,
\quad i=1,\dots,d.
\]
\end{corollary}

\begin{proof}
For $\mathbf m,\mathbf n\in\mathbb N_0^d$,
\begin{align*}
\int_{\mathbb R^d}\Psi_{\mathbf m}(\mathbf z)\Psi_{\mathbf n}(\mathbf z)\,\nu_d(d\mathbf z)
&=
\prod_{i=1}^d
\int_{\mathbb R}\psi_{m_i}(z_i)\psi_{n_i}(z_i)\,\nu(dz_i) \\
&=
\prod_{i=1}^d \delta_{m_i n_i}
=
\delta_{\mathbf m\mathbf n}.
\end{align*}
Completeness follows from the standard tensor-product construction of
orthonormal bases. The identity for $\Psi_{\mathbf e_i}$ uses
$H_1(z)=2z$.
\end{proof}

\begin{lemma}[Diagonal selection under orthonormal row constraints]
\label{lem:diag-pick-topN}
Let $N\in\mathbb N$, let $\{d_j\}_{j\ge1}$ be a nonnegative nonincreasing
sequence, and let
\[
\mathbf A=(a_{ij})_{1\le i\le N,\;j\ge1},
\qquad
\mathbf B=(b_{ij})_{1\le i\le N,\;j\ge1}
\]
be real matrices with countably many columns such that
\[
\mathbf A\mathbf A^\top=\mathbf I_N,
\qquad
\mathbf B\mathbf B^\top=\mathbf I_N.
\]
Let
\[
\mathbf D=\operatorname{diag}(d_1,d_2,\dots).
\]
Then
\[
\big\|\mathbf A\mathbf D\mathbf B^\top\big\|_*
\;\le\;
\sum_{i=1}^N d_i.
\]
Moreover, equality is attained, for example, by
\[
\mathbf A=\mathbf B=\big[\mathbf I_N\ \ 0\ \ 0\ \ \cdots\big].
\]
\end{lemma}

\begin{proof}
Let $\mathbf a_j,\mathbf b_j\in\mathbb R^N$ denote the $j$-th columns of
$\mathbf A^\top$ and $\mathbf B^\top$, respectively. Then
\[
\sum_{j\ge1}\mathbf a_j\mathbf a_j^\top=\mathbf I_N,
\qquad
\sum_{j\ge1}\mathbf b_j\mathbf b_j^\top=\mathbf I_N,
\]
hence
\[
\sum_{j\ge1}\|\mathbf a_j\|^2=N,
\qquad
\sum_{j\ge1}\|\mathbf b_j\|^2=N.
\]
Also, since $\mathbf A^\top\mathbf A\preceq \mathbf I$ and
$\mathbf B^\top\mathbf B\preceq \mathbf I$, we have
\[
0\le \|\mathbf a_j\|^2\le 1,
\qquad
0\le \|\mathbf b_j\|^2\le 1
\qquad \text{for all } j\ge 1.
\]

By the variational characterization of the nuclear norm,
\[
\big\|\mathbf A\mathbf D\mathbf B^\top\big\|_*
=
\max_{\mathbf R,\mathbf S\in O(N)}
\operatorname{tr}\!\big(\mathbf R\mathbf A\mathbf D\mathbf B^\top \mathbf S^\top\big)
=
\max_{\mathbf R,\mathbf S\in O(N)}
\sum_{j\ge1} d_j\,\langle \mathbf R\mathbf a_j,\mathbf S\mathbf b_j\rangle.
\]
Using Cauchy--Schwarz and $2ab\le a^2+b^2$,
\[
\langle \mathbf R\mathbf a_j,\mathbf S\mathbf b_j\rangle
\le \|\mathbf a_j\|\,\|\mathbf b_j\|
\le \frac{1}{2}\Big(\|\mathbf a_j\|^2+\|\mathbf b_j\|^2\Big).
\]
Therefore
\[
\big\|\mathbf A\mathbf D\mathbf B^\top\big\|_*
\le
\frac12\sum_{j\ge1} d_j\|\mathbf a_j\|^2
+
\frac12\sum_{j\ge1} d_j\|\mathbf b_j\|^2.
\]
Now, if $\{u_j\}_{j\ge1}$ satisfies $0\le u_j\le 1$ and
$\sum_{j\ge1}u_j=N$, then the monotonicity of $\{d_j\}_{j\ge1}$ implies
\[
\sum_{j\ge1} d_j u_j \le \sum_{i=1}^N d_i;
\]
indeed, the maximum is achieved by taking $u_1=\cdots=u_N=1$ and
$u_j=0$ for $j>N$. Applying this with
$u_j=\|\mathbf a_j\|^2$ and $u_j=\|\mathbf b_j\|^2$ yields
\[
\sum_{j\ge1} d_j\|\mathbf a_j\|^2 \le \sum_{i=1}^N d_i,
\qquad
\sum_{j\ge1} d_j\|\mathbf b_j\|^2 \le \sum_{i=1}^N d_i.
\]
Hence
\[
\big\|\mathbf A\mathbf D\mathbf B^\top\big\|_*
\le \sum_{i=1}^N d_i.
\]

Finally, for
$\mathbf A=\mathbf B=\big[\mathbf I_N\ \ 0\ \ 0\ \ \cdots\big]$ we have
\[
\mathbf A\mathbf D\mathbf B^\top
=
\operatorname{diag}(d_1,\dots,d_N),
\]
whose nuclear norm is exactly $\sum_{i=1}^N d_i$.
\end{proof}

\begin{corollary}[Strict-gap characterization of maximizers]
\label{col:diag-pick-topN}
Under the assumptions of \autoref{lem:diag-pick-topN}, assume in addition that
\[
d_N>d_{N+1}.
\]
If
\[
\big\|\mathbf A\mathbf D\mathbf B^\top\big\|_*
=
\sum_{i=1}^N d_i,
\]
then
\[
\mathbf A=\big[\mathbf A_1\ \ 0\ \ 0\ \ \cdots\big],
\qquad
\mathbf B=\big[\mathbf B_1\ \ 0\ \ 0\ \ \cdots\big]
\]
for some $\mathbf A_1,\mathbf B_1\in O(N)$.
\end{corollary}

\begin{proof}
If equality holds in \autoref{lem:diag-pick-topN}, then equality must hold in
both weighted-sum bounds
\[
\sum_{j\ge1} d_j\|\mathbf a_j\|^2 \le \sum_{i=1}^N d_i,
\qquad
\sum_{j\ge1} d_j\|\mathbf b_j\|^2 \le \sum_{i=1}^N d_i.
\]
Because $d_N>d_{N+1}$ and
\[
0\le \|\mathbf a_j\|^2\le 1,\qquad \sum_{j\ge1}\|\mathbf a_j\|^2=N,
\]
the only way to achieve
\[
\sum_{j\ge1} d_j\|\mathbf a_j\|^2 = \sum_{i=1}^N d_i
\]
is to place all mass on the first $N$ coordinates, i.e.
\[
\|\mathbf a_j\|^2=1 \ \text{for } 1\le j\le N,
\qquad
\|\mathbf a_j\|^2=0 \ \text{for } j>N.
\]
The same argument gives
\[
\|\mathbf b_j\|^2=1 \ \text{for } 1\le j\le N,
\qquad
\|\mathbf b_j\|^2=0 \ \text{for } j>N.
\]
Hence all columns indexed by $j>N$ vanish, so
\[
\mathbf A=\big[\mathbf A_1\ \ 0\ \ 0\ \ \cdots\big],
\qquad
\mathbf B=\big[\mathbf B_1\ \ 0\ \ 0\ \ \cdots\big]
\]
with $\mathbf A_1,\mathbf B_1\in\mathbb R^{N\times N}$. Since
$\mathbf A\mathbf A^\top=\mathbf B\mathbf B^\top=\mathbf I_N$, we obtain
\[
\mathbf A_1\mathbf A_1^\top=\mathbf I_N,
\qquad
\mathbf B_1\mathbf B_1^\top=\mathbf I_N.
\]
Because $\mathbf A_1$ and $\mathbf B_1$ are square, this means
$\mathbf A_1,\mathbf B_1\in O(N)$.
\end{proof}

\begin{proof}[Proof of \autoref{thm:affine-id}]
By \autoref{prop:rep-inv}, it suffices to characterize the maximizers of the
source-space objective over
\[
\hat{\mathcal F}_{\mathcal S}\times \hat{\mathcal F}'_{\mathcal S}.
\]
Let
\[
\mathbf K
:=
\mathbf\Sigma_{\mathbf s\mathbf s}^{-1/2}
\mathbf\Sigma_{\mathbf s\mathbf s'}
\mathbf\Sigma_{\mathbf s'\mathbf s'}^{-1/2}
=
\mathbf U_0\,\operatorname{diag}(\rho_1,\dots,\rho_d)\,\mathbf V_0^\top
\]
be a singular value decomposition, where
$\mathbf U_0,\mathbf V_0\in O(d)$ and
\[
1>\rho_1\ge \cdots \ge \rho_d>0.
\]
Define the canonical Gaussian coordinates
\[
\mathbf u
:=
\frac{1}{\sqrt 2}\,
\mathbf U_0^\top
\mathbf\Sigma_{\mathbf s\mathbf s}^{-1/2}
(\mathbf s-\bm\mu_{\mathbf s}),
\qquad
\mathbf v
:=
\frac{1}{\sqrt 2}\,
\mathbf V_0^\top
\mathbf\Sigma_{\mathbf s'\mathbf s'}^{-1/2}
(\mathbf s'-\bm\mu_{\mathbf s'}).
\]
Then $\sqrt2\,\mathbf u$ and $\sqrt2\,\mathbf v$ are standard Gaussian, and
\[
\mathrm{Cov}(\sqrt2\,\mathbf u,\sqrt2\,\mathbf v)
=
\operatorname{diag}(\rho_1,\dots,\rho_d).
\]
Hence \autoref{prop:multi_variante_hermite} applies.

The maps
\[
\mathbf s
=
\bm\mu_{\mathbf s}
+
\sqrt2\,\mathbf\Sigma_{\mathbf s\mathbf s}^{1/2}\mathbf U_0\,\mathbf u,
\qquad
\mathbf s'
=
\bm\mu_{\mathbf s'}
+
\sqrt2\,\mathbf\Sigma_{\mathbf s'\mathbf s'}^{1/2}\mathbf V_0\,\mathbf v
\]
are affine bijections. Therefore, maximizing over centered, whitened maps of
$\mathbf s$ and $\mathbf s'$ is equivalent to maximizing over centered,
whitened maps of $\mathbf u$ and $\mathbf v$.

Let $(\bar{\mathbf h},\bar{\mathbf h}')$ be any feasible pair in these
canonical coordinates. Since
$\{\Psi_{\mathbf n}\}_{\mathbf n\in\mathbb N_0^d}$ is an orthonormal basis of
$L^2(\nu_d)$ by \autoref{cor:multi_hermite_orthogonality}, each component
admits an $L^2$ expansion
\[
\bar h_r(\mathbf u)
=
\sum_{\mathbf n\in\mathcal I}
a_{r,\mathbf n}\,\Psi_{\mathbf n}(\mathbf u),
\qquad
\bar h'_q(\mathbf v)
=
\sum_{\mathbf n\in\mathcal I}
b_{q,\mathbf n}\,\Psi_{\mathbf n}(\mathbf v),
\qquad
\mathcal I:=\mathbb N_0^d\setminus\{\mathbf 0\},
\]
for $r,q=1,\dots,d$. There is no $\mathbf n=\mathbf 0$ term because the maps
are centered. Let
\[
\mathbf A=(a_{r,\mathbf n})_{1\le r\le d,\;\mathbf n\in\mathcal I},
\qquad
\mathbf B=(b_{q,\mathbf n})_{1\le q\le d,\;\mathbf n\in\mathcal I}
\]
denote the corresponding coefficient matrices. Orthogonality of the basis and
the whitening constraints imply
\[
\mathbf A\mathbf A^\top=\mathbf I_d,
\qquad
\mathbf B\mathbf B^\top=\mathbf I_d.
\]

Next, by \autoref{prop:multi_variante_hermite},
\[
\mathbb E\!\left[\Psi_{\mathbf m}(\mathbf u)\Psi_{\mathbf n}(\mathbf v)\right]
=
\rho^{\mathbf n}\delta_{\mathbf m\mathbf n},
\qquad
\rho^{\mathbf n}:=\prod_{i=1}^d \rho_i^{n_i}.
\]
Hence the cross-covariance matrix of $(\bar{\mathbf h},\bar{\mathbf h}')$ is
\begin{equation}
\label{eq:cross-covariance-affine-id}
\mathbf C
:=
\mathrm{Cov}\!\big(\bar{\mathbf h}(\mathbf u),\bar{\mathbf h}'(\mathbf v)\big)
=
\mathbf A\,
\operatorname{diag}\!\big(\rho^{\mathbf n}\big)_{\mathbf n\in\mathcal I}\,
\mathbf B^\top.
\end{equation}

Now enumerate the multi-indices in $\mathcal I$ as
$\mathcal I=\{\mathbf n^{(1)},\mathbf n^{(2)},\dots\}$ so that
\[
d_j:=\rho^{\mathbf n^{(j)}}
\]
is nonincreasing. Every first-order multi-index $\mathbf e_i$ gives
\[
\rho^{\mathbf e_i}=\rho_i.
\]
On the other hand, if $|\mathbf n|:=\sum_{i=1}^d n_i\ge 2$, then
\[
\rho^{\mathbf n}
=
\prod_{i=1}^d \rho_i^{n_i}
\le \rho_1^{|\mathbf n|}
\le \rho_1^2
<
\rho_d
\]
by \autoref{assump:canonical-correlation-separation}. Therefore the top $d$
diagonal entries of \eqref{eq:cross-covariance-affine-id} are exactly
\[
\rho_1,\dots,\rho_d,
\]
corresponding to the first-order indices
$\mathbf e_1,\dots,\mathbf e_d$, and moreover there is a strict spectral gap
\[
d_d=\rho_d>d_{d+1}.
\]

Applying \autoref{lem:diag-pick-topN} with $N=d$ to
\eqref{eq:cross-covariance-affine-id} yields
\[
J_{\mathcal S}(\bar{\mathbf h},\bar{\mathbf h}')
=
\|\mathbf C\|_*
\le
\sum_{i=1}^d \rho_i.
\]
This upper bound is attained by the purely first-order maps
\[
\bar{\mathbf h}_{\mathrm{lin}}(\mathbf u)=\sqrt2\,\mathbf u,
\qquad
\bar{\mathbf h}'_{\mathrm{lin}}(\mathbf v)=\sqrt2\,\mathbf v,
\]
because these maps are centered and whitened, and their cross-covariance is
\[
\mathrm{Cov}(\sqrt2\,\mathbf u,\sqrt2\,\mathbf v)
=
\operatorname{diag}(\rho_1,\dots,\rho_d),
\]
whose singular values are exactly $\rho_1,\dots,\rho_d$. Hence
every population maximizer must satisfy
\[
\|\mathbf C\|_*=\sum_{i=1}^d \rho_i.
\]
By the strict-gap characterization in \autoref{col:diag-pick-topN}, this forces
all coefficients outside the first-order block to vanish. Thus there exist
orthogonal matrices $\mathbf A_1,\mathbf B_1\in O(d)$ such that
\[
\mathbf A=\big[\mathbf A_1\ \ 0\ \ 0\ \ \cdots\big],
\qquad
\mathbf B=\big[\mathbf B_1\ \ 0\ \ 0\ \ \cdots\big].
\]
Using \autoref{cor:multi_hermite_orthogonality},
\[
\Psi_{\mathbf e_i}(\mathbf u)=\sqrt2\,u_i,
\qquad
\Psi_{\mathbf e_i}(\mathbf v)=\sqrt2\,v_i.
\]
Therefore every maximizer has the form
\[
\bar{\mathbf h}^*(\mathbf u)=\mathbf A_1(\sqrt2\,\mathbf u),
\qquad
\bar{\mathbf h}'^*(\mathbf v)=\mathbf B_1(\sqrt2\,\mathbf v).
\]
Substituting the definitions of $\mathbf u$ and $\mathbf v$ gives
\[
\bar{\mathbf h}^*(\mathbf u)
=
\mathbf A_1\mathbf U_0^\top
\mathbf\Sigma_{\mathbf s\mathbf s}^{-1/2}
(\mathbf s-\bm\mu_{\mathbf s}),
\]
\[
\bar{\mathbf h}'^*(\mathbf v)
=
\mathbf B_1\mathbf V_0^\top
\mathbf\Sigma_{\mathbf s'\mathbf s'}^{-1/2}
(\mathbf s'-\bm\mu_{\mathbf s'}).
\]
Since $\mathbf A_1\mathbf U_0^\top$ and
$\mathbf B_1\mathbf V_0^\top$ are orthogonal, there exist
$\mathbf Q,\mathbf Q'\in O(d)$ such that
\[
\tilde{\mathbf h}^*(\mathbf s)
=
\mathbf Q\,
\mathbf\Sigma_{\mathbf s\mathbf s}^{-1/2}
(\mathbf s-\bm\mu_{\mathbf s}),
\qquad
\tilde{\mathbf h}'^{*}(\mathbf s')
=
\mathbf Q'\,
\mathbf\Sigma_{\mathbf s'\mathbf s'}^{-1/2}
(\mathbf s'-\bm\mu_{\mathbf s'}).
\]
Recalling that
\[
\tilde{\mathbf h}^*=\tilde{\mathbf f}^*\circ \mathbf g,
\qquad
\tilde{\mathbf h}'^{*}=\tilde{\mathbf f}'^{*}\circ \mathbf g',
\]
this proves the whitened statement in \autoref{thm:affine-id}. Finally, if
\[
\tilde{\mathbf h}^*
=
\mathbf W_{\mathbf h}
\big(\mathbf h^*-\mathbb E[\mathbf h^*]\big),
\qquad
\tilde{\mathbf h}'^{*}
=
\mathbf W_{\mathbf h'}
\big(\mathbf h'^*-\mathbb E[\mathbf h'^*]\big)
\]
are the corresponding population-whitened maps, then
\[
\mathbf h^*(\mathbf s)
=
\mathbf W_{\mathbf h}^{-1}\mathbf Q\,
\mathbf\Sigma_{\mathbf s\mathbf s}^{-1/2}
(\mathbf s-\bm\mu_{\mathbf s})
+
\mathbb E[\mathbf h^*],
\]
\[
\mathbf h'^*(\mathbf s')
=
\mathbf W_{\mathbf h'}^{-1}\mathbf Q'\,
\mathbf\Sigma_{\mathbf s'\mathbf s'}^{-1/2}
(\mathbf s'-\bm\mu_{\mathbf s'})
+
\mathbb E[\mathbf h'^*].
\]
Hence the unwhitened representation maps are affine, which proves the theorem.
\end{proof}

\paragraph{Why the strict gap is needed.}
The argument above uses the strict inequality
\[
\rho_d>\rho_1^2
\]
exactly once: it ensures that the first-order block
$\{\mathbf e_1,\dots,\mathbf e_d\}$ is separated from every higher-order
multi-index. If this fails, then a second-order term such as
$2\mathbf e_1$ can have coefficient $\rho_1^2\ge \rho_d$, and a population
maximizer need not be supported entirely on the first-order block. In that
regime, nonlinear CCA can mix linear and higher-order terms, and strict affine
identifiability is no longer guaranteed.
\begin{corollary}[Single-orbit structure of population maximizers under affine identifiability]
\label{cor:cca_single_orbit}
Assume the hypotheses of \autoref{thm:affine-id}, and let
\((\tilde{\mathbf f}^\star,\tilde{\mathbf f}^{\prime\star})\)
be any population maximizer of \autoref{equ:CCA_objective}.
Then the full set of population maximizers is exactly
\[
\Big\{
(\mathbf Q\tilde{\mathbf f}^\star,\mathbf Q'\tilde{\mathbf f}^{\prime\star})
:
\mathbf Q,\mathbf Q'\in O(d_{\mathcal Z})
\Big\}
\]
as a subset of
\(
L^2(P_{\mathbf x};\mathbb R^{d_{\mathcal Z}})
\times
L^2(P_{\mathbf x'};\mathbb R^{d_{\mathcal Z}}).
\)
In particular, under the assumptions of \autoref{thm:affine-id},
all population maximizers lie in a single
\(O(d_{\mathcal Z})\times O(d_{\mathcal Z})\)-orbit.
\end{corollary}

\begin{proof}
Let
\((\tilde{\mathbf f},\tilde{\mathbf f}')\)
be any other population maximizer of \autoref{equ:CCA_objective}.
By \autoref{thm:affine-id}, there exist orthogonal matrices
\(
\mathbf R_1,\mathbf R_2,\mathbf R_1',\mathbf R_2'
\in O(d_{\mathcal Z})
\)
such that
\[
(\tilde{\mathbf f}^\star\circ \mathbf g)(\mathbf s)
=
\mathbf R_1\,
\mathbf\Sigma_{\mathbf s\mathbf s}^{-1/2}
(\mathbf s-\bm\mu_{\mathbf s}),
\qquad
(\tilde{\mathbf f}\circ \mathbf g)(\mathbf s)
=
\mathbf R_2\,
\mathbf\Sigma_{\mathbf s\mathbf s}^{-1/2}
(\mathbf s-\bm\mu_{\mathbf s}),
\]
and similarly
\[
(\tilde{\mathbf f}^{\prime\star}\circ \mathbf g')(\mathbf s')
=
\mathbf R_1'\,
\mathbf\Sigma_{\mathbf s'\mathbf s'}^{-1/2}
(\mathbf s'-\bm\mu_{\mathbf s'}),
\qquad
(\tilde{\mathbf f}'\circ \mathbf g')(\mathbf s')
=
\mathbf R_2'\,
\mathbf\Sigma_{\mathbf s'\mathbf s'}^{-1/2}
(\mathbf s'-\bm\mu_{\mathbf s'}).
\]

Define
\[
\mathbf Q:=\mathbf R_2\mathbf R_1^\top\in O(d_{\mathcal Z}),
\qquad
\mathbf Q':=\mathbf R_2'\mathbf R_1'^\top\in O(d_{\mathcal Z}).
\]
Then
\[
(\tilde{\mathbf f}\circ \mathbf g)(\mathbf s)
=
\mathbf Q\,(\tilde{\mathbf f}^\star\circ \mathbf g)(\mathbf s),
\qquad
(\tilde{\mathbf f}'\circ \mathbf g')(\mathbf s')
=
\mathbf Q'\,(\tilde{\mathbf f}^{\prime\star}\circ \mathbf g')(\mathbf s').
\]

By the pullback isometry from \autoref{lem:pushforward}.1,
\[
\|\tilde{\mathbf f}-\mathbf Q\tilde{\mathbf f}^\star\|_{L^2(P_{\mathbf x})}
=
\|(\tilde{\mathbf f}-\mathbf Q\tilde{\mathbf f}^\star)\circ \mathbf g\|_{L^2(P_{\mathbf s})}
=
0,
\]
and likewise
\[
\|\tilde{\mathbf f}'-\mathbf Q'\tilde{\mathbf f}^{\prime\star}\|_{L^2(P_{\mathbf x'})}
=
\|(\tilde{\mathbf f}'-\mathbf Q'\tilde{\mathbf f}^{\prime\star})\circ \mathbf g'\|_{L^2(P_{\mathbf s'})}
=
0.
\]
Thus every population maximizer lies in the
\(O(d_{\mathcal Z})\times O(d_{\mathcal Z})\)-orbit of
\((\tilde{\mathbf f}^\star,\tilde{\mathbf f}^{\prime\star})\).

The reverse inclusion is exactly \autoref{cor:cca_orbit_maximizers}.
\end{proof}

\subsection{Preservation of Higher-Order Nonlinearities Beyond First-Order Dominance}
\label{app:higher_order_nonlinearities}

In the main text, \autoref{assump:canonical-correlation-separation} requires first-order canonical dominance, specifically $\rho_{d_{\mathcal{S}}}>\rho_1^2$, to guarantee strictly affine identifiability. This section provides a self-contained illustration of the representation learning dynamics when this assumption is relaxed. Specifically, we demonstrate how the canonical spectrum interacts with the representation capacity $d_{\mathcal{Z}}$ to determine the survival of higher-order nonlinearities.

\paragraph{Theoretical Mechanism.}
As established in the proof of \autoref{thm:affine-id}, the normalized orthogonal polynomial expansion (e.g., the Mehler expansion for Gaussian priors) diagonalizes the CCA objective. Under this expansion, a multivariate polynomial term of multi-index degree $\mathbf{n}=(n_1,\dots,n_{d_{\mathcal{S}}})$ corresponds to a canonical correlation given by $\prod_{i=1}^{d_{\mathcal{S}}}\rho_i^{n_i}$. Because nonlinear CCA strictly prioritizes orthogonal components that maximize the sum of inter-view correlations, it greedily selects the $d_{\mathcal{Z}}$ terms with the largest corresponding values of $\prod_{i=1}^{d_{\mathcal{S}}}\rho_i^{n_i}$, regardless of their polynomial degree. Consequently, higher-order nonlinear terms of strongly correlated source components can overshadow the first-order (linear) terms of more weakly correlated components.

\paragraph{Infinite Capacity Guarantee.}
Because the true canonical correlations satisfy $0<\rho_i<1$ for all $i\in\{1,\dots,d_{\mathcal{S}}\}$, the correlation of any higher-order term diminishes exponentially as its total polynomial degree $|\mathbf{n}|\to\infty$. Therefore, while a limited capacity $d_{\mathcal{Z}}$ can lead to the truncation of weakly correlated linear terms, the asymptotic limit of infinite representation capacity ($d_{\mathcal{Z}}\to\infty$) guarantees the eventual recovery of all first-order linear terms. However, in this unconstrained regime, the desired linear features are inherently entangled with a multitude of higher-order artifacts.

\paragraph{A Concrete Example.}
To rigorously illustrate this phenomenon, consider a two-dimensional latent space ($d_{\mathcal{S}}=2$) with ground-truth canonical correlations $\rho_1=0.9$ and $\rho_2=0.5$. Notice that this violates the first-order dominance condition since $\rho_2<\rho_1^2$ ($0.5<0.81$).

Evaluating the canonical correlations for the lowest-degree polynomial terms yields:
\begin{itemize}
    \item First-order terms: $s_1$ yields $\rho_1=0.9$, and $s_2$ yields $\rho_2=0.5$.
    \item Higher-order terms of $s_1$: $s_1^2$ yields $\rho_1^2=0.81$, $s_1^3$ yields
    $\rho_1^3=0.729$, $s_1^4$ yields $\rho_1^4=0.6561$, $s_1^5$ yields $\rho_1^5=0.59049$
    and $s_1^6$ yields $\rho_1^6=0.531441$.
\end{itemize}

Sorting these available orthogonal terms in descending order of their canonical correlations generates the sequence: $\{0.9, 0.81, 0.729, 0.6561, 0.59049, 0.531441,0.5, \dots\}$.

If we constrain the representation capacity to match the source dimension ($d_{\mathcal{Z}}=2$), nonlinear CCA greedily selects the top two components. In this case, it captures $\{s_1,s_1^2\}$, introducing a second-order nonlinearity of $s_1$ and entirely failing to recover the linear term $s_2$.

If the representation capacity is expanded to $d_{\mathcal{Z}}=7$, the
algorithm selects the top seven components:
$\{s_1, s_1^2, s_1^3, s_1^4, s_1^5, s_1^6, s_2\}$. Here, the linear
term $s_2$ is successfully recovered, consistent with the infinite capacity
guarantee, but the learned representation is now dominated by higher-order
nonlinearities up to the sixth degree. This example clearly demonstrates
why the strict spectral gap $\rho_{d_{\mathcal{S}}}>\rho_1^2$ is
mathematically necessary to isolate purely affine representations under finite capacity.
\subsection{Proof of \autoref{thm:emp-consistency}}
\label{app:proof-emp-consistency}

We write $\|\cdot\|$ for the spectral norm and $\|\cdot\|_*$ for the nuclear norm.

\begin{lemma}[Stability of ridge whitening near the identity]
\label{lem:ridge-whitening-identity}
Let $\mathbf A\in\mathbb R^{d\times d}$ be symmetric positive semidefinite and
assume $\|\mathbf A-\mathbf I_d\|\le \frac12$. Then, for every $\epsilon\in[0,1]$,
\[
\big\|(\mathbf A+\epsilon \mathbf I_d)^{-1/2}-\mathbf I_d\big\|
\;\le\;
C\,\|\mathbf A-\mathbf I_d\|
+
\big|(1+\epsilon)^{-1/2}-1\big|,
\]
where $C>0$ is an absolute constant. In particular,
\[
\big\|(\mathbf A+\epsilon \mathbf I_d)^{-1/2}-\mathbf I_d\big\|
=
O\!\big(\|\mathbf A-\mathbf I_d\|+\epsilon\big).
\]
\end{lemma}

\begin{proof}
Since $\mathbf A$ is symmetric, it admits an eigendecomposition
$\mathbf A=\mathbf U\operatorname{diag}(\lambda_1,\dots,\lambda_d)\mathbf U^\top$.
The condition $\|\mathbf A-\mathbf I_d\|\le \frac12$ implies
$\lambda_i\in[\frac12,\frac32]$ for all $i$.
Therefore
\[
(\mathbf A+\epsilon\mathbf I_d)^{-1/2}
=
\mathbf U\operatorname{diag}\!\big((\lambda_i+\epsilon)^{-1/2}\big)_{i=1}^d\mathbf U^\top,
\]
and hence
\[
\big\|(\mathbf A+\epsilon\mathbf I_d)^{-1/2}
-(1+\epsilon)^{-1/2}\mathbf I_d\big\|
=
\max_{1\le i\le d}
\big|(\lambda_i+\epsilon)^{-1/2}-(1+\epsilon)^{-1/2}\big|.
\]
Now apply the mean value theorem to the scalar map
$t\mapsto (t+\epsilon)^{-1/2}$ on $[\frac12,\frac32]$:
\[
\big|(\lambda_i+\epsilon)^{-1/2}-(1+\epsilon)^{-1/2}\big|
\le
\sup_{t\in[\frac12,\frac32],\,\epsilon\in[0,1]}
\frac{1}{2(t+\epsilon)^{3/2}}
\,|\lambda_i-1|
\le
C\,|\lambda_i-1|.
\]
Taking the maximum over $i$ gives
\[
\big\|(\mathbf A+\epsilon\mathbf I_d)^{-1/2}
-(1+\epsilon)^{-1/2}\mathbf I_d\big\|
\le
C\,\|\mathbf A-\mathbf I_d\|.
\]
Finally,
\[
\big\|(1+\epsilon)^{-1/2}\mathbf I_d-\mathbf I_d\big\|
=
\big|(1+\epsilon)^{-1/2}-1\big|,
\]
and the claim follows by the triangle inequality.
\end{proof}

\begin{proof}
For each $n$, let
\[
(\hat{\mathbf f}_n,\hat{\mathbf f}'_n)
:=
(\tilde{\mathbf f}_{\bm\theta_n},\tilde{\mathbf f}'_{\bm\theta_n'})
\]
be any empirical $\delta_n$-maximizer from Assumption~A4.
Fix a feasible pair
\[
(\tilde{\mathbf f},\tilde{\mathbf f}')
\in
\tilde{\mathcal F}_{\mathcal X}\times \tilde{\mathcal F}'_{\mathcal X'}.
\]
Write
\[
\mathbf z:=\tilde{\mathbf f}(\mathbf x),
\qquad
\mathbf z':=\tilde{\mathbf f}'(\mathbf x').
\]
Since the feasible classes in \autoref{assump:function-class} are whitened,
\[
\mathbf\Sigma_{\mathbf z\mathbf z}=\mathbf I_{d_{\mathcal Z}},
\qquad
\mathbf\Sigma_{\mathbf z'\mathbf z'}=\mathbf I_{d_{\mathcal Z}}.
\]
Hence the population normalized cross-covariance simplifies to
\[
\mathbf K
=
\mathbf\Sigma_{\mathbf z\mathbf z'},
\qquad
J(\tilde{\mathbf f},\tilde{\mathbf f}')
=
\|\mathbf K\|_*.
\]
For the empirical quantities, define
\[
\hat{\mathbf W}_{\mathbf z}
:=
(\hat{\mathbf\Sigma}_{\mathbf z\mathbf z}+\epsilon\mathbf I)^{-1/2},
\qquad
\hat{\mathbf W}_{\mathbf z'}
:=
(\hat{\mathbf\Sigma}_{\mathbf z'\mathbf z'}+\epsilon\mathbf I)^{-1/2},
\]
and
\[
\hat{\mathbf K}
:=
\hat{\mathbf W}_{\mathbf z}\,
\hat{\mathbf\Sigma}_{\mathbf z\mathbf z'}\,
\hat{\mathbf W}_{\mathbf z'}.
\]
Then
\[
\hat J(\tilde{\mathbf f},\tilde{\mathbf f}')
=
\|\hat{\mathbf K}\|_*.
\]

\paragraph{Step 1: stability of empirical whitening.}
Let
\[
\Delta_n
:=
\sup_{\tilde{\mathbf f}\in\tilde{\mathcal F}_{\mathcal X},\,
      \tilde{\mathbf f}'\in\tilde{\mathcal F}'_{\mathcal X'}}
\Big(
\|\hat{\mathbf\Sigma}_{\mathbf z\mathbf z}-\mathbf I\|
\vee
\|\hat{\mathbf\Sigma}_{\mathbf z'\mathbf z'}-\mathbf I\|
\vee
\|\hat{\mathbf\Sigma}_{\mathbf z\mathbf z'}-\mathbf\Sigma_{\mathbf z\mathbf z'}\|
\Big).
\]
By Assumption~A3,
\[
\Delta_n=o_p(\epsilon),
\]
hence also $\Delta_n=o_p(1)$ because $\epsilon\to 0$.

On the event $\{\Delta_n\le \frac12\}$, every eigenvalue of
$\hat{\mathbf\Sigma}_{\mathbf z\mathbf z}$ and
$\hat{\mathbf\Sigma}_{\mathbf z'\mathbf z'}$ lies in
$[\frac12,\frac32]$, uniformly over the feasible classes.
Therefore \autoref{lem:ridge-whitening-identity} yields
\[
\sup_{\tilde{\mathbf f},\tilde{\mathbf f}'}
\big\|\hat{\mathbf W}_{\mathbf z}-\mathbf I\big\|
\le
C\sup_{\tilde{\mathbf f},\tilde{\mathbf f}'}
\big\|\hat{\mathbf\Sigma}_{\mathbf z\mathbf z}-\mathbf I\big\|
+
\big|(1+\epsilon)^{-1/2}-1\big|
=
o_p(\epsilon)+O(\epsilon)
=
o_p(1),
\]
and similarly
\[
\sup_{\tilde{\mathbf f},\tilde{\mathbf f}'}
\big\|\hat{\mathbf W}_{\mathbf z'}-\mathbf I\big\|
=
o_p(1).
\]
In particular,
\[
\sup_{\tilde{\mathbf f},\tilde{\mathbf f}'}
\big\|\hat{\mathbf W}_{\mathbf z}\big\|
+
\sup_{\tilde{\mathbf f},\tilde{\mathbf f}'}
\big\|\hat{\mathbf W}_{\mathbf z'}\big\|
=
O_p(1).
\]

\paragraph{Step 2: uniform convergence of the normalized cross-covariance.}
For every feasible pair, $\mathbf z$ and $\mathbf z'$ are whitened. Hence
\[
\|\mathbf\Sigma_{\mathbf z\mathbf z'}\|\le 1.
\]
Indeed, for arbitrary unit vectors $\mathbf u,\mathbf v\in\mathbb R^{d_{\mathcal Z}}$,
\[
\big|\mathbf u^\top \mathbf\Sigma_{\mathbf z\mathbf z'}\mathbf v\big|
=
\big|\mathrm{Cov}(\mathbf u^\top\mathbf z,\mathbf v^\top\mathbf z')\big|
\le
\sqrt{\mathrm{Var}(\mathbf u^\top\mathbf z)\,\mathrm{Var}(\mathbf v^\top\mathbf z')}
=
1,
\]
so the operator norm is at most $1$.
Therefore, on the event $\{\Delta_n\le \frac12\}$,
\[
\sup_{\tilde{\mathbf f},\tilde{\mathbf f}'}
\|\hat{\mathbf\Sigma}_{\mathbf z\mathbf z'}\|
\le
\sup_{\tilde{\mathbf f},\tilde{\mathbf f}'}
\|\mathbf\Sigma_{\mathbf z\mathbf z'}\|
+
\Delta_n
\le
1+\Delta_n
\le
\frac32.
\]

Now expand
\begin{align*}
\hat{\mathbf K}-\mathbf K
&=
\hat{\mathbf W}_{\mathbf z}\,
\hat{\mathbf\Sigma}_{\mathbf z\mathbf z'}\,
\hat{\mathbf W}_{\mathbf z'}
-
\mathbf\Sigma_{\mathbf z\mathbf z'} \\
&=
(\hat{\mathbf W}_{\mathbf z}-\mathbf I)\,
\hat{\mathbf\Sigma}_{\mathbf z\mathbf z'}\,
\hat{\mathbf W}_{\mathbf z'}
+
(\hat{\mathbf\Sigma}_{\mathbf z\mathbf z'}-\mathbf\Sigma_{\mathbf z\mathbf z'})\,
\hat{\mathbf W}_{\mathbf z'}
+
\mathbf\Sigma_{\mathbf z\mathbf z'}\,
(\hat{\mathbf W}_{\mathbf z'}-\mathbf I).
\end{align*}
Taking suprema over the feasible classes and using Step~1 gives
\begin{align*}
\sup_{\tilde{\mathbf f},\tilde{\mathbf f}'}
\|\hat{\mathbf K}-\mathbf K\|
&\le
\sup_{\tilde{\mathbf f},\tilde{\mathbf f}'}
\|\hat{\mathbf W}_{\mathbf z}-\mathbf I\|\,
\sup_{\tilde{\mathbf f},\tilde{\mathbf f}'}
\|\hat{\mathbf\Sigma}_{\mathbf z\mathbf z'}\|\,
\sup_{\tilde{\mathbf f},\tilde{\mathbf f}'}
\|\hat{\mathbf W}_{\mathbf z'}\| \\
&\quad
+
\sup_{\tilde{\mathbf f},\tilde{\mathbf f}'}
\|\hat{\mathbf\Sigma}_{\mathbf z\mathbf z'}-\mathbf\Sigma_{\mathbf z\mathbf z'}\|\,
\sup_{\tilde{\mathbf f},\tilde{\mathbf f}'}
\|\hat{\mathbf W}_{\mathbf z'}\| \\
&\quad
+
\sup_{\tilde{\mathbf f},\tilde{\mathbf f}'}
\|\mathbf\Sigma_{\mathbf z\mathbf z'}\|\,
\sup_{\tilde{\mathbf f},\tilde{\mathbf f}'}
\|\hat{\mathbf W}_{\mathbf z'}-\mathbf I\| \\
&=
o_p(1).
\end{align*}

\paragraph{Step 3: objective consistency.}
Since the population and empirical objectives are the nuclear norms of
$\mathbf K$ and $\hat{\mathbf K}$, respectively,
\[
\big|\hat J(\tilde{\mathbf f},\tilde{\mathbf f}')
-
J(\tilde{\mathbf f},\tilde{\mathbf f}')\big|
=
\big|\|\hat{\mathbf K}\|_*-\|\mathbf K\|_*\big|
\le
\|\hat{\mathbf K}-\mathbf K\|_*.
\]
Using $\|\mathbf A\|_*\le d_{\mathcal Z}\|\mathbf A\|$ for
$\mathbf A\in\mathbb R^{d_{\mathcal Z}\times d_{\mathcal Z}}$, we obtain
\[
\sup_{\tilde{\mathbf f}\in\tilde{\mathcal F}_{\mathcal X},\,
      \tilde{\mathbf f}'\in\tilde{\mathcal F}'_{\mathcal X'}}
\big|\hat J(\tilde{\mathbf f},\tilde{\mathbf f}')
-
J(\tilde{\mathbf f},\tilde{\mathbf f}')\big|
\le
d_{\mathcal Z}
\sup_{\tilde{\mathbf f},\tilde{\mathbf f}'}
\|\hat{\mathbf K}-\mathbf K\|
\xrightarrow{\mathbb P} 0.
\]
This proves Claim~1.

\paragraph{Step 4: estimator consistency up to orthogonal transformations.}
Let
\[
(\tilde{\mathbf f}^\star,\tilde{\mathbf f}^{\prime\star})
\in
\tilde{\mathcal F}_{\mathcal X}\times\tilde{\mathcal F}'_{\mathcal X'}
\]
be any population maximizer, and write
\[
J^\star
:=
J(\tilde{\mathbf f}^\star,\tilde{\mathbf f}^{\prime\star}).
\]
Define its orbit
\[
\mathcal O^\star
:=
\big\{
(\mathbf Q\tilde{\mathbf f}^\star,\mathbf Q'\tilde{\mathbf f}^{\prime\star})
:\ \mathbf Q,\mathbf Q'\in O(d_{\mathcal Z})
\big\}.
\]
We equip the quotient by this orthogonal action with the distance
\[
d_{\mathcal Q}\!\big((\tilde{\mathbf f},\tilde{\mathbf f}'),\mathcal O^\star\big)
:=
\inf_{\mathbf Q,\mathbf Q'\in O(d_{\mathcal Z})}
\Big(
\|\mathbf Q\tilde{\mathbf f}-\tilde{\mathbf f}^\star\|_{L^2(P_{\mathbf x})}
+
\|\mathbf Q'\tilde{\mathbf f}'-\tilde{\mathbf f}^{\prime\star}\|_{L^2(P_{\mathbf x'})}
\Big).
\]

By Assumption~A4,
\[
\hat J(\hat{\mathbf f}_n,\hat{\mathbf f}'_n)
\ge
\hat J(\tilde{\mathbf f}^\star,\tilde{\mathbf f}^{\prime\star})
-\delta_n.
\]
Therefore
\begin{align*}
0
\le
J^\star-J(\hat{\mathbf f}_n,\hat{\mathbf f}'_n)
&\le
\big|
J(\tilde{\mathbf f}^\star,\tilde{\mathbf f}^{\prime\star})
-
\hat J(\tilde{\mathbf f}^\star,\tilde{\mathbf f}^{\prime\star})
\big| \\
&\quad
+
\big|
\hat J(\hat{\mathbf f}_n,\hat{\mathbf f}'_n)
-
J(\hat{\mathbf f}_n,\hat{\mathbf f}'_n)
\big|
+
\delta_n \\
&\le
2\sup_{\tilde{\mathbf f},\tilde{\mathbf f}'}
\big|\hat J(\tilde{\mathbf f},\tilde{\mathbf f}')-
J(\tilde{\mathbf f},\tilde{\mathbf f}')\big|
+
\delta_n
\xrightarrow{\mathbb P} 0,
\end{align*}
where the last step uses Claim~1 and $\delta_n\to 0$.

We now invoke the orbit-separation part of Assumption~A1. If
\[
d_{\mathcal Q}\!\big((\hat{\mathbf f}_n,\hat{\mathbf f}'_n),\mathcal O^\star\big)
\not\xrightarrow{\mathbb P} 0,
\]
then there exist $\eta,\alpha>0$ and a subsequence (not relabeled) such that
\[
\mathbb P\!\left(
d_{\mathcal Q}\!\big((\hat{\mathbf f}_n,\hat{\mathbf f}'_n),\mathcal O^\star\big)
\ge \eta
\right)\ge \alpha
\qquad\text{for all }n.
\]
On this event, Assumption~A1 gives
\[
J^\star-J(\hat{\mathbf f}_n,\hat{\mathbf f}'_n)\ge \kappa(\eta)>0,
\]
which contradicts
\[
J^\star-J(\hat{\mathbf f}_n,\hat{\mathbf f}'_n)\xrightarrow{\mathbb P} 0.
\]
Hence
\[
d_{\mathcal Q}\!\big((\hat{\mathbf f}_n,\hat{\mathbf f}'_n),\mathcal O^\star\big)
\xrightarrow{\mathbb P} 0.
\]
Since $O(d_{\mathcal Z})\times O(d_{\mathcal Z})$ is compact and the distance
functional is continuous, the infimum in $d_{\mathcal Q}$ is attained.
Equivalently, there exist random orthogonal matrices
$\mathbf Q_n,\mathbf Q_n'\in O(d_{\mathcal Z})$ such that
\[
\|\mathbf Q_n\hat{\mathbf f}_n-\tilde{\mathbf f}^\star\|_{L^2(P_{\mathbf x})}
+
\|\mathbf Q_n'\hat{\mathbf f}_n'-\tilde{\mathbf f}^{\prime\star}\|_{L^2(P_{\mathbf x'})}
\xrightarrow{\mathbb P} 0.
\]
This proves Claim~2.

\paragraph{Step 5: latent recovery.}
Let
\[
\tilde{\mathbf h}^\star
:=
\tilde{\mathbf f}^\star\circ\mathbf g,
\qquad
\tilde{\mathbf h}^{\prime\star}
:=
\tilde{\mathbf f}^{\prime\star}\circ\mathbf g'.
\]
By the pullback isometry established in the proof of \autoref{prop:rep-inv},
for every $\mathbf Q,\mathbf Q'\in O(d_{\mathcal Z})$,
\[
\|\mathbf Q(\hat{\mathbf f}_n\circ \mathbf g)-\tilde{\mathbf h}^\star\|_{L^2(P_{\mathbf s})}
=
\|\mathbf Q\hat{\mathbf f}_n-\tilde{\mathbf f}^\star\|_{L^2(P_{\mathbf x})},
\]
and likewise
\[
\|\mathbf Q'(\hat{\mathbf f}_n'\circ \mathbf g')-\tilde{\mathbf h}^{\prime\star}\|_{L^2(P_{\mathbf s'})}
=
\|\mathbf Q'\hat{\mathbf f}_n'-\tilde{\mathbf f}^{\prime\star}\|_{L^2(P_{\mathbf x'})}.
\]
Therefore Claim~2 immediately implies
\begin{align*}
\inf_{\mathbf Q,\mathbf Q'\in O(d_{\mathcal Z})}
\Big(
&
\|\mathbf Q(\hat{\mathbf f}_n\circ \mathbf g)-\tilde{\mathbf h}^\star\|_{L^2(P_{\mathbf s})}
\\
&\qquad\qquad
+
\|\mathbf Q'(\hat{\mathbf f}_n'\circ \mathbf g')-\tilde{\mathbf h}^{\prime\star}\|_{L^2(P_{\mathbf s'})}
\Big)
\xrightarrow{\mathbb P} 0.
\end{align*}
Finally, by \autoref{thm:affine-id},
$\tilde{\mathbf h}^\star$ and $\tilde{\mathbf h}^{\prime\star}$
are the marginally whitened latent factors up to orthogonal transformations.
This proves Claim~3.
\end{proof}
\section{Necessity of whitening}
\label{app:whitening}

For precision, the necessity claim has to be stated carefully. Centering and unit covariance play different roles, and only the latter is the genuinely restrictive whitening condition. What is mathematically indispensable is a scale-fixing normalization equivalent to whitening. In our formulation this normalization is imposed by restricting the encoder classes to zero-mean, identity-covariance maps. The proof of \autoref{thm:affine-id} uses this normalization in two logically distinct ways: centering removes the constant polynomial mode, while unit covariance yields the orthonormal coefficient geometry required by the diagonal selection argument.

We first record that normalized CCA can always be represented on whitened encoders.

\begin{proposition}[Whitening is without loss of generality for normalized CCA]
\label{prop:whitening-wlog}
Let
\(
\mathbf h \in L^2(P_{\mathbf s};\mathbb R^{d_{\mathcal Z}})
\)
and
\(
\mathbf h' \in L^2(P_{\mathbf s'};\mathbb R^{d_{\mathcal Z}})
\)
be centered maps with
\[
\mathbf{\Sigma}_{\mathbf h}:=\mathrm{Cov}\big(\mathbf h(\mathbf s)\big)\succ 0,
\qquad
\mathbf{\Sigma}_{\mathbf h'}:=\mathrm{Cov}\big(\mathbf h'(\mathbf s')\big)\succ 0.
\]
Define their whitened versions by
\[
\tilde{\mathbf h}:=\mathbf{\Sigma}_{\mathbf h}^{-1/2}\mathbf h,
\qquad
\tilde{\mathbf h}':=\mathbf{\Sigma}_{\mathbf h'}^{-1/2}\mathbf h'.
\]
If
\[
J_{\mathcal S}^{\mathrm{cca}}(\mathbf h,\mathbf h')
:=
\sum_{i=1}^{d_{\mathcal Z}}
\sigma_i\!\Big(
\mathbf{\Sigma}_{\mathbf h}^{-1/2}
\mathrm{Cov}\big(\mathbf h(\mathbf s),\mathbf h'(\mathbf s')\big)
\mathbf{\Sigma}_{\mathbf h'}^{-1/2}
\Big),
\]
then
\[
J_{\mathcal S}^{\mathrm{cca}}(\mathbf h,\mathbf h')
=
J_{\mathcal S}(\tilde{\mathbf h},\tilde{\mathbf h}'),
\]
where \(J_{\mathcal S}\) is the whitened source-space objective in \autoref{equ:CCA_objective_source}. In particular, every candidate pair for normalized CCA admits an equivalent representation by whitened encoders.
\end{proposition}

\begin{proof}
Since \(\mathbf h\) and \(\mathbf h'\) are centered,
\[
\mathrm{Cov}\big(\tilde{\mathbf h}(\mathbf s)\big)
=
\mathbf{\Sigma}_{\mathbf h}^{-1/2}
\mathbf{\Sigma}_{\mathbf h}
\mathbf{\Sigma}_{\mathbf h}^{-1/2}
=
\mathbf I_{d_{\mathcal Z}},
\qquad
\mathrm{Cov}\big(\tilde{\mathbf h}'(\mathbf s')\big)
=
\mathbf{\Sigma}_{\mathbf h'}^{-1/2}
\mathbf{\Sigma}_{\mathbf h'}
\mathbf{\Sigma}_{\mathbf h'}^{-1/2}
=
\mathbf I_{d_{\mathcal Z}}.
\]
Hence \(\tilde{\mathbf h}\in\hat{\mathcal F}_{\mathcal S}\) and
\(\tilde{\mathbf h}'\in\hat{\mathcal F}'_{\mathcal S}\). Moreover,
\[
\mathrm{Cov}\big(\tilde{\mathbf h}(\mathbf s),\tilde{\mathbf h}'(\mathbf s')\big)
=
\mathbf{\Sigma}_{\mathbf h}^{-1/2}
\mathrm{Cov}\big(\mathbf h(\mathbf s),\mathbf h'(\mathbf s')\big)
\mathbf{\Sigma}_{\mathbf h'}^{-1/2}.
\]
Taking singular values and summing them proves the identity.
\end{proof}

The next proposition shows why some such normalization is necessary.

\begin{proposition}[Without scale normalization the covariance objective is ill-posed]
\label{prop:raw-cca-unbounded}
Define the unnormalized objective
\[
J_{\mathcal S}^{\mathrm{raw}}(\mathbf h,\mathbf h')
:=
\sum_{i=1}^{d_{\mathcal Z}}
\sigma_i\!\Big(
\mathrm{Cov}\big(\mathbf h(\mathbf s),\mathbf h'(\mathbf s')\big)
\Big)
\]
over centered square-integrable encoder pairs. If there exists one pair
\((\mathbf h,\mathbf h')\) such that
\(J_{\mathcal S}^{\mathrm{raw}}(\mathbf h,\mathbf h')>0\),
then
\[
\sup_{\mathbf h,\mathbf h'} J_{\mathcal S}^{\mathrm{raw}}(\mathbf h,\mathbf h')
=+\infty.
\]
\end{proposition}

\begin{proof}
For any \(\lambda>0\),
\[
\mathrm{Cov}\big(\lambda\mathbf h(\mathbf s),\lambda\mathbf h'(\mathbf s')\big)
=
\lambda^2\,
\mathrm{Cov}\big(\mathbf h(\mathbf s),\mathbf h'(\mathbf s')\big),
\]
so
\[
J_{\mathcal S}^{\mathrm{raw}}(\lambda\mathbf h,\lambda\mathbf h')
=
\lambda^2 J_{\mathcal S}^{\mathrm{raw}}(\mathbf h,\mathbf h').
\]
If \(J_{\mathcal S}^{\mathrm{raw}}(\mathbf h,\mathbf h')>0\), letting
\(\lambda\to\infty\) proves the claim. Under
\autoref{assump:gaussian-latent-pair}, this hypothesis holds by taking the first
pair of canonical variates and, if \(d_{\mathcal Z}>1\), padding the remaining
coordinates with zeros.
\end{proof}

The preceding propositions isolate the formal necessity of whitening. In the proof of \autoref{thm:affine-id}, whitening enters at three specific places.

\textbf{1. Basis alignment and exclusion of the constant mode.}\
The Hermite-Mehler expansions in
\autoref{lemma:hermite_expansion_bivariate_gaussian} and
\autoref{prop:multi_variante_hermite} are written in the standardized source
coordinates introduced there. This basis alignment is determined by the latent
Gaussian law after source standardization and is independent of the encoder
parameterization. For the encoder expansions, the only degree-zero condition
needed is centering. Indeed, if
\[
\tilde{\mathbf h}_r(\mathbf s)
=
\sum_{\mathbf n\in\mathbb N^{d_{\mathcal S}}}
\alpha_{r,\mathbf n}\Psi_{\mathbf n}(\mathbf u),
\qquad
\tilde{\mathbf h}'_q(\mathbf s')
=
\sum_{\mathbf n\in\mathbb N^{d_{\mathcal S}}}
\beta_{q,\mathbf n}\Psi_{\mathbf n}(\mathbf v),
\]
then \(\Psi_{\mathbf 0}\) is the unique constant basis element and
\(\mathbb E[\Psi_{\mathbf n}]=0\) for every \(\mathbf n\neq \mathbf 0\). Hence
\[
\mathbb E[\tilde{\mathbf h}_r(\mathbf s)]=0
\iff
\alpha_{r,\mathbf 0}=0,
\qquad
\mathbb E[\tilde{\mathbf h}'_q(\mathbf s')]=0
\iff
\beta_{q,\mathbf 0}=0.
\]
Thus the constant mode disappears because the encoders are centered. The
unit-covariance part of whitening is used later.

\textbf{2. Feasible-set geometry for the CCA objective.}\
After applying \autoref{prop:whitening-wlog}, we may work entirely with
whitened encoders. Writing
\[
\tilde{\mathbf h}_r(\mathbf s)
=
\sum_{\mathbf n\neq \mathbf 0}\alpha_{r,\mathbf n}\Psi_{\mathbf n}(\mathbf u),
\qquad
\tilde{\mathbf h}'_q(\mathbf s')
=
\sum_{\mathbf n\neq \mathbf 0}\beta_{q,\mathbf n}\Psi_{\mathbf n}(\mathbf v),
\]
and collecting coefficients into matrices
\(
\mathbf A=(\alpha_{r,\mathbf n})
\)
and
\(
\mathbf B=(\beta_{q,\mathbf n}),
\)
Corollary~\ref{cor:multi_hermite_orthogonality} gives
\[
\mathrm{Cov}\big(\tilde{\mathbf h}(\mathbf s)\big)=\mathbf A\mathbf A^\top,
\qquad
\mathrm{Cov}\big(\tilde{\mathbf h}'(\mathbf s')\big)=\mathbf B\mathbf B^\top.
\]
Hence whitening is exactly the constraint
\[
\mathbf A\mathbf A^\top=\mathbf I_{d_{\mathcal Z}},
\qquad
\mathbf B\mathbf B^\top=\mathbf I_{d_{\mathcal Z}}.
\]
Under these constraints, the cross-covariance diagonalizes as
\[
\mathrm{Cov}\big(\tilde{\mathbf h}(\mathbf s),\tilde{\mathbf h}'(\mathbf s')\big)
=
c\,\mathbf A\,\mathrm{diag}(t_{\mathbf n})\,\mathbf B^\top,
\]
and \autoref{lem:diag-pick-topN} applies directly. Without the unit-covariance
constraints, the coefficient matrices are no longer row-orthonormal, and the
proof no longer reduces to selecting the largest diagonal entries
\(\{|t_{\mathbf n}|\}\). This is the precise point where whitening enters the
identifiability argument.

\textbf{3. Scale fixing and collapse avoidance.}\
\autoref{prop:raw-cca-unbounded} already shows that without any scale-fixing
normalization the raw covariance objective is unbounded. After passing to the
whitened representation, the remaining first-order block is also
well-conditioned. Once \autoref{assump:canonical-correlation-separation}
excludes all higher-order indices, the optimizer is confined to the first-order
block \(\{\mathbf e_1,\dots,\mathbf e_{d_{\mathcal S}}\}\). Let
\(\mathbf A_{\mathrm{lin}}\) and \(\mathbf B_{\mathrm{lin}}\) denote the
corresponding \(d_{\mathcal Z}\times d_{\mathcal S}\) coefficient matrices.
Because the full coefficient matrices are whitened, we still have
\[
\mathbf A_{\mathrm{lin}}\mathbf A_{\mathrm{lin}}^\top
=
\mathbf I_{d_{\mathcal Z}},
\qquad
\mathbf B_{\mathrm{lin}}\mathbf B_{\mathrm{lin}}^\top
=
\mathbf I_{d_{\mathcal Z}}.
\]
When \(d_{\mathcal Z}=d_{\mathcal S}\), this forces
\(
\mathbf A_{\mathrm{lin}},\mathbf B_{\mathrm{lin}}\in O(d_{\mathcal Z})
\).
Therefore the recovered linear maps are full-rank and perfectly conditioned in
the whitened coordinates: no output coordinate can vanish, duplicate another
one, or absorb an arbitrary scale. This is exactly what yields the orthogonal
ambiguities in \autoref{thm:affine-id}. Without whitening, even after
higher-order terms are removed, the remaining linear coefficient matrices need
only be invertible (if at all) and may be arbitrarily ill-conditioned.

\begin{remark}
The discussion above yields the precise form of the necessity claim used in the
main text. The indispensable ingredient is a scale-fixing normalization
equivalent to whitening. In our formulation it is imposed as a hard whitening
constraint on the encoder classes; alternatively, it can be implemented inside
the objective as in standard normalized CCA, in which case
\autoref{prop:whitening-wlog} shows that the analysis may still be carried out
on whitened encoders. What fails without such normalization is twofold: the raw
covariance objective is unbounded by \autoref{prop:raw-cca-unbounded}, and the
row-orthonormal coefficient geometry required by \autoref{lem:diag-pick-topN}
disappears. Hence the affine identifiability proof does not go through without
whitening or an equivalent variance normalization.
\end{remark}

\section{Proof Sketches for the Remaining Candidate Distributions}
\label{sec:sketch}
\label{sec:sketches-non-gauss}

The Gaussian proof of \autoref{thm:affine-id} relies on three Gaussian-specific
ingredients: the scalar Hermite--Mehler expansion in
\autoref{lemma:hermite_expansion_bivariate_gaussian}, its tensor-product
extension in \autoref{prop:multi_variante_hermite}, and the resulting
orthogonality statement in \autoref{cor:multi_hermite_orthogonality}. For the
remaining candidate priors, the same optimization argument carries over once
these ingredients are replaced by the corresponding scalar Lancaster expansion
and the associated orthonormal polynomial basis
\citep{lancaster1958structure,eagleson1964polynomial}. The only
family-dependent objects are therefore the one-dimensional orthogonal
polynomials and the associated diagonal coefficient sequence.

\subsection{Unified Proof Template}

We state the common reduction at the level needed by the CCA proof. Throughout
this subsection, let \(I_i=\mathbb N_0\) for Poisson, negative binomial, and
gamma marginals, and let \(I_i=\{0,\dots,m_i\}\) for hypergeometric marginals,
where \(m_i\) is the maximal polynomial degree permitted by the finite support
of the \(i\)-th coordinate. In the hypergeometric case, all sums below are
therefore finite.

\begin{proposition}[Unified Lancaster reduction for the remaining candidate priors]
\label{prop:lancaster-reduction}
Assume that, for each coordinate \(i\in\{1,\dots,d_{\mathcal S}\}\), the pair
\((s_i,s_i')\) has a bivariate Lancaster law with common marginal \(\nu_i\),
orthonormal polynomial basis
\(\{\psi_{i,n}\}_{n\in I_i}\subset L^2(\nu_i)\) satisfying
\(\psi_{i,0}\equiv 1\), and coefficient sequence
\(\{\lambda_{i,n}\}_{n\in I_i}\) with \(\lambda_{i,0}=1\) such that
\begin{equation}
\label{eq:scalar_lancaster_expansion}
dP_{s_i,s_i'}(u,v)
=
\left(
\sum_{n\in I_i}\lambda_{i,n}\,\psi_{i,n}(u)\psi_{i,n}(v)
\right)\,
d\nu_i(u)\,d\nu_i(v).
\end{equation}
Assume moreover that the coordinate pairs
\(\{(s_i,s_i')\}_{i=1}^{d_{\mathcal S}}\) are independent across \(i\). Define
the tensor-product basis and multivariate Lancaster coefficients by
\[
\Psi_{\mathbf n}(\mathbf s)
:=
\prod_{i=1}^{d_{\mathcal S}}\psi_{i,n_i}(s_i),
\qquad
\lambda_{\mathbf n}
:=
\prod_{i=1}^{d_{\mathcal S}}\lambda_{i,n_i},
\qquad
\mathbf n=(n_1,\dots,n_{d_{\mathcal S}})\in \mathcal I:=\prod_{i=1}^{d_{\mathcal S}} I_i.
\]
Let \(\tilde{\mathbf h},\tilde{\mathbf h}'\) be whitened latent maps in source
space and expand their coordinates as
\[
\tilde h_r(\mathbf s)
=
\sum_{\mathbf n\in \mathcal I\setminus\{\mathbf 0\}}
\alpha_{r,\mathbf n}\,\Psi_{\mathbf n}(\mathbf s),
\qquad
\tilde h_q'(\mathbf s')
=
\sum_{\mathbf n\in \mathcal I\setminus\{\mathbf 0\}}
\beta_{q,\mathbf n}\,\Psi_{\mathbf n}(\mathbf s'),
\]
for \(r,q\in\{1,\dots,d_{\mathcal Z}\}\), and define the coefficient matrices
\[
\mathbf A
=
(\alpha_{r,\mathbf n})_{1\le r\le d_{\mathcal Z},\,\mathbf n\in \mathcal I\setminus\{\mathbf 0\}},
\qquad
\mathbf B
=
(\beta_{q,\mathbf n})_{1\le q\le d_{\mathcal Z},\,\mathbf n\in \mathcal I\setminus\{\mathbf 0\}}.
\]
Then:
\begin{enumerate}[label=\arabic*.]
    \item \textbf{Whitening constraints.}
    Because the tensor-product basis is orthonormal and the constant mode is
    excluded, the whitening conditions imply
    \[
    \mathbf A\mathbf A^\top=\mathbf I_{d_{\mathcal Z}},
    \qquad
    \mathbf B\mathbf B^\top=\mathbf I_{d_{\mathcal Z}}.
    \]

    \item \textbf{Diagonal cross-covariance.}
    The source-space cross-covariance diagonalizes as
    \[
    \mathrm{Cov}\big(\tilde{\mathbf h}(\mathbf s),\tilde{\mathbf h}'(\mathbf s')\big)
    =
    \mathbf A\,
    \mathrm{diag}\big((\lambda_{\mathbf n})_{\mathbf n\in \mathcal I\setminus\{\mathbf 0\}}\big)\,
    \mathbf B^\top .
    \]

    \item \textbf{Reduction to diagonal selection.}
    Consequently, by \autoref{lem:diag-pick-topN}, the source-space CCA
    objective equals the sum of the \(d_{\mathcal Z}\) largest absolute values
    among \(\{\lambda_{\mathbf n}:\mathbf n\in\mathcal I\setminus\{\mathbf 0\}\}\).
\end{enumerate}
If \(d_{\mathcal Z}=d_{\mathcal S}\) and
\begin{equation}
\label{eq:non_gaussian_first_order_dominance}
\min_{1\le i\le d_{\mathcal S}} |\lambda_{i,1}|
>
\sup_{\substack{\mathbf n\in \mathcal I\setminus\{\mathbf 0\}\\
\mathbf n\notin\{\mathbf e_1,\dots,\mathbf e_{d_{\mathcal S}}\}}}
|\lambda_{\mathbf n}|,
\end{equation}
then every population maximizer uses only the degree-one tensor basis
functions. Equivalently, there exist orthogonal matrices
\(\mathbf Q,\mathbf Q'\in O(d_{\mathcal Z})\) such that
\[
\tilde{\mathbf h}^*(\mathbf s)
=
\mathbf Q\,
\big(\psi_{1,1}(s_1),\dots,\psi_{d_{\mathcal S},1}(s_{d_{\mathcal S}})\big)^\top,
\]
\[
\tilde{\mathbf h}^{\prime *}(\mathbf s')
=
\mathbf Q'\,
\big(\psi_{1,1}(s_1'),\dots,\psi_{d_{\mathcal S},1}(s_{d_{\mathcal S}}')\big)^\top.
\]
Hence, whenever each first-degree basis function \(\psi_{i,1}\) is affine in
its argument, the optimal source-space maps are affine up to orthogonal
transformations.
\end{proposition}

\begin{sketchproof}
\textbf{1. Reduce to the latent space.}
By \autoref{prop:rep-inv}, it suffices to maximize \(J_{\mathcal S}\) over
whitened source-space maps \((\tilde{\mathbf h},\tilde{\mathbf h}')\).

\smallskip
\textbf{2. Expand the encoders in the orthonormal polynomial basis.}
Since \(\{\Psi_{\mathbf n}\}_{\mathbf n\in\mathcal I}\) is an orthonormal
tensor-product basis of \(L^2(P_{\mathbf s})\) and whitening enforces zero
mean, the constant term \(\mathbf n=\mathbf 0\) is absent from both
expansions. The identity-covariance constraints then yield
\(\mathbf A\mathbf A^\top=\mathbf I\) and
\(\mathbf B\mathbf B^\top=\mathbf I\).

\smallskip
\textbf{3. Diagonalize the cross-covariance.}
Independence across coordinates and the scalar Lancaster expansion
\eqref{eq:scalar_lancaster_expansion} imply
\[
\mathbb E\big[\Psi_{\mathbf m}(\mathbf s)\Psi_{\mathbf n}(\mathbf s')\big]
=
\lambda_{\mathbf n}\,\delta_{\mathbf m\mathbf n},
\qquad
\mathbf m,\mathbf n\in\mathcal I.
\]
Substituting the encoder expansions gives
\[
\mathrm{Cov}\big(\tilde{\mathbf h}(\mathbf s),\tilde{\mathbf h}'(\mathbf s')\big)
=
\mathbf A\,
\mathrm{diag}\big((\lambda_{\mathbf n})_{\mathbf n\in\mathcal I\setminus\{\mathbf 0\}}\big)\,
\mathbf B^\top .
\]

\smallskip
\textbf{4. Reduce CCA to selecting diagonal entries.}
By \autoref{lem:diag-pick-topN}, the CCA objective is maximized by selecting
the \(d_{\mathcal Z}\) largest absolute values among
\(\{\lambda_{\mathbf n}:\mathbf n\neq \mathbf 0\}\). Under
\eqref{eq:non_gaussian_first_order_dominance}, those top coefficients are
exactly the degree-one indices
\(\mathbf e_1,\dots,\mathbf e_{d_{\mathcal S}}\).

\smallskip
\textbf{5. Conclude affinity.}
Thus every maximizer lies in the span of the first-degree basis functions.
Since the first-degree orthonormal polynomial of a one-dimensional orthogonal
polynomial system is always affine in the underlying scalar variable, the
optimal source-space maps are affine. Pushing them back to observation space
via \autoref{prop:rep-inv} yields the same conclusion for the learned
encoders.
\end{sketchproof}

\begin{remark}
The non-Gaussian argument does \emph{not} require the Gaussian-specific
identity \(\lambda_{\mathbf n}=\prod_i \rho_i^{n_i}\). The proof only needs
diagonalization in an orthonormal polynomial basis and an ordering condition
such as \eqref{eq:non_gaussian_first_order_dominance}. Whenever a particular
family admits the geometric specialization \(\lambda_{i,n}=\rho_i^n\), the
dominance condition reduces to the Gaussian-style separation
\(\rho_{d_{\mathcal S}}>\rho_1^2\).
\end{remark}

\subsection{Instantiations for the Four Families}

For each family below, the first-degree orthonormal polynomial is precisely the
standardized coordinate,
\[
\psi_1(x)=\frac{x-\mu}{\sigma},
\]
with \(\mu\) and \(\sigma^2\) the marginal mean and variance. In particular,
\[
\lambda_{i,1}
=
\mathbb E\!\left[\psi_{i,1}(s_i)\psi_{i,1}(s_i')\right]
=
\mathrm{Corr}(s_i,s_i').
\]
Therefore, once higher-degree terms are excluded by
\eqref{eq:non_gaussian_first_order_dominance},
\autoref{prop:lancaster-reduction} immediately yields affine recovery.

\paragraph{Poisson \((\lambda)\).}
The marginal law is
\[
\nu(x)=e^{-\lambda}\frac{\lambda^x}{x!},
\qquad x\in \mathbb N_0.
\]
The associated orthogonal family is the Charlier family \(C_m(x;\lambda)\).
After normalization, the first-degree term is
\[
\psi_1(x)=\frac{x-\lambda}{\sqrt{\lambda}}.
\]

\paragraph{Negative Binomial \((r,p)\).}
We use the convention
\[
\nu(x)=\binom{r+x-1}{x}p^r(1-p)^x,
\qquad x\in \mathbb N_0,\quad 0<p<1.
\]
The associated orthogonal family is the Meixner family
\(M_m(x;\beta,c)\) with \(\beta=r\) and \(c=1-p\). Its mean and variance are
\[
\mu=\frac{r(1-p)}{p},
\qquad
\sigma^2=\frac{r(1-p)}{p^2},
\]
hence
\[
\psi_1(x)=\frac{x-\mu}{\sigma}.
\]

\paragraph{Hypergeometric \((N,K,n)\).}
The marginal law is
\[
\nu(x)=\frac{\binom{K}{x}\binom{N-K}{n-x}}{\binom{N}{n}},
\qquad
x=\max\{0,n-(N-K)\},\dots,\min\{n,K\}.
\]
The associated orthogonal family is the Hahn family associated with the
hypergeometric weight; its degree is finite because the support is finite. Its
mean and variance are
\[
\mu=n\frac{K}{N},
\qquad
\sigma^2=n\frac{K}{N}\left(1-\frac{K}{N}\right)\frac{N-n}{N-1},
\]
so the first-degree normalized polynomial is
\[
\psi_1(x)=\frac{x-\mu}{\sigma}.
\]

\paragraph{Gamma \((k,\theta)\).}
The marginal law is
\[
d\nu(x)=\frac{1}{\Gamma(k)\theta^k}x^{k-1}e^{-x/\theta}\mathbf 1_{(0,\infty)}(x)\,dx.
\]
The associated orthogonal family is the generalized Laguerre family
\(L_m^{(k-1)}(x/\theta)\). Its mean and variance are
\[
\mu=k\theta,
\qquad
\sigma^2=k\theta^2,
\]
hence
\[
\psi_1(x)=\frac{x-k\theta}{\sqrt{k}\,\theta}.
\]

\subsection{Connection to the Gaussian Proof}

The relationship to the Gaussian case is now transparent.

\begin{itemize}
    \item \textbf{Only the basis changes.}
    Replace the Hermite--Mehler system by the appropriate Lancaster system:
    Charlier for Poisson, Meixner for negative binomial, Hahn for
    hypergeometric, and generalized Laguerre for gamma.

    \item \textbf{The diagonalization step is identical.}
    Once the joint law is expanded in an orthonormal polynomial basis, the
    cross-covariance of the encoder coefficients is diagonal, and the CCA
    objective reduces to a diagonal selection problem exactly as in the Gaussian
    proof.

    \item \textbf{The optimization step is unchanged.}
    \autoref{lem:diag-pick-topN} still implies that the maximizer selects the
    \(d_{\mathcal Z}\) largest absolute Lancaster coefficients.

    \item \textbf{Affine recovery again follows from first-degree dominance.}
    If the degree-one coefficients dominate all higher-degree coefficients, only
    first-degree basis elements survive at the optimum. Since those first-degree
    basis elements are affine in the latent coordinates, the conclusion of
    \autoref{thm:affine-id} carries over verbatim.
\end{itemize}
\section{Implementation Details}\label{sec:implmentation}
\paragraph{Training Configurations.}
We adapt the experimental protocols of \citep{zimmermann2021contrastive, matthes2023towards}
to our dual-encoder framework. All models are optimized via Adam \citep{kingma2014adam}
with a constant learning rate of $10^{-4}$. For the synthetic dataset, we train for
$10^5$ iterations with a batch size of $1024$, requiring approximately $1.5$ hours for
$d_{\mathcal{S}}=40$ on a single NVIDIA RTX A5000 GPU. For 3DIdent, we train for $10^4$
iterations with a batch size of $512$, requiring roughly $4$ hours. Empirically,
we observe that CCA-based objectives converge $2$--$3\times$ faster than standard
contrastive losses. With the exception of qualitative visualizations, all reported
metrics are averaged across five independent random seeds. Comprehensive implementation
details and extended results are provided in \autoref{sec:results}.

For encoder architectures, we adopt the same design as \citep{zimmermann2021contrastive} for both synthetic and 3DIdent datasets,
except that for synthetic data we employ a lighter residual network.
This network comprises two hidden layers of sizes $10\cdot d_\mathcal{S}$ and $20\cdot d_\mathcal{S}$, followed by three residual blocks and an output layer.
Each residual block contains two layers of width $20\cdot d_\mathcal{S}$.
We apply leaky-ReLU activations and batch normalization to all hidden layers.

The decoders are implemented as approximately
invertible multi-layer perceptrons. The encoders $\mathbf f$ and $\mathbf f'$ are parameterized independently
using residual connections and batch normalization for optimization stability.

Training the synthetic model with $d_\mathcal{S}=10$ on a single RTX~A5000 GPU takes approximately one hour,
whereas experiments on 3DIdent require roughly four hours using eight RTX~A5000 GPUs.
\paragraph{Candidate Distribution Setups.}
For each latent dimension, we generate a paired source $(\mathbf{s}, \mathbf{s}') \in \mathbb{R}^{d_\mathcal S} \times \mathbb{R}^{d_\mathcal S}$
according to the additive latent model
$\mathbf{s} = \mathbf{a} + \mathbf{c}$, $\mathbf{s}' = \mathbf{b} + \mathbf{c}$,
where $\mathbf{a}, \mathbf{b}, \mathbf{c}$ are independent random vectors.
Unless otherwise stated, the coordinates are independent across dimensions except for Gaussian case as stated in \autoref{assump:gaussian-latent-pair}.

\textit{Joint Gaussian.}
We sample $(\mathbf{s}, \mathbf{s}') \sim \mathcal{N}(\mathbf{0}, \bm{\Sigma})$ with block covariance
\[
\bm{\Sigma} =
\begin{pmatrix}
\mathbf{I}_n & \mathbf{A}\,\mathrm{diag}(\bm{\rho})\,\mathbf{B}^\top \\
\mathbf{B}\,\mathrm{diag}(\bm{\rho})\,\mathbf{A}^\top & \mathbf{I}_n
\end{pmatrix},
\]
where $\mathbf{A}, \mathbf{B}$ are independent random orthogonal matrices
(constructed via QR decomposition of Gaussian matrices with $\det=+1$).
The canonical correlations $\bm{\rho} = (\rho_1, \ldots, \rho_n)$ are linearly spaced in
$[0.3,\,0.5]$.
This yields standard-normal marginals and the desired cross-covariance structure
$\mathbf{A}\,\mathrm{diag}(\bm{\rho})\,\mathbf{B}^\top$.

\textit{Joint Gamma.}
For each coordinate $j$, we draw three independent random variables
$c_j \sim \mathrm{Gamma}(k=1, \mathrm{rate}=1)$ and
$a_j, b_j \sim \mathrm{Gamma}(k=2, \mathrm{rate}=1)$,
and set
\[
s_j = a_j + c_j, \qquad s'_j = b_j + c_j.
\]
Then $\mathbb{E}[s_j] = \mathbb{E}[s'_j] = 3$,
$\mathrm{Var}(s_j) = \mathrm{Var}(s'_j) = 3$, and
$\mathrm{Cov}(s_j, s'_j) = 1$.

\textit{Joint Poisson.}
For each coordinate $j$,
$c_j \sim \mathrm{Poisson}(1)$ and
$a_j, b_j \sim \mathrm{Poisson}(2)$ independently, and
\[
s_j = a_j + c_j, \qquad s'_j = b_j + c_j.
\]
This yields $\mathbb{E}[s_j] = \mathbb{E}[s'_j] = 3$,
$\mathrm{Var}(s_j) = \mathrm{Var}(s'_j) = 3$, and
$\mathrm{Cov}(s_j, s'_j) = 1$.

\textit{Joint Negative Binomial.}
Each coordinate is generated as
$c_j \sim \mathrm{NegBin}(r=1, p=0.5)$ and
$a_j, b_j \sim \mathrm{NegBin}(r=2, p=0.5)$,
and we set
\[
s_j = a_j + c_j, \qquad s'_j = b_j + c_j.
\]
Under the PyTorch parameterization,
$\mathbb{E}[\mathrm{NegBin}(r, p)] = r(1-p)/p$ and
$\mathrm{Var} = r(1-p)/p^2$.
Hence $\mathbb{E}[s_j] = 3$,
$\mathrm{Var}(s_j) = 6$, and
$\mathrm{Cov}(s_j, s'_j) = 2$.
Note that this construction uses the \emph{negative binomial} distribution
(rather than the bounded binomial).

\textit{Joint Hypergeometric.}
For each dimension $j$, we first sample population and success counts
$N_j \sim \text{Uniform}\{10, \ldots, 20\}$ and
$M_j \sim \text{Uniform}\{1, \ldots, N_j - 1\}$,
and fix $n_{1j} = n_{2j} = 1$.
We draw without replacement $n_{1j} + n_{2j} = 2$ items
from a population of $N_j$ containing $M_j$ successes,
and define
\[
s_j = \text{\#successes in first draw}, \qquad
s'_j = \text{\#successes in second draw}.
\]
The marginals are $\mathrm{Hypergeometric}(N_j, M_j, n=1)$,
equivalently $\mathrm{Bernoulli}(p_j = M_j/N_j)$,
and the two draw indicators are negatively correlated with
\[
\mathrm{Cov}(s_j, s'_j)
= -\frac{M_j (N_j - M_j)}{N_j^2 (N_j - 1)}.
\]

\paragraph{Implementation Notes.}
Each sampling routine draws $\mathbf{s}$ and caches the paired
$\mathbf{s}'$ for retrieval during conditional evaluation;
it thus implements a paired sampler rather than a conditional generator.
All coordinates are sampled independently across $j$.

\section{Further Experimental Details and Additional Experimental Results}\label{sec:results}

\autoref{tab:synthetic_identifiability_pa_max} and \autoref{tab:synthetic_identifiability_pa_mean} gives the subspace errors measured by maximal and mean principal angels on synthetic data across five candidate distributions.
\begin{table*}[ht!]
    \centering
    \resizebox{\linewidth}{!}{
    \begin{tabular}{l*{10}{c}}
        \toprule
        Methods   & \multicolumn{2}{c}{Gaussian}  & \multicolumn{2}{c}{Negative Binomial}  & \multicolumn{2}{c}{Gamma}    & \multicolumn{2}{c}{Poisson}    & \multicolumn{2}{c}{Hypergeometric} \\
        \cmidrule(r){2-3}
        \cmidrule(r){4-5}
        \cmidrule(r){6-7}
        \cmidrule(r){8-9}
        \cmidrule(r){10-11}
            & $\mathbf f$ & $\mathbf f'$   & $\mathbf f$ & $\mathbf f'$   &$\mathbf f$ & $\mathbf f'$   &$\mathbf f$ & $\mathbf f'$   &$\mathbf f$ & $\mathbf f'$  \\
        \midrule
        SwAV & 89.26 $\pm$0.55 & 88.88 $\pm$0.45 & 89.64 $\pm$0.2 & 89.65 $\pm$ 0.26 & 89.49 $\pm$0.25 & 89.32 $\pm$0.46 & 89.67 $\pm$0.27 & 89.5 $\pm$ 0.33 & 86.48 $\pm$0.27 & 89.5 $\pm$0.33\\
        BarlowTwins    & 89.44$\pm$0.86 &89.51 $\pm$0.63             & 85.53 $\pm$ 0.86 & 85.36 $\pm$ 0.59 & 86.16 $\pm$ 1.15 & 88.6 $\pm$ 0.58 & 87.73 $\pm$0.86 & 87.43 $\pm$0.91 &88.93 $\pm$0.51 &89.59 $\pm$0.58\\
        VICReg  &  88.99 $\pm$0.54 & 89.07 $\pm$0.61 & 89.36 $\pm$0.48 & 89.16 $\pm$0.73 & 89.6 $\pm$0.24 & 89.03 $\pm$0.59 & 89.66 $\pm$0.41 & 88.38 $\pm$0.4 & 88.95 $\pm$0.43 & 89.29 $\pm$0.22 \\
        W-MSE & \textbf{7.83 $\pm$0.31} & \textbf{8.11 $\pm$0.24} & \textbf{8.5 $\pm$0.18} & \textbf{8.17 $\pm$0.19} & 8.4 $\pm$0.37 & \textbf{8.01 $\pm$0.32} & \textbf{7.63 $\pm$ 0.42} & 7.61 $\pm$ 0.2 & 8.72 $\pm$0.22 & 8.5 $\pm$ 0.18\\
        DGCCA & 89.33 $\pm$ 1.19 & 88.17 $\pm$0.76 & 89.12 $\pm$0.51 & 89.38 $\pm$0.68 & 88.60 $\pm$0.61 & 89.06 $\pm$0.62 & 89.21 $\pm$0.61 & 89.3 $\pm$ 0.62 & 89.21 $\pm$0.7 & 88.39 $\pm$0.59\\
        DeepCCA   & 8.88 $\pm$0.28 & 8.22 $\pm$0.23 &  8.79 $\pm$0.66 & 9.65 $\pm$0.3  & \textbf{7.93 $\pm$0.4} & 8.81 $\pm$0.45 & 11.43 $\pm$0.59& \textbf{7.37 $\pm$0.45} &\textbf{8.61 $\pm$0.24}&\textbf{7.39 $\pm$0.45}\\
        \bottomrule
     \end{tabular}}
        \caption{Comparison of the maximal principal angles $PA_{max}$$\downarrow$(°) of both encoders $\mathbf f, \mathbf f'$ on synthetic data ($d_\mathcal{S}=d_\mathcal{Z}=10$).}
        \label{tab:synthetic_identifiability_pa_max}
\end{table*}
\begin{table*}[ht!]
    \centering
    \resizebox{\linewidth}{!}{

    \begin{tabular}{l*{10}{c}}
        \toprule
        Methods   & \multicolumn{2}{c}{Gaussian}  & \multicolumn{2}{c}{Negative Binomial}  & \multicolumn{2}{c}{Gamma}    & \multicolumn{2}{c}{Poisson}    & \multicolumn{2}{c}{Hypergeometric} \\
        \cmidrule(r){2-3}
        \cmidrule(r){4-5}
        \cmidrule(r){6-7}
        \cmidrule(r){8-9}
        \cmidrule(r){10-11}
            & $\mathbf f$ & $\mathbf f'$   & $\mathbf f$ & $\mathbf f'$   &$\mathbf f$ & $\mathbf f'$   &$\mathbf f$ & $\mathbf f'$   &$\mathbf f$ & $\mathbf f'$  \\
        \midrule
        SwAV & 54.95 $\pm$1.75 & 55.18 $\pm$0.95 & 59.13 $\pm$10.12 & 62.48 $\pm$ 9.98 & 56.68 $\pm$0.59 & 55.17 $\pm$ 0.6 & 57.57 $\pm$0.46 & 56.39 $\pm$0.33 &63.21 $\pm$9.22 & 68.92 $\pm$8.69 \\
        BarlowTwins    & 33.91$\pm$0.16 &32.78 $\pm$0.23             & 30.27 $\pm$0.18 & 28.66 $\pm$0.15 & 33.6 $\pm$ 0.22 & 32.29 $\pm$ 0.17 & 42.18 $\pm$ 0.15 & 40.58 $\pm$ 0.22 &60.77 $\pm$ 0.29 & 60.72 $\pm$ 0.23\\
        VICReg  &  68.47 $\pm$0.44 & 67.89 $\pm$0.63 & 67.56 $\pm$0.45 & 68.91 $\pm$0.32 & 74.27 $\pm$0.18 & 70.39 $\pm$0.25 & 75.46 $\pm$0.27 & 72.95 $\pm$0.31 & 65.8 $\pm$0.43 & 64.75 $\pm$0.29 \\
        W-MSE & 5.12 $\pm$0.09 & 5.15 $\pm$0.03 & 5.21 $\pm$ 0.09 & \textbf{4.74 $\pm$0.07} & 4.75 $\pm$0.06 & \textbf{4.57 $\pm$0.1} & \textbf{4.69 $\pm$ 0.1} & 4.37 $\pm$0.04 & 5.38 $\pm$0.05 & \textbf{5.13 $\pm$0.02}\\
        DGCCA & 67.22 $\pm$ 0.38 & 65.91 $\pm$0.71 & 65.65 $\pm$0.23 & 64.89 $\pm$0.17 & 67.27 $\pm$0.11 & 66.12 $\pm$0.32 & 68.24 $\pm$ 0.25 & 68.82 $\pm$0.33 & 63.12 $\pm$ 0.23 & 63.58 $\pm$0.17\\
        DeepCCA   & \textbf{5.06 $\pm$0.08} & \textbf{4.92 $\pm$0.09} &  \textbf{5.00 $\pm$0.11} & 5.16 $\pm$0.05 & \textbf{4.65 $\pm$0.1} & 4.86 $\pm$0.08 & 5.18 $\pm$0.07& \textbf{4.06 $\pm$0.08} &\textbf{5.37 $\pm$0.07}&6.17 $\pm$0.07\\
        \bottomrule
    \end{tabular}}
        \caption{Comparison of the mean principal angles $PA_{\text{mean}}$$\downarrow$(°) of both encoders $\mathbf f, \mathbf f'$ on synthetic data ($d_\mathcal{S}=d_\mathcal{Z}=10$).}
        \label{tab:synthetic_identifiability_pa_mean}
\end{table*}
\end{document}